%% file: main.tex
\documentclass[%
 reprint,
 amsmath,amssymb,
 aps,
]{revtex4-2}

\def\thesection{\arabic{section}}

\usepackage{graphicx}
\usepackage{dcolumn}
\usepackage{bm}
\usepackage{stmaryrd}
\usepackage{bbm}
\usepackage{amssymb}
\usepackage{algorithm}
\usepackage{algorithmic}

\usepackage{microtype}
\usepackage{graphicx}
\usepackage{subfigure}
\usepackage{booktabs} 
\usepackage{enumitem}
\usepackage{bm}

\usepackage{hyperref}
\usepackage{physics}

\usepackage{amsmath}
\usepackage{amssymb}
\usepackage{mathtools}
\usepackage{amsthm}
\usepackage{xfrac}
\usepackage{bbm}
\usepackage{bigints}
\usepackage{xcolor}

\usepackage[capitalize,noabbrev]{cleveref}

\theoremstyle{plain}
\newtheorem{theorem}{Theorem}[section]

\newtheorem{Conjecture}[theorem]{Conjecture}

\theoremstyle{definition}

\theoremstyle{remark}

\DeclareMathOperator{\x}{\mathbf{x}}

\newcommand{\camera}[1]{{\color{black} #1}}

\begin{document}
\preprint{}

\title{Optimal Learning of Deep Random Networks of Extensive-width}

\author{Hugo Cui}
\affiliation{Statistical Physics Of Computation lab,
Institute of Physics, \'Ecole Polytechnique F\'ed\'erale de Lausanne, 1015 Lausanne, Switzerland}
\author{Florent Krzakala}
\affiliation{Information Learning and Physics lab,
Institute of Electrical Engineering, \'Ecole Polytechnique F\'ed\'erale de Lausanne, 1015 Lausanne, Switzerland}
\author{Lenka Zdeborov\'a}
\affiliation{Statistical Physics Of Computation lab,
Institute of Physics, \'Ecole Polytechnique F\'ed\'erale de Lausanne, 1015 Lausanne, Switzerland}

\begin{abstract}
We consider the problem of learning a target function corresponding to a deep, extensive-width, non-linear neural network with random Gaussian weights. We consider the asymptotic limit where the number of samples, the input dimension and the network width are proportionally large. We propose a closed-form expression for the Bayes-optimal test error, for regression and classification tasks. 
We further compute closed-form expressions for the test errors of ridge regression, kernel and random features regression. 
We find, in particular, that optimally regularized ridge regression, as well as kernel regression, achieve Bayes-optimal performances, while the logistic loss yields a near-optimal test error for classification. We further show numerically that when the number of samples grows faster than the dimension, ridge \& kernel  methods become suboptimal, while 
neural networks achieve test error close to zero from quadratically many samples. 
\end{abstract}
\maketitle

\section{Introduction}

Learning with neural networks has proven to be an extraordinarily versatile tool to approximate (learn) non-trivial functions from data. Many fundamental theoretical questions, however, remain open. For instance, the determination, for a given target function, of just \textit{how many} training data samples are needed in order to learn the target to a given precision? This is tantamount to determining the minimal error that can be achieved from a training set of a given size.\\

While for a generic target function and generic training set this question is very challenging, valuable insight can be accessed by studying simplified settings with Gaussian input data and specific target functions with known functional forms.
Of particular interest is the rich class of functions given by {\it random neural networks}. The lowest achievable test error is known to be obtained through Bayesian inference of the parameters of the target function, assuming (as we will) the distribution of the parameters is given. The {\it Bayes-optimal} test error corresponds to the information-theoretically minimal  test error that any algorithm can achieve. In the context of Gaussian data, with target functions being {\it single-layer random neural networks} the problem was studied as early as in \cite{Opper1991GeneralizationPO,seung1992statistical,watkin1993statistical,schwarze1993learning}. More recently, \cite{Barbier2017OptimalEA} provide a rigorous characterization of the Bayes-optimal error in the asymptotic proportional regime, where the number of samples is proportional to the input dimension and both of them are large with a fixed ratio $\alpha$. 
These results were then extended in \cite{schwarze1993learning,Aubin2018TheCM} to neural networks with one {\it narrow} hidden layer, whose width remains of order one in the above limit of large dimension and a proportional number of samples.\\ 

In practice, neural networks are trained using Empirical Risk Minimization (ERM) methods, and it is hence also important to know whether those methods are able to achieve the Bayes-optimal error.
\cite{thrampoulidis2018precise,montanari2019generalization,Hastie2019SurprisesIH,Mei2019TheGE,Aubin2020GeneralizationEI,Loureiro2021CapturingTL}, between others, addressed this question for Gaussian data, providing closed-form formulas for the ERM test error for generalized {\it linear} models for target functions corresponding to single-layer neural network with random weights from a number of samples proportional to the dimension.\\

Here, we pursue these lines of work and study a target function given by a {\it deep non-linear neural network with random weights}, in the limit where the layers-widths and the input dimension are comparably large, hereafter referred to as the \textit{extensive-width} regime. We call such target function the \textit{deep extensive-width random network}. We consider Gaussian input data. Our main question is the characterization of the test error that can be achieved information-theoretically from a given number of samples, as well as its reachability with ERM approaches. While the assumptions of Gaussian input data and the prescribed target function seem far from current machine-learning practice, from a theoretical point of view, these questions remain challenging and widely open even in such a simplified setting (even for a single hidden layer).
It is hard to imagine that we could obtain a plausible theory of deep learning without being able to answer such questions first.\\

\paragraph{Main contributions}

For the target function corresponding to the deep extensive-width random network and random Gaussian input data we obtain the following results:
\begin{itemize}[wide=1pt, noitemsep, topsep=0pt]
\item We conjecture a closed-form characterization for the asymptotic Bayes-optimal error, for regression and classification tasks, in the proportional regime where the number of samples $n$ scales linearly with the input dimension $d$.  


\item A fundamental step in our derivation, of independent interest, is the deep (Bayes) Gaussian Equivalence Property conjecture (GEP) , which specifies the Gaussian statistics of the output of deep networks whose weights are Gaussian, or sampled from the Bayes posterior. We show how the GEP follows from Bayes theory and the asymptotic concentration of random variables in the proportional regime. 

\item We contrast the Bayes-optimal test error to test errors achieved by simple ERM methods. For regression, ridge and kernel regression are found to achieve the Bayes-optimal mean-squared error, provided they are optimally regularized. An explicit formula for optimal regularization is provided. These results establish that it is impossible to learn more than a linear estimator of the target extensive-width network from linearly many samples. In the case of classification, logistic and ridge classification are found to yield test errors close (but not equal) to the Bayes error.

\item We provide a numerical exploration of the  regime where the number of samples $n$ tends to infinity \textit{faster} than linearly with the input dimension $d$, in which the deep (Bayes) GEPs can no longer be employed. We show that ridge and kernel  methods then cease to be optimal, while gradient-trained neural networks manage to almost perfectly learn the target, evidencing the superiority of neural nets.\\

\end{itemize}
A repository with the code employed in the present work can be found \href{https://github.com/HugoCui/Bayes_extensive}{here}.

\subsection{Related works}

\paragraph{Bayesian learning of neural networks}
It is well known that Bayesian learning using networks of infinite width (i.e. width much larger than the number of samples and the input dimension) is equivalent to kernel regression \cite{Neal1996PriorsFI,Lee2017DeepNN,Lee2019WideNN, Matthews2018GaussianPB, Hron2020ExactPD}. A theoretical analysis for extensive-width, however, proved for a long time a challenging endeavor.  \camera{\cite{Yaida2019NonGaussianPA, Roberts2021ThePO, ZavatoneVeth2021AsymptoticsOR}  computed (perturbative) first-order corrections to the mean test error with respect to the infinite width limit, but only accommodate a finite number of training samples. The recent work of \cite{ZavatoneVeth2022ContrastingRA, Hanin2022BayesianIW} respectively provide an asymptotic and non-asymptotic study of Bayesian learning, but are limited to  linear activations. }\cite{Li2021StatisticalMO} and \cite{Ariosto2022StatisticalMO} conjecture that in the proportional regime, i.e. $n\sim d$, the estimator yielded by extensive-width networks with ReLU or sign activations is still given by the associated Gaussian Process (GP) kernel, with the width only rescaling the variance term in the test error. We note that these works rely on a heuristic Gaussianity assumption and provide expressions depending explicitly on the entire dataset. Here instead we address specifically the Bayes-optimal performance for a random network target function and Gaussian inputs, which allows us to provide closed-form \textit{scalar} formulae and leverage the principled GEP to characterize the statistics. \camera{Finally, while all the previously cited work study the case of Bayesian regression with a square log likelihood, the present work also covers classification settings.}\\

\paragraph{Replica method in ML} The replica method has been applied in a sizeable body of work to access asymptotic characterizations of the test error (Bayes or ERM) in a variety of setups \cite{seung1992statistical,watkin1993statistical}. While being heuristic, its predictions have been proven rigorously in many cases, e.g. \cite{talagrand2006parisi,Barbier2017OptimalEA}. This toolbox has been successfully deployed to analyze architectures with one trainable layer, including generalized linear models \cite{advani2016statistical,Aubin2020GeneralizationEI,Cui2019LargeDF,Maillard2020PhaseRI,Loureiro2021CapturingTL}, narrow networks with frozen readout \cite{Aubin2018TheCM}, random features (RF) \cite{Gerace2020GeneralisationEI} and kernel methods \cite{Canatar2020SpectralBA,Cui2021GeneralizationER,Cui2022ErrorRF}. Recently \cite{ZavatoneVeth2022ContrastingRA} studied the multiple layers case, in the framework of linear networks. Here we go a step further and analyze deep non-linear networks. \\

\camera{
\paragraph{The proportional regime} The proportional $n\sim d$ regime has been investigated for shallow networks in a sizeable body of work, leveraging tools like the convex gaussian minimax theorem \cite{thrampoulidis2018precise,Aubin2020GeneralizationEI,Loureiro2021CapturingTL, montanari2019generalization}, random matrix theory \cite{el2008spectrum,Pennington2019NonlinearRM,Louart2017ARM} or approximate message-passing \cite{Aubin2018TheCM,Gabrie2019MeanfieldIM}, in addition to the replica method.\\
}

\paragraph{Gaussian Equivalence} 
The equivalence between the asymptotic test error of simple ERM algorithms with that of the associated problem where the data samples are replaced by Gaussian covariates with matching population covariance has been observed in many situations, starting with the seminal work \cite{el2008spectrum} on kernel matrices. In particular, \cite{Goldt2021TheGE,goldt2022gaussian,Montanari2022UniversalityOE,Hu2020UniversalityLF} have proven a \textit{Gaussian Equivalence} principle that shows that, in the proportional regime, one can often replace projected data with Gaussian ones. Such equivalences were used, for instance, in \cite{Loureiro2021CapturingTL} to characterize the ERM test error in a variety of setups,
in terms solely of the population covariances of the target/learner networks. \camera{Concomitant works \cite{DRF2023, Bosch2023PreciseAA} characterize the Gaussian universality of the test error of deep learners with fixed random weights and trainable readout.}

\section{Setting}

\label{sec:setting}
We consider the problem of learning from a train set $\mathcal{D}=\{\x^\mu,y^\mu\}_{\mu=1}^n$, with $n$ independently sampled Gaussian covariates $\x^\mu\in \mathbb{R}^d\sim\mathcal{N}(0,\Sigma)$. The covariance $\Sigma$ is assumed to admit a well-defined limiting spectral distribution $\mu$ as $d\rightarrow \infty$ with finite non-zero first and second moments. The labels $y^\mu$ are assumed to be generated by a  $L$-layers deep network with random weights. Denoting $\xi\sim \mathcal{N}(0,\Delta)$ a Gaussian additive output noise, we have
\begin{align}
\label{eq:teacher}
   y^\mu =f_\star\Big[\frac{1}{\sqrt{k_{L}}}{\bf a}_\star^\top  \underbrace{\left(
    \varphi^\star_{L}\circ  \dots\circ \varphi^\star_2\circ\varphi_1^\star\right)}_{L}
    (\x^\mu)+\xi \Big],\\
    \text{with layers}\,\,\varphi^\star_\ell(\x)=\sigma_\ell\left(
    \frac{1}{\sqrt{k_{\ell-1}}}W^\star_\ell \cdot \x
   \right). \nonumber
\end{align}
$(\sigma_\ell)_{\ell=1}^{L}$ is a sequence of activation functions, which are assumed to be odd for simplifying technical reasons. The readout function $f_\star(\cdot)$ will be taken to be the identity function for regression, and the sign function for classification. The width of the $\ell$-th layer is denoted $k_\ell$, and the associated weight matrix is $W^\star_\ell\in\mathbb{R}^{k_\ell\times k_{\ell-1}}$, with elements sampled i.i.d from $\mathcal{N}(0,\Delta_\ell)$. Similarly, the readout weight vector $a_\star  $ is sampled from $\mathcal{N}(0,\Delta_a \mathbb{I}_{k_L})$.\\

We wish to characterize the Bayes-optimal test errors when learning on data produced by the target function \eqref{eq:teacher}. We consider that all the hyperparameters $L, k_\ell, \sigma_\ell, \Delta_a, \Delta_\ell,\Delta$ of the architecture \eqref{eq:teacher} are known, but the weights $a_\star,\{W_\ell^\star\}_{\ell=1}^L$ are not known to the learner.\\ 

Throughout Sec.~\ref{sec:Bayes_regression} to \ref{sec:ERM_regression}, we consider the \textit{proportional regime}: the high-dimensional asymptotic limit where $\forall \ell, ~n,d,k_\ell \!\rightarrow\! \infty$ with fixed $\mathcal{O}(1)$ ratios $\alpha\equiv \sfrac{n}{d}$ and $\gamma^\star_\ell\equiv\sfrac{k_\ell}{d}$. The parameters $L,\Delta_\ell,\Delta_a,\Delta$ are assumed to be $\mathcal{O}(1)$. The \textit{quadratic regime} $n\sim d^2\sim k_\ell^2$
is numerically explored in Sec. \ref{sec:quadratic}. 
It is known that learning a target of large width $k$ (resp. using a network of large width) with a finite number of samples $n=\mathcal{O}_k(1)$ simplifies drastically to the problem of learning a Gaussian process (resp. kernel regression) \cite{Neal1996PriorsFI,Lee2017DeepNN, Matthews2018GaussianPB}. We consider here widths $\{k_\ell\}_{\ell=1}^L$ at most \textit{comparable}, and not very large compared to, the input dimension $d$ and the number of samples $n$, which makes for a non-trivial, and much richer, learning problem.

\section{Bayes-optimal Error}

\label{sec:Bayes_regression}
The Bayes-optimal error for data generated using the target function \eqref{eq:teacher} is achieved by sampling the weights ${\bf a}, \{W_\ell\}_\ell$ from a posterior measure involving a neural network of matching architecture. We thus define 
\begin{align}
\label{eq:student}
&    \hat{y}(\x)=\frac{1}{\sqrt{k_{L}}}{\bf a}^\top  \underbrace{\left(
    \varphi_{L}\circ \varphi_{L-1}\circ \dots\circ \varphi_2\circ\varphi_1\right)}_{L}
    (\x) ,\\
 &\text{with layers}\,\,
    \varphi_\ell(\x)=\sigma_\ell\left(
    \frac{1}{\sqrt{k_{\ell-1}}}W_\ell \cdot \x
    \right).
    \end{align}
The Bayes-optimal Mean Squared Error (MSE) is then 
\begin{align}
\label{eq:MSE}
\epsilon_{g,\mathrm{reg}}^{\mathrm{BO}}\!=\!\mathbb{E}_{\mathcal{D}, \{W_\ell^\star\}_{\ell=1}^L, {\bf a}_\star}\mathbb{E}_{\x,y}\!\left[\!
    \big(y\!-\!\langle \hat{y}(\x)\rangle_{{\bf a},\{W_\ell\}_{\ell=1}^L\sim\mathbb{P}}\big)^2\!
    \right]
\end{align}
where $\x,y$ should be understood as a test sample. 
The Bayes-optimal classification error (defined as the probability to wrongly classify a test sample) is given by
\begin{align}
    \label{eq:Bayes_classif} &\epsilon_{g,\mathrm{class}}^{\mathrm{BO}}=\mathbb{E}_{\mathcal{D}, \{W_\ell^\star\}_{\ell=1}^L, {\bf a}_\star}\notag\\
    &\mathbb{P}_{\x,y}\left[
    y\ne \mathrm{sign} \left(\langle \mathrm{sign}(\hat{y}(\x))\rangle_{{\bf a},\{W_\ell\}_{\ell=1}^L\sim\mathbb{P}}\right)
    \right].
\end{align}
In (\ref{eq:MSE},\ref{eq:Bayes_classif}), the learner network is averaged over the posterior
\begin{align}
    \label{eq:Bayes_post_classif}
    & \mathbb{P}\left[ {\bf a},\{ W_\ell \}_{ \ell=1 }^L | \mathcal{D} \right]  \propto e^{-\frac{||{\bf a}||^2}{2\Delta_a}-\sum\limits_{\ell}\frac{||W_\ell||_F^2}{2\Delta_\ell}
}\notag\\
&\times \prod\limits_{\mu=1}^n\int \frac{d\xi e^{-\frac{1}{2\Delta}\xi^2}}{\sqrt{2\pi\Delta}}\delta\left[
y^\mu-f_\star(\hat{y}(\x^\mu)+\xi)
\right].
\end{align}
The Bayes errors \eqref{eq:MSE} and \eqref{eq:Bayes_classif} provide information-theoretic lower bounds on the test error for learning the target \eqref{eq:teacher}, in the sense that no learning algorithm can reach better performance when learning from the dataset $\mathcal{D}$.\\

Accessing numerically the Bayes errors \eqref{eq:MSE} and \eqref{eq:Bayes_classif} requires sampling an $\mathcal{O}(d^2)$-dimensional distribution, a difficult task. It is on the other hand possible to theoretically derive closed-form formulas using the replica method \cite{Replica, Mzard2009InformationPA} that allows characterizing the Bayes error in terms of the moments of independent instances of $\hat{y}(\x)$ (the eponymous replicas) drawn from the posterior eq.~\eqref{eq:Bayes_post_classif}. In the replica calculation, one averages over the randomness in the model and in order to be able to carry out such  averages in a closed form, the Gaussian equivalence property described in the next section is crucial.

\subsection{The Bayesian Gaussian Equivalence Property}

A seminal step in our analytical approach is the property that we can replace the statistics of the output $\hat{y}(\x)$ with respect to the randomness of the input $\x$ by Gaussian, with a covariance depending linearly on the covariance of the weight matrices $W_l$. In fact, \cite{Li2021StatisticalMO,Ariosto2022StatisticalMO} do rely on a related Gaussianity assumption, which \cite{Ariosto2022StatisticalMO} heuristically justify for $L=1$ using the Breuer-Major theorem, for generic datasets. Since in the present work, we consider the specific Bayes-optimal setting, we are in a position to state a  more principled conjecture which follows from the GEP \cite{Goldt2020ModellingTI} and the Nishimori identities \cite{Nishimori2001StatisticalPO,Iba1998TheNL}.

\begin{Conjecture}
\label{prop:BGP_2l}(\textbf{Shallow Bayes GEP}) Consider ${\bf x}$ a random Gaussian vector. Then for $L\!=\!1$, in the extensive-width asymptotic limit $d,k_1\rightarrow\infty$ with fixed $\mathcal{O}(1)$ ratio $\gamma_1=\sfrac{k_1}{d}$, any finite number of replicas $\hat{y}^1(\x;W^1_1,{\bf a}^1),...,\hat{y}^s(\x;W^s_1,{\bf a}^s)$ independently drawn from the Bayes posterior ~\eqref{eq:Bayes_post_classif} are jointly Gaussian. Furthermore, their correlation reads $\mathbb{E}_{\x}\hat{y}^a(\x)\hat{y}^b(\x)=\sfrac{{\bf a}^{a\top} \Omega_1^{ab} {\bf a}^b}{k_1}$ where $\Omega_1^{ab}$ is the population covariance of the last layer post-activations $\mathbb{E}_{\x}\varphi_1^a(\x)\varphi_1^b(\x)^\top$  that reads
\begin{equation}
    \Omega_1^{ab}=\left(\kappa_1^{(1)}\right)^2\frac{W_1^a \Sigma W_1^{b\top}}{d}+ \delta_{a,b}\left(\kappa_*^{(1)}\right)^2\mathbb{I}_{k_1}\, ,
    \label{cov-L1}
\end{equation}
where $\kappa_1^{(1)}=\sfrac{\mathbb{E}_{z}\left[z\sigma_1(z)\right]}{r_1}$ and $(\kappa_*^{(1) })^2=\mathbb{E}_{z}\left[\sigma_1(z)^2\right]-r_1(\kappa_1^{(1)})^2$, with $r_1=\Delta_1\sfrac{\Tr\Omega_1}{d}$ and $z\sim\mathcal{N}(0,r_1)$.
\end{Conjecture}

We now explain how Conjecture \ref{prop:BGP_2l} is motivated. In the proportional regime, for $a=b$, \camera{conditional on the} matrix $W_1$, the Gaussian Equivalence Theorem of \cite{Goldt2021TheGE,Hu2020UniversalityLF,Montanari2022UniversalityOE} prove indeed that the model (\ref{eq:student}) for $L=1$ 
\camera{shares the same second-order post-activation statistics as the noisy \textit{linear} network} $\hat y = {\bf a}^T (\kappa_1^{(1)} \sfrac{W_1 {\x}}{\sqrt{d}} + \kappa_* ^{(1)}Z)$ (with $Z$ a random Gaussian variable), thus leading to the covariance (\ref{cov-L1}). 
\camera{This} so-called one-dimensional central limit theorem (1dCLT) \cite{Goldt2021TheGE,Hu2020UniversalityLF,Montanari2022UniversalityOE} holds under some strict assumptions on the weight matrix $W_1$, that are satisfied in particular for random matrices with independent entries.\\

In the Bayesian setting one needs to integrate over the posterior distribution of the matrix $W_1$, learned from the data. For Conjecture \ref{prop:BGP_2l} to be valid, the conditions of the 1dCLT must be satisfied with high probability over the learned matrices, which is by no means a trivial requirement. This is where the properties of Bayes-optimal inference come in handy: indeed, a classic property of Bayesian learning (often called the Nishimori property \cite{Nishimori2001StatisticalPO,Iba1998TheNL,Zdeborov2015StatisticalPO}) is that the statistics of weights drawn from the Bayes posterior is exactly the same as the one of the target network weights. This is a direct consequence of the Bayes formula (see e.g. section 1.2.3. in \cite{Zdeborov2015StatisticalPO}). As a consequence, the learned matrices are following Gaussian statistics as well (given this is the statistics of the target ones by definition), and thus respect the conditions of the 1dCLT.\\

\camera{When considering different replicas ($a\ne b$), the Nishimori conditions ensure that one of the two replicas can be taken, without loss of generality, to be the target weight $W_1^\star$. Since $W_1$ is learnt and therefore generically correlated with the target weights $W_1^\star$, the assessment of the covariance $\Omega^{ab}_1$ is a challenging task. However, the results of \cite{Aubin2018TheCM} suggest that $W_1$ is asymptotically uncorrelated with $W_1^\star$ for sample complexities $\alpha \lesssim k_1$. Since we consider here $\alpha=\mathcal{O}(1)\ll k_1$, this motivates the following conjecture:
\begin{Conjecture}
\label{conj:non-spec}
    (Non-specialization) for $L\!=\!1$, in the asymptotic limit $n,d,k_1\rightarrow\infty$ with fixed $\mathcal{O}(1)$ ratio $\gamma_1=\sfrac{k_1}{d}$ and $\alpha=\sfrac{n}{d}$, let $W_1$ be sampled from the Bayes posterior \eqref{eq:Bayes_post_classif}. Then with high probability $W_1$ has vanishing overlap with $W_1^\star$, i.e.
    \begin{align}
        \frac{1}{d}\underset{1\le i,j \le k_1}{\max}\left( W_1^\star \Sigma W_1^\top\right)_{i,j} =\mathcal{O}\left(\sfrac{1}{\sqrt{d}}\right).
    \end{align}
\end{Conjecture}
\ref{conj:non-spec} implies that the second term in the right-hand side of \eqref{cov-L1} is only present for $a=b$. The detailed derivation of \eqref{cov-L1} is presented in Appendix \,\ref{App:GET}
}

 \subsection{Deep (Bayesian) Gaussian Equivalence Property}
 
 We next discuss how these results generalize to deep networks ($L\ge 2$). 
 While a sizeable body of work has been devoted to the distribution induced by the random weights for fixed inputs \cite{Lee2017DeepNN, Matthews2018GaussianPB, Hanin2022BayesianIW,Hanin2022CorrelationFI,Yaida2019NonGaussianPA}, little is known, in the deep case, for the distribution induced by the input distribution, for \textit{fixed} weights.  While there is no proof of the equivalence of the 1dCLT of \cite{Goldt2021TheGE,Hu2022SharpAO} for $L\!\ge\!2$, we provide, in  App.~\ref{App:GET}, numerical evidence of the following conjecture:

 \begin{Conjecture}\label{prop:dGP} The output $\hat{y}(\x)$ of a deep random network, \camera{conditional on its Gaussian weights $\{W_\ell\}_{\ell=1}^L,{\bf a}$}, in the extensive-width limit $d\rightarrow \infty$ and $\forall \ell, k_\ell\rightarrow\infty$ with fixed ratios $\gamma_\ell=\sfrac{k_\ell}{d}$, is asymptotically Gaussian with respect to $\x$.
 \end{Conjecture}

\camera{Conjecture \ref{prop:dGP} thus extends the first part of \ref{prop:BGP_2l} to the deep setting. This intuitively follows from the fact that higher order cumulants of the post-activations at intermediary layers are asymptotically suppressed (as shown in \cite{Fischer2022DecomposingNN}) and thus approximately Gaussian -- allowing one to iterate \ref{prop:BGP_2l}.} App.\,\ref{App:GET} further establishes a closed-form expression for the variance of $\hat{y}(\x)$, which like the shallow case \ref{prop:BGP_2l} is amenable to being interpreted in terms of an equivalent noisy network. We defer the discussion of the  latter to the subsection \ref{subsec:BO_error}. Finally, the Nishimori property again ensures that conjecture \ref{prop:dGP} transfers to weights sampled from the Bayes posterior \eqref{eq:Bayes_post_classif}. Defining the following recursion on $\{r_\ell\}_{\ell=1}^L$, $\{\kappa_1^{(\ell)}\}_{\ell=1}^L$ and $\{\kappa_*^{(\ell)}\}_{\ell=1}^L$:
\begin{align}
\label{eq:kappa_multilayer}
&r_{\ell+1}=  \Delta_{\ell+1}\mathbb{E}_{z\sim\mathcal{N}(0,r_\ell)}\left[\sigma_{\ell}(z)^2\right],\notag\\
    &\kappa_1^{(\ell)}=\frac{1}{r_\ell}\mathbb{E}_{z\sim\mathcal{N}(0,r_\ell)}\left[z\sigma_\ell(z)\right],\notag\\
   &\kappa_*^{(\ell) }=\sqrt{\mathbb{E}_{z\sim\mathcal{N}(0,r_\ell)}\left[\sigma_\ell(z)^2\right]-r_\ell\left(\kappa_1^{(\ell)}\right)^2},
\end{align}
with $r_1\equiv\Delta_1\sfrac{\tr\Sigma}{d}$, the deep version of (\ref{prop:BGP_2l}) \camera{and \ref{conj:non-spec}} reads:
\begin{Conjecture}\label{prop:BGP_ML}
    \textit{\textbf{(Deep Bayes GEP)}} in the extensive width asymptotic limit $d\!\rightarrow\!\infty$ and $\forall \ell, k_\ell\!\rightarrow\!\infty$ with fixed ratios $\gamma_\ell\!=\!\sfrac{k_\ell}{d}$, \camera{let $\hat{y}^1(\x),...,\hat{y}^s(\x)$ be any finite number of replicas  independently drawn from the Bayes posterior eq.~\eqref{eq:Bayes_post_classif}. Then }$\hat{y}^1(\x),...,\hat{y}^s(\x)$ are jointly Gaussian  with correlation  $\mathbb{E}_{\x}\hat{y}^a(\x)\hat{y}^b(\x)=\sfrac{{\bf a}^{a\top} \Omega_L^{ab} {\bf a}^b}{k_1}$, where the population covariance $\Omega_L^{ab}\equiv \mathbb{E}_{\x}(\varphi^a_L\circ...\varphi_1^a(\x))(\varphi^b_L\circ...\varphi_1^b(\x))^\top$  is given by 
\begin{align}
\label{eq:Omega_Psi_multilayer}
   \Omega_{\ell}^{ab}= \big(\kappa_1^{(\ell) }\big)^2  \frac{W^a_{\ell} \Omega^{ab}_{\ell-1} W^{b\top}_{\ell}}{k_{\ell-1}}+\delta_{ab} \big(\kappa_*^{(\ell) }\big)^2 \mathbb{I}_{k_{\ell}} \,.
\end{align}
\camera{Finally, defining $\Omega_\ell^{a\star}\equiv \mathbb{E}_{\x}(\varphi^a_\ell\circ...\varphi_1^a(\x))(\varphi^\star_\ell\circ...\varphi_1^\star(\x))^\top$ for any $a$, there is no specialization, i.e. with high probability
\begin{align}
    \frac{1}{d}\underset{1\le i,j \le k_\ell}{\max}\left( W_\ell^a\Omega^{a\star}_{\ell-1}(W_\ell^\star)^\top\right)_{i,j}=\mathcal{O}\left(\sfrac{1}{\sqrt{d}}\right).
\end{align}}
\end{Conjecture}
  In \eqref{eq:Omega_Psi_multilayer}, $\Omega_0^{ab}\equiv\Sigma$. \camera{The derivation of \eqref{eq:Omega_Psi_multilayer} can be found in Appendix\,\ref{App:GET}, eqs \eqref{eq:App:GET:linearization_single_layer} and \eqref{App:eq:propgation_Phi_DRM}}. We precise that \ref{prop:BGP_ML} holds for any sequence of activations $\{\sigma_\ell\}_{\ell=1}^L$ satisfying $\forall\ell,~\mathbb{E}_{z\sim\mathcal{N}(0,r_\ell)}\left[\sigma_\ell(z)\right]=0$. \camera{This is in particular always true for odd activations. We adopt in this work the latter (stronger) assumption for the sake of definiteness and clarity}. An important note is that the population covariance between post-activations at \textit{any} two layers $1\le \ell,\ell^\prime\le L$ (not just $\ell=\ell^\prime=L$) can be generically computed. Because the post-activations result from the propagation of the Gaussian variable $\x$ through several non-linear layers, this computation (which we detail in App.~\ref{App:GET}) is non-trivial for $L\ge 2$ and of independent interest.

\subsection{Bayes-optimal errors}
\label{subsec:BO_error}
\vspace{-2mm}
Conjectures \ref{prop:BGP_2l} and \ref{prop:BGP_ML} are used in equations \eqref{eq:App:2L:local_fields_stats} of App.~\ref{App:2layer} for shallow networks and \eqref{eq:App:Multi:appli_GEP} of App.~\ref{App:multilayer} for deep networks, allowing to characterize the Bayes error in terms of the sole second-order statistic $q=\mathbb{E}_{\mathcal{D}, W_1^\star, {\bf a}_\star}\langle \mathbb{E}_{\x}\hat{y}^a(\x)\hat{y}^b(\x) \rangle_{\mathbb{P}}$. $q$ is known as the \textit{self overlap} in the statistical physics literature. The replica computation then proceeds in a rather standard way, \camera{provided one employs the so-called \textit{replica-symmetry ansatz}, which is always correct in Bayes-optimal settings (see e.g. \cite{Zdeborov2015StatisticalPO})}. One finally reaches the following characterizations in Apps.~\ref{App:2layer} and \ref{App:multilayer}.\\

{\bf For regression}, the Bayes-optimal MSE reads
\begin{align}
\label{eq:Bayes_regression}
\! \! \epsilon_{g,\mathrm{reg}}^{\mathrm{BO}}\!\!=&\! \prod\limits_{\ell=1}^L\big(\kappa_1^{(\ell)}\big)^2\!  \Big(\!\Delta_a\big(\int 
 z\dd \mu(z)\big) \! \prod\limits_{\ell=1}^L \Delta_\ell-q\!\Big) \! +\epsilon_r
\end{align}
with the self-overlap $q$ satisfying the equation
\begin{align}
\label{eq:SP_Bayes}
    q=\frac{1}{2}\int\frac{\alpha\prod\limits_{\ell=1}^L\left(\kappa_1^{(\ell)}\right)^2 z^2\Delta_a^2\prod\limits_{\ell=1}^L \Delta_\ell^2}{\epsilon_{g,\mathrm{reg}}^{\mathrm{BO}}+\alpha\prod\limits_{\ell=1}^L\left(\kappa_1^{(\ell)}\right)^2z\Delta_a\prod\limits_{\ell=1}^L \Delta_\ell}\dd \mu(z). 
\end{align}
We have denoted the residual error
\begin{align}
\label{eq:residual} \! \! \epsilon_r\! \equiv \!\sum\limits_{\ell_0=1}^{L-1}\! \big(\kappa_*^{(\ell_0)}\big)^2\! \Delta_a \!\! \!\!\prod\limits_{\ell=\ell_0+1}^L \! \!\!\!\big(\kappa_1^{(\ell)}\big)^2 \!  \Delta_\ell\! +\! \big(\kappa_*^{(L)}\big)^2 \!  \Delta_a \! + \! \Delta. 
\end{align}

{\color{orange} 
\begin{figure}
    \centering
    \includegraphics[scale=0.57]{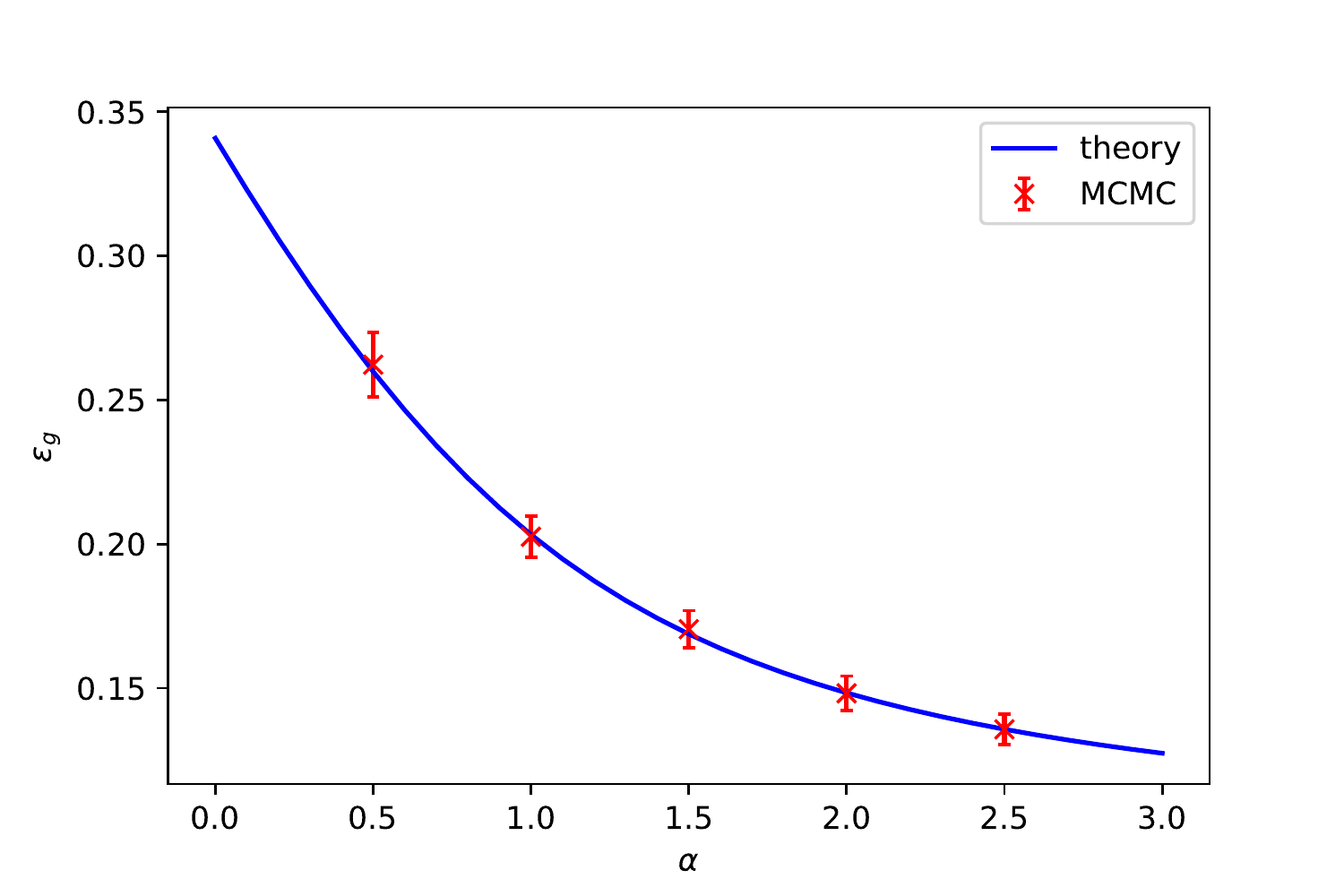}
    \caption{(solid lines) Theoretical predition for the Bayes MSE \eqref{eq:Bayes_regression}, for a one-hidden layer rectangular neural network ($\gamma_1=1$) with shifted ReLU activation $\sigma_1(\cdot)=(\cdot)_+-\sfrac{1}{\sqrt{2\pi}}$. (red crosses) Monte Carlo simulations using the Gibbs sampling algorithm of \cite{piccioli2023gibbs}. The MSE was estimated over the last $15000$ iterations of the algorithm, after $15000$ initial thermalization steps. Error bars represent a single standard deviation over $N=20$ instances of the target network. The simulations were performed in dimension $d=1000$.}
    \label{fig:MCMC}
\end{figure}
}

{\bf For classification,} the Bayes-optimal error reads
\begin{equation}
\label{eq:error_classif}
\epsilon_{g,\mathrm{class}}^{\mathrm{BO}}\!=\! \frac{1}{\pi}\arccos\left[\frac{ \sqrt{\prod\limits_{\ell=1}^L\left(\kappa_1^{(\ell)}\right)^2q}}{{\sqrt{\Delta_a \! \int \!
 z\dd \mu(z) \! \prod\limits_{\ell=1}^L \!\big(\kappa_1^{(\ell)}\big)^2\! \! \Delta_\ell+\epsilon_r}}} \right],
\end{equation}
where the self-overlap $q$ satisfies the system of equations
\begin{align}
\label{eq:classification_BO_SP}
\!
    \begin{cases}
q=\int\frac{{\scriptstyle\hat{q}\Delta_a^2\prod\limits_{\ell=1}^L \Delta_\ell^2z^2}}{\scriptstyle\hat{q}z\Delta_a\prod\limits_{\ell=1}^L \Delta_\ell+1}\dd\mu(z)\\
    \hat{q}=\frac{2\alpha\prod\limits_{\ell=1}^L\left(\kappa_1^{(\ell)}\right)^2}{\Delta_a\int 
 z\dd \mu(z)\prod\limits_{\ell=1}^L \left(\kappa_1^{(\ell)}\right)^2\Delta_\ell+\epsilon_r-\prod\limits_{\ell=1}^L\left(\kappa_1^{(\ell)}\right)^2 q}\notag\\
    ~~ \int \frac{d\xi}{(2\pi)^{\frac{3}{2}}}\frac{2e^{-\frac{1}{2}\frac{\Delta_a\int 
 z\dd \mu(z)\prod\limits_{\ell=1}^L \left(\kappa_1^{(\ell)}\right)^2\Delta_\ell+\epsilon_r+\prod\limits_{\ell=1}^L\left(\kappa_1^{(\ell)}\right)^2 q}{\Delta_a\int 
 z\dd \mu(z)\prod\limits_{\ell=1}^L \left(\kappa_1^{(\ell)}\right)^2\Delta_\ell+\epsilon_r-\prod\limits_{\ell=1}^L\left(\kappa_1^{(\ell)}\right)^2 q}\xi^2}}{1-\mathrm{erf}\left(\frac{\prod\limits_{\ell=1}^L\kappa_1^{(\ell)}\sqrt{q}\xi}{\sqrt{2\left(\Delta_a\int 
 z\dd \mu(z)\prod\limits_{\ell=1}^L \left(\kappa_1^{(\ell)}\right)^2\Delta_\ell+\epsilon_r-\prod\limits_{\ell=1}^L\left(\kappa_1^{(\ell)}\right)^2 q\right)}}\right)}
    \end{cases}.
\end{align}

{\color{black}
As previously discussed, numerically sampling the Bayes posterior \eqref{eq:Bayes_post_classif} is a hard task. However, for regression in the shallow $L=1$ case and a shifted ReLU activation $\sigma_1(\cdot)=(\cdot)_+-\sfrac{1}{\sqrt{2\pi}}$ (with the shift ensuring the condition $\mathbb{E}_{z\sim\mathcal{N}(0,r_1)}\left[\sigma_1(z)\right]=0$ discussed below Conjecture \ref{prop:BGP_ML} is satisfied), the recent work of \cite{piccioli2023gibbs} provides an efficient implementation of a Gibbs sampler. We use this algorithm to run simulations for this particular target network, which can be observed in Fig.\,\ref{fig:MCMC} to agree well with the theoretical prediction \eqref{eq:Bayes_regression}.\\
}

\paragraph{Equivalent shallow network}
Remarkably, the Bayes errors \eqref{eq:Bayes_regression} and \eqref{eq:error_classif} are equal to the Bayes errors of a simple single-layer target function with random weights
\begin{equation}
\label{eq:linear_model}
    y^{\mathrm{eq}}(\x)=f_\star\left(
    \frac{\sqrt{\rho} \bm{\theta}^\top \x}{\sqrt{d}}  +\sqrt{\epsilon_r} \xi
    \right),
\end{equation}
where $\xi\sim\mathcal{N}(0,1)$ and $\bm{\theta}$ is a Gaussian weight vector with independent Gaussian entries of unit variance. We have defined the effective signal strength
$$
\rho\equiv \Delta_a\prod\limits_{\ell=1}^L \left(\kappa_1^{(\ell)}\right)^2\Delta_\ell.
$$
We refer the reader to App.~\ref{App:Shallow} for a derivation of this equivalence. To gain intuition on this equivalence, 
observe that the deep Bayes GEP \ref{prop:BGP_ML} applied to a single replica implies that the deep non-linear network \eqref{eq:teacher} is characterized by the same second-order activations statistics as a network with noisy \textit{linear} layers
\begin{equation}
\label{eq:linearization}
\varphi^{\mathrm{eq.}}_\ell(\x)=\kappa_1^{(\ell)}
    \frac{1}{\sqrt{k_{\ell-1}}}W^\star_\ell \cdot \x+\kappa_*^{(\ell)}\mathcal{N}(0,\mathbb{I}_{k_\ell}).
\end{equation}
In turn, this deep noisy network reduces equivalently to the shallow network \eqref{eq:linear_model}. Interestingly, note that while the multilayer target \eqref{eq:teacher} is deterministic for a given instance of the weights, the equivalent target \eqref{eq:linear_model} displays a stochastic output noise $\xi$. This noise subsumes the effect of the higher order terms introduced by the non-linearities, which are not learnt in the proportional regime \cite{Mei2021GeneralizationEO}.\\

\begin{figure}[t!]
\vspace{-0.9cm}
    \centering 
\includegraphics[scale=0.53]{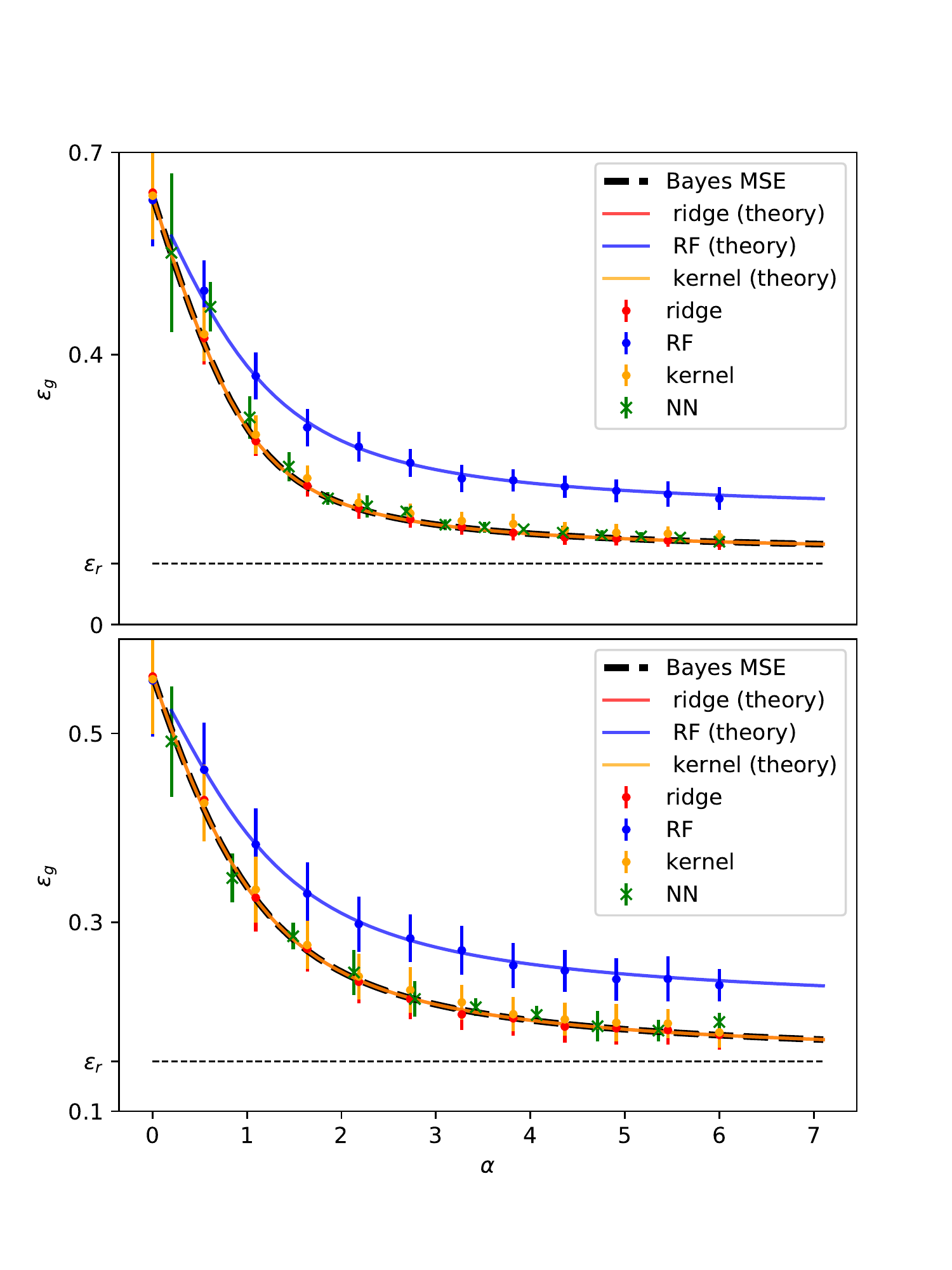}
      \vspace{-1cm}
    \caption{Targets \eqref{eq:teacher} with $L=1$ (top) and $L=2$ (bottom) hidden layers , with $\sigma_{1,2}(x)=\tanh(2x)$ activation, widths $ k_{1,2}=700$, and $\Delta_a=\Delta_{1,2}=1,~\Delta=0$, in dimension $d=500$. The Bayes-optimal MSE \eqref{eq:Bayes_regression} (dashed black) is contrasted to the replica predictions and simulations for optimally regularized ridge regression \eqref{eq:eg_ridge_teacherML} (red), optimally regularized random features \eqref{eq:SP_RF} with $\sigma(x)=\tanh(2x)$ non-linearity (blue), and optimally regularized kernel regression \eqref{eq:SP_Kernel} with 
$\sigma(x)=\tanh(2x)$ non-linearity (orange). Green dots represent simulations for a one (top) and two (bottom) hidden layers neural network of width $1500$, optimized with full-batch GD, learning rate $\eta=8.10^{-3}$ and weight decay $\lambda=0.1$.  Dashed grey lines represent the residual error $\epsilon_r$ \eqref{eq:residual}. \camera{Error bars represent one standard deviation over 30 trials.}}
    \label{fig:regression}
\end{figure}

Fig.~\ref{fig:regression} shows the Bayes MSE, eq.~\eqref{eq:Bayes_regression}, for networks with tanh activation, with $L=1,2$ hidden layers. This is contrasted to the MSE achieved by an expressive neural network (NN) with twice the target width, optimized end-to-end with full batch gradient descent. Fig.\,\ref{fig:classification} presents the same experiment in the classification setting. As expected, even this expressive learning algorithm cannot achieve a lower error than the information-theoretic lower bounds \eqref{eq:Bayes_regression}, \eqref{eq:error_classif}.


\section{ERM with Linear Methods}

\label{sec:ERM_regression}

Eqs.~\eqref{eq:Bayes_regression} and \eqref{eq:error_classif} provide the information-theoretic minimal error for deep extensive-width targets \eqref{eq:teacher}. However, \cite{Barbier2017OptimalEA, Aubin2018TheCM} evidenced that the Bayes error is not always attainable, in practice, by known polynomial time algorithms. In this section, we investigate the performance of some standard ERM methods. We provide a tight asymptotic characterization of the test error of each algorithm and show that for regression the Bayes error is, in fact, also achievable algorithmically. We address in succession: ridge regression, RF regression, kernel regression, logistic regression and ridge classification.\\

\camera{We give, for each of the considered ERM algorithms, a sharp asymptotic characterization of the associated test error.  The fact that the deep non-linear target \eqref{eq:teacher} shares the same Bayes error as the equivalent shallow model \eqref{eq:linear_model} suggests that the test error of ERM methods should also be identical. Applying Theorem $1$ of \cite{Loureiro2021CapturingTL} on the equivalent shallow target \eqref{eq:linear_model} thus leads to the formulas provided here. This heuristic line of reasoning is put on a firm basis in some settings in the concomitant work of \cite{DRF2023}, where the formulas characterizing the performance of the considered ERM methods are derived. We discuss this further in App.\,\ref{App:ERM}.}

\vspace{-2mm}
\subsection{Ridge regression}
\vspace{-2mm}
\label{subsec:ridge}
We first consider ridge regression, corresponding to the minimization of the risk
\begin{equation}
\label{eq:ridge_risk}
    \mathcal{R}(w)=\sum\limits_{\mu=1}^n\left(y^\mu-\frac{w\cdot \x^\mu}{\sqrt{d}}\right)^2+\frac{\lambda}{2}||w||^2,
\end{equation}
with a $\ell_2$ regularization term of strength $\lambda$. The associated test error can be computed by combining the deep GEP \eqref{eq:kappa_multilayer} and the theorem of \cite{Loureiro2021CapturingTL}, as
\begin{align}
\label{eq:eg_ridge_teacherML}
\epsilon_g=&\rho \int 
 z\dd \mu(z)+q-2\prod\limits_{\ell=1}^{L}\kappa_1^{(\ell)} m+\epsilon_r,
\end{align}
where $m,q,V$ are the solutions of the system of equations
\begin{align}
\label{eq:SP_ridge_teacherML}
\!\!\!\!
\begin{cases}
\hat{V}=\frac{\alpha}{1+V}\\
\hat{q}=\alpha\frac{\epsilon_g}{(1+V)^2}\\
\hat{m}=\frac{\prod\limits_{\ell=1}^{L}\kappa_1^{(\ell)}\alpha}{1+V}
\end{cases}\!\!\!\!\! \!\!\!
&&\begin{cases}
V\!=\!\int
\frac{z}{\lambda +\hat{
V}z}\dd\mu(z)
\\
q\!=\!\int
\frac{\Delta_a \prod\limits_{\ell=1}^{L}\Delta_\ell \hat{m}^2z^3+\hat{q}z^2}{(\lambda +\hat{
V}z)^2}\dd\mu(z)
\\
m\!=\!\Delta_a  \!\prod\limits_{\ell=1}^{L} \! \Delta_\ell \hat{m} \! \int \!
\frac{z^2}{\lambda +\hat{
V}z}\dd\mu(z)
\end{cases}
\end{align}
The derivation and further discussion are provided in App.\,\ref{App:ERM}. \\

\begin{figure}[t!]

    \centering
    \includegraphics[scale=0.54]{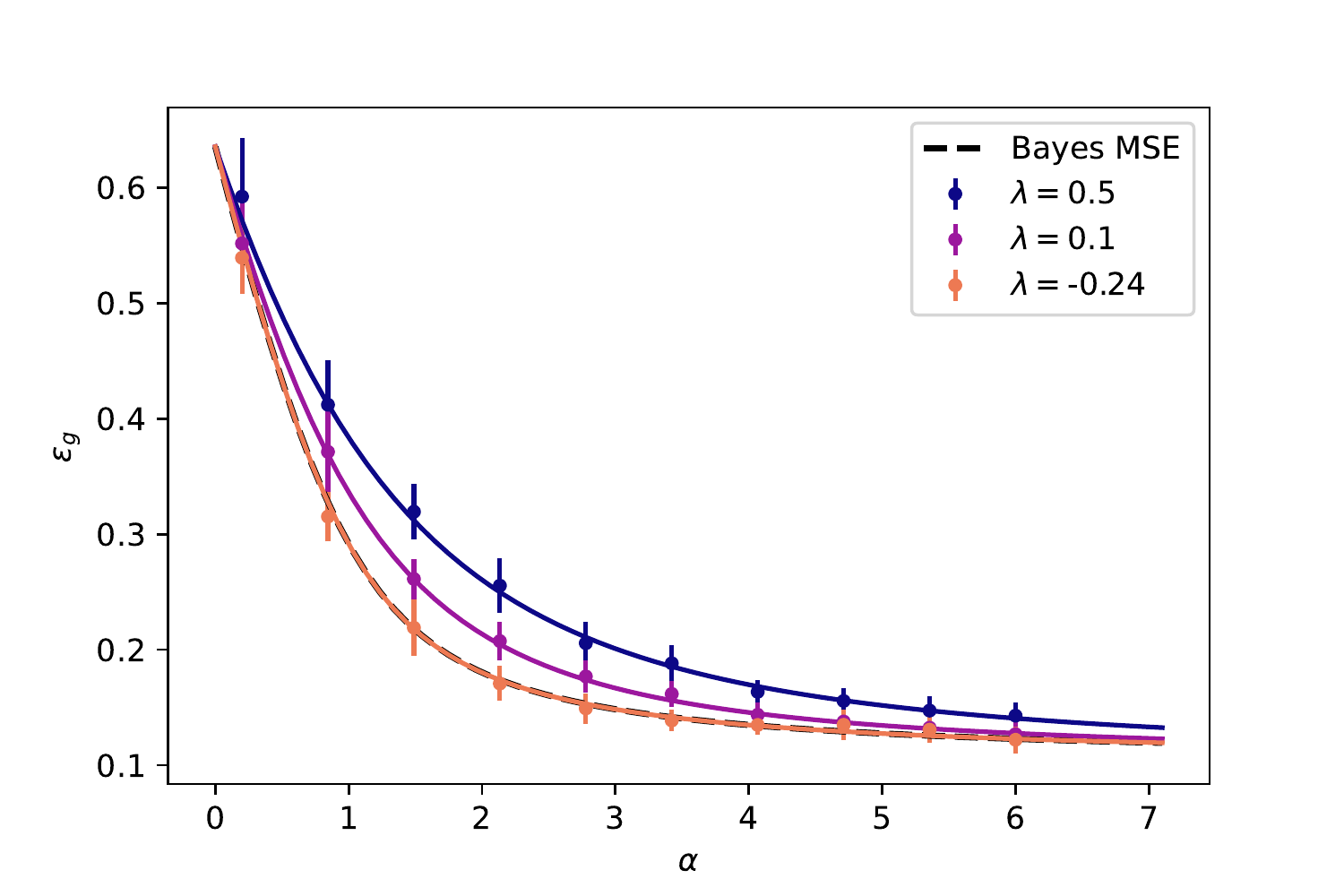}
  
    \caption{Arccosine kernel regression on a two-layers target of width $k_1=700$, $\Delta_a=\Delta_1=1$, $\Delta=0$, with $\sigma_1(x)=\tanh(2x)$ activation, in dimension $d=500$. Different curves correspond to different regularizations $\lambda=0.5, 0.1$ and the optimal $\lambda^\star\approx -0.24$ \eqref{eq:Kernel_opt}.  
    Solid lines correspond to the replica 
 predictions \eqref{eq:SP_Kernel}. The learning curve for the optimal \textit{negative} regularizer $\lambda^\star$ (red) superimposes with the Bayes-optimal MSE \eqref{eq:Bayes_regression}. \camera{Error bars represent one standard deviation over 30 trials.}
}
    \label{fig:neg_reg}
\end{figure}

\paragraph{Ridge regression is Bayes-optimal} The optimal regularization $\lambda^\star$ leading to minimal test error is shown in App.~\ref{App:regu_ERM} to admit the compact expression
\begin{align}
\label{eq:opt_reg_ridge}
    \lambda^\star=\frac{\epsilon_r}{\rho}.
\end{align}
The expression \eqref{eq:opt_reg_ridge} mirrors the result of \cite{SahraeeArdakan2022KernelMA} for GP targets, see App.~\ref{App:ERM}. Eq.~\eqref{eq:opt_reg_ridge} intuitively corresponds to requiring that the regularization $\lambda$ used in the ERM \eqref{eq:ridge_risk} should be equal to the true noise-to-signal ratio of the equivalent target \eqref{eq:linear_model}.\\

For $\lambda=\lambda^\star$, the equation \eqref{eq:SP_ridge_teacherML} reduces to \eqref{eq:Bayes_regression}, implying that optimally regularized ridge regression achieves the Bayes-optimal MSE. The solution of 
\eqref{eq:eg_ridge_teacherML}, \eqref{eq:SP_ridge_teacherML}, \eqref{eq:opt_reg_ridge}, with the corresponding numerical simulations, is plotted in Fig.\,\ref{fig:regression}, and can indeed be seen to exactly fall on the Bayes-optimal baseline \eqref{eq:Bayes_regression}. This implies that in the proportional high-dimensional limit $n\sim d$, no algorithm can learn more accurately the non-linear function \eqref{eq:teacher} than optimally regularized linear regression. This echoes the claims of \cite{SahraeeArdakan2022KernelMA} when the target is a GP.

\vspace{-2mm}
\subsection{Random Features}
\vspace{-2mm}
Random features learning \cite{Rahimi2007RandomFF} was initially introduced as a means to speed up kernel methods. They correspond to the ERM
\begin{equation}
\label{eq:RF_risk}
    \mathcal{R}(w)\!=\!\sum\limits_{\mu=1}^n \! \left(y^\mu-
    \frac{1}{\sqrt{k}}w\cdot\sigma\big(
    \frac{F\cdot \x^\mu}{\sqrt{d}}
    \big)
    \right)^2\!\!+\frac{\lambda}{2}||w||^2,
\end{equation}

where $\sigma$ is a nonlinearity with associated GEP coefficients \eqref{eq:kappa_multilayer} $\kappa_1,\kappa_*$, and $F\in\mathbb{R}^{k\times d}$ is the random feature matrix, assumed in the following to possess i.i.d Gaussian entries. RF learning corresponds formally to ridge regression in a $k-$dimensional space, often taken larger than the $d-$ dimensional input space to allow for overparametrization. Again, we consider the proportional regime $n,d,k\rightarrow\infty$ and introduce the width/dimension ratio $\gamma\equiv k/d$. In the following, for simplicity, we restrict ourselves to the case of isotropic covariates $\Sigma=\mathbb{I}_d$.
Sharp asymptotics for the test error of such models have been provided in \cite{Mei2019TheGE,Gerace2020GeneralisationEI} for single-layer targets. We tie those results in with \eqref{eq:kappa_multilayer} in the case of the deep random target \eqref{eq:teacher}. The asymptotic test error is shown in App.\,\ref{App:ERM} to be given again by \eqref{eq:eg_ridge_teacherML}, where $q,m$ satisfy
\begin{align}
\label{eq:SP_RF}
&\begin{cases}
\hat{V}=\frac{\frac{\alpha}{\gamma}}{1+V}\\
\hat{q}=\frac{\alpha}{\gamma}\frac{\epsilon_g}{(1+V)^2}\\
\hat{m}=\sqrt{\Delta_a \prod\limits_{\ell=1}^{L}\Delta_\ell }\sqrt{\gamma}\frac{\prod\limits_{\ell=1}^{L}\kappa_1^{(\ell)}\frac{\alpha}{\gamma}}{1+V}
\end{cases}
\\
&\begin{cases}
&V=\frac{1}{\hat{V}}
-\frac{\lambda}{\hat{V}^2\kappa_1^2}g\left(
-\frac{\lambda+\hat{V}\kappa_*^2}{\hat{V}\kappa_1^2}
\right)
\\
&q=\frac{\hat{m}^2+\hat{q}}{\hat{V}^2}-\frac{1}{\kappa_1^2\hat{V}^2}\left(\frac{2\lambda(\hat{m}^2+\hat{q})}{\hat{V}}+\hat{m}^2\kappa_*^2
\right)g\left(
-\frac{\lambda+\hat{V}\kappa_*^2}{\hat{V}\kappa_1^2}
\right)\notag\\
&\qquad+\frac{\lambda}{\kappa_1^4\hat{V}^3}\left(\frac{\lambda(\hat{m}^2+\hat{q})}{\hat{V}}+\hat{m}^2\kappa_*^2
\right)g^\prime\left(
-\frac{\lambda+\hat{V}\kappa_*^2}{\hat{V}\kappa_1^2}
\right)
\\
&m=\sqrt{\gamma} \frac{\hat{m}}{\hat{V}}\left[
1-\frac{1}{\kappa_1^2}\left(
\frac{\lambda}{\hat{V}}+\kappa_*^2
\right)g\left(
-\frac{\lambda+\hat{V}\kappa_*^2}{\hat{V}\kappa_1^2}
\right)
\right].
\end{cases}
\end{align}
$g(z)$ is the Stieljes transform of the Marchenko-Pastur distribution with aspect ratio $\gamma$. The solution of \eqref{eq:SP_RF} is plotted in Fig.~\ref{fig:regression}, with the regularization $\lambda$ being numerically optimized over, and is observed to yield higher MSE than the optimal \eqref{eq:Bayes_regression}. Intuitively, this is because from \eqref{eq:linear_model} the target function \eqref{eq:teacher} effectively acts as a linear function on the original space, while the RF transformation $\sigma(F\cdot)$ \eqref{eq:RF_risk} introduces a mismatch between the original basis where the target acts and the features basis in which the linear regression readout is carried out. This intuition is formalized in App.\,\ref{App:regu_ERM}. This mismatch can be shown to be benign only in the infinite overparametrization (infinite width) $\gamma\rightarrow \infty$ limit, has discussed in the next subsection.


\vspace{-2mm}
\subsection{Kernels}
\vspace{-2mm}
In the  $\gamma\rightarrow\infty$ limit, RF learning \eqref{eq:RF_risk} becomes equivalent to kernel regression \cite{Neal1996PriorsFI,Lee2017DeepNN}. The saddle-point equations \eqref{eq:SP_RF} characterizing the generalization error \eqref{eq:eg_ridge_teacherML} then reduce to 
\begin{align}
\label{eq:SP_Kernel}
\!\!\!\!
\begin{cases}
\hat{V}=\frac{\alpha}{1+V}\\
\hat{q}=\alpha\frac{\epsilon_g}{(1+V)^2}\\
\hat{m}=\alpha\frac{\prod\limits_{\ell=1}^{L}\kappa_1^{(\ell)}}{1+V}
\end{cases}
\!\!\!\!
&&\begin{cases}
V&=\frac{\kappa_*^2}{\lambda}+\frac{\kappa_1^2 }{\lambda+\hat{V}\kappa_1^2 }
\\
q&=\frac{\Delta_a \prod\limits_{\ell=1}^{L}\Delta_\ell \hat{m}^2\kappa_1^4+\hat{q}\kappa_1^4}{(\lambda+\hat{V}\kappa_1^2 )^2}
\\
m&=\Delta_a \prod\limits_{\ell=1}^{L^\star}\Delta_\ell \hat{m}
\frac{\kappa_1^2 }{\lambda+\hat{V}\kappa_1^2 }
\end{cases}.
\end{align}
A complete derivation of the eqs \eqref{eq:SP_Kernel} is given in App.\,\ref{App:ERM}. \\

\paragraph{Kernel regression is Bayes-optimal} The regularization $\lambda$ minimizing the test error \eqref{eq:eg_ridge_teacherML} for kernel regression \eqref{eq:SP_Kernel} can be shown as in the ridge case to admit the simple expression
\begin{equation}
\label{eq:Kernel_opt}
    \lambda^\star=\kappa_1^2\left(
    \frac{\epsilon_r}{\rho}-\frac{\kappa_*^2}{\kappa_1^2}
    \right).
\end{equation}
We discuss the connection of this formula with the one provided in \cite{SahraeeArdakan2022KernelMA} for GP targets in App.~\ref{App:2layer}. Again, the expression \eqref{eq:Kernel_opt} also admits a very intuitive interpretation. An informal takeaway from \cite{Mei2021GeneralizationEO, Hu2022SharpAO} is indeed that in the linear $n\sim d$ regime, kernel regression can be seen as effectively implementing ridge regression with an implicit $\ell_2$ regularization equal to $\sfrac{\lambda+\kappa_*^2}{\kappa_1^2}$. Requiring this implicit regularization to match the true target noise-to-signal ratio \eqref{eq:linear_model} naturally leads to \eqref{eq:Kernel_opt}. The test error \eqref{eq:eg_ridge_teacherML} evaluated at the solution of \eqref{eq:SP_Kernel}, at the optimal regularization \eqref{eq:Kernel_opt}, is plotted in Fig.\,\ref{fig:regression}, and can be seen to exactly superimpose with the Bayes-optimal baseline \eqref{eq:Bayes_regression}. \\

 When the activation $\sigma$ of the kernel is very non-linear (as quantified by the ratio $\sfrac{\kappa_*^2}{\kappa_1^2}$) and implements too strong an implicit regularization \cite{Wu2020OnTO,Hastie2019SurprisesIH} compared to the actual target noise, the optimal regularization \eqref{eq:Kernel_opt} can be \textit{negative}. This negative ridge phenomenon can for example be observed when learning a target $2-$layer network with $\tanh$ activation with a kernel using the more non-linear $\mathrm{sign}$ activation, see Fig.~\ref{fig:neg_reg}.  Note that even in such cases where the optimal explicit regularization $\lambda$ is negative, the risk remains convex due to the implicit $\ell_2$ regularization.

\begin{figure}[t!]

    \centering
\includegraphics[scale=0.55]{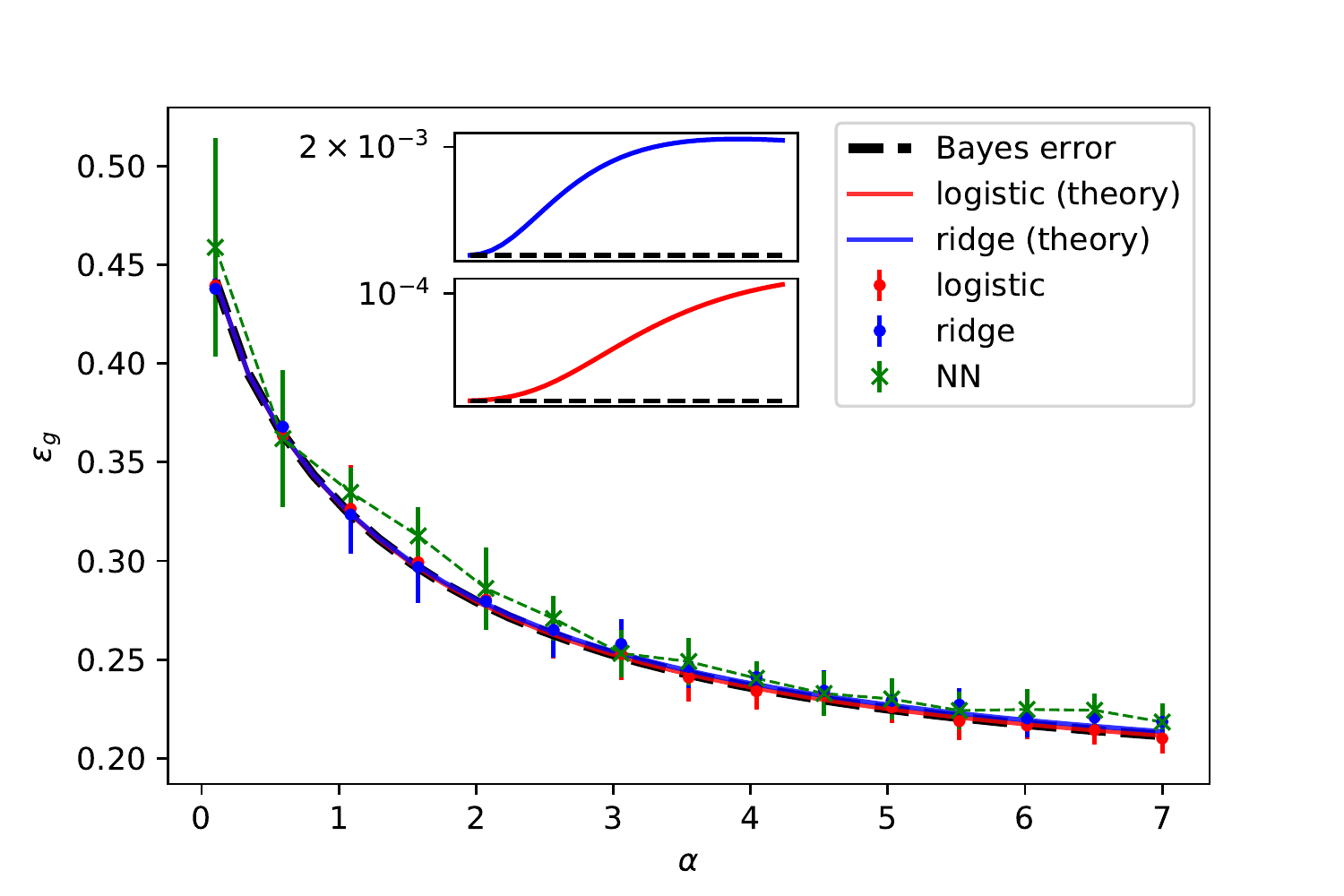}

    \caption{Learning curves for classification, with a three-layers target \eqref{eq:teacher} with $\mathrm{tanh}(2\cdot)$ activation and width $k_{1,2}=700$, $\Delta_a=\Delta_{1,2}=1$, $\Delta=0$, in dimension $d=500$. The black dashed line represents the Bayes-optimal error \eqref{eq:error_classif}. The theoretical learning curves for optimally regularized logistic regression (ridge classification) are shown in red (blue), alongside the corresponding numerical simulations. Green dots show the test error of a three layers fully connected network trained end-to-end with full-batch Adam, learning rate $0.003$ and weight decay $0.01$, after $2000$ epochs. \camera{Error bars represent one standard deviation over 30 trials.} (inset) Zoom in on the theoretical learning curves for logistic regression (red) and ridge classification (blue), with the Bayes-optimal baseline \eqref{eq:error_classif} subtracted. Logistic regression and ridge classification can be seen to be very close, but not equal, to the Bayes-optimal baseline (up to $10^{-4}$ and $10^{-3}$ respectively).
   }
    \label{fig:classification}
\end{figure}

\vspace{-2mm}
\subsection{Logistic and ridge classification}
\vspace{-2mm}
 This last subsection addresses ERM in the classification setting. We consider two standard classification methods, namely logistic regression and ridge classification. They correspond to ERM on the risk
\begin{equation}
\label{eq:classification_risk}
    \mathcal{R}(w)=\sum\limits_{\mu=1}^n \ell \left(y^\mu,\frac{w\cdot \x^\mu}{\sqrt{d}}\right)+\frac{\lambda}{2}||w||^2
\end{equation}
with $\ell(y,z)=\ln(1+e^{-yz})$ and $\ell(y,z)=\sfrac{1}{2}(y-z)^2$ respectively. For simplicity, we assume the noiseless $\Delta=0$ setting. An asymptotic expression for the test error of logistic regression can again be reached and is detailed in App.\,\ref{App:Classification}. These theoretical predictions are plotted in Fig.~\ref{fig:classification}, and contrasted to the Bayes-optimal baseline \eqref{eq:error_classif}. At odds with the regression case, the learning curves of logistic regression and ridge classification do not exactly superimpose with the Bayes-optimal baseline but lie extremely close.  Fig.\,\ref{fig:classification} shows that for a $\sigma(x)=\tanh(2x)$ activation the difference is of order of $10^{-4}$ (for logistic regression) and $10^{-3}$ (for ridge classification). \camera{Such a gap has also been observed by \cite{Aubin2020GeneralizationEI} for targets without hidden layer.} 

\begin{figure}[t!]

    \centering
    \includegraphics[scale=0.6]{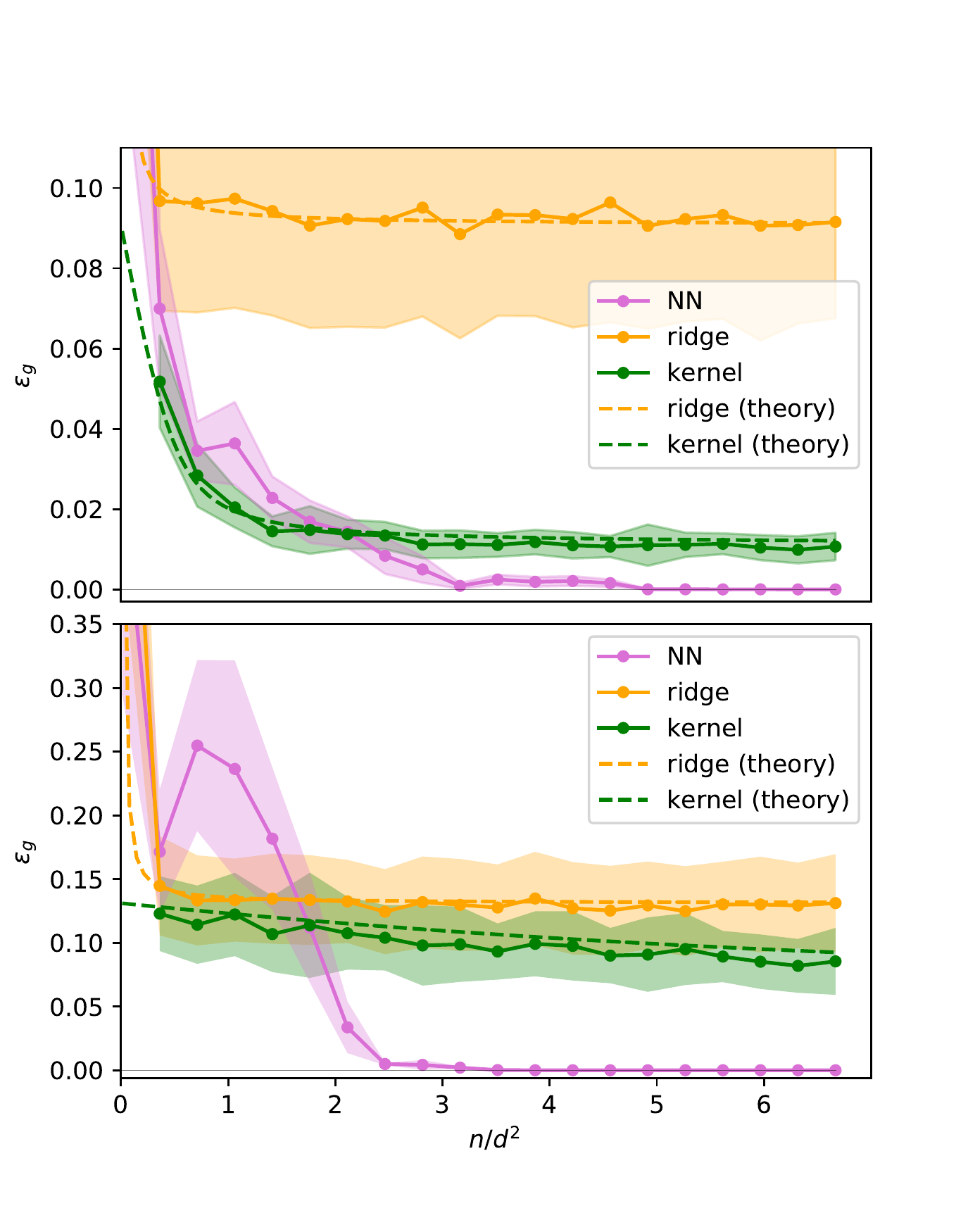}
        
    \caption{MSE of regression on a ReLU (top) erf$(2x)$ (bottom) $2-$layers target of width $k_1=20$, $\Delta_a=\Delta_1=1,~\Delta=0$, in dimension $d=30$. Dots represent simulations for optimally regularized ridge regression (orange), optimally regularized arccosine (top), and arcsine (bottom) kernel (green). Dashed lines represent the theoretical predictions for the MSE of the kernel in polynomial regimes of \cite{Hu2020UniversalityLF}. Purple dots indicate the MSE of a $2-$layers fully connected neural network of width $k=30$ trained end-to-end using Adam (purple), batch size $\sfrac{n}{3}$ and learning rate $\eta=3.10^{-3}$, over $2000$ epochs. \camera{Error bars represent one standard deviation over 10 trials.} For a quadratic number of samples $n\sim d^2$ the network learns the target perfectly (up to errors of order $10^{-5}$) while kernel regression only learns a quadratic approximation of the target and yields MSEs larger by an order of $10^3$. Ridge regression can only learn the best linear approximation and leads to even higher MSE.
    }
    \label{fig:Quad}
\end{figure}

\section{Beyond the Proportional Regime}

\label{sec:quadratic}
Section \ref{sec:ERM_regression} establishes that ridge regression and kernel regression are Bayes-optimal in the linear $n\sim d$ regime for the target \eqref{eq:teacher}.
These conclusions and the ones of \cite{SahraeeArdakan2022KernelMA} in the context of GP targets, and are reminiscent on a high level of the empirical observations of \cite{Arora2019HarnessingTP} that neural networks can only marginally outperform kernel methods for small train sets on several benchmark real datasets. Note that the study \cite{Lee2020FiniteVI} also employs networks with comparable width to the dataset size \& dimension, and reaches similar conclusions.\\

In fact, the information-theoretic optimality of linear/kernel ERM methods is essentially due to the fact that a proportional number of samples $n\sim d$ is not enough to learn features beyond linear approximations. The conclusions drawn in other scaling regimes are anticipated to be very different. 
In particular, beyond the linear $n\sim d$ regime, the test error should pick up, and depend on, non-Gaussian asymptotic corrections to the GEP \cite{Goldt2021TheGE}, Bayes GEP \ref{prop:BGP_2l}, and deep Bayes GEP \ref{prop:BGP_ML}.
Besides, information-theoretic intuition suggests that since the target \eqref{eq:teacher} is parametrized by $\mathcal{O}(d^2)$ real numbers, a quadratic number of samples $n\sim d^2$ is needed, and sufficient, to learn it perfectly. In other words, one expects the Bayes-optimal error to vanish in the quadratic regime. This means in particular that kernel methods (which are known to learn at best a quadratic approximation of the target in this scaling regime \cite{Misiakiewicz2022SpectrumOI,xiao2022precise,Hu2022SharpAO}), and ridge regression (which learns a linear approximation) will cease to be optimal.
In this section, this intuition is further explored numerically.\\

Fig.\,\ref{fig:Quad} contrasts the MSE of an Adam-optimized neural network, optimally regularized ridge regression, and optimally regularized arcosine kernel regression, in the $n\sim d^2$ regime, for a ReLU network target with one hidden layer. For completeness, we also plot the theoretical predictions for the test error for kernel regression derived in \cite{Hu2022SharpAO}. While ridge (kernel) regression can only learn the best linear (quadratic) approximation of the non-linear target, the neural network manages to learn the target almost perfectly (up to an MSE of $\mathcal{O}(10^{-5})$). \camera{These results show the superiority of neural networks over kernel methods in the quadratic regime for the multi-layer target \eqref{eq:teacher}, and complement similar superiority results (see e.g. \cite{Ghorbani2019LinearizedTN,Ghorbani2020WhenDN}) for single-layer targets.}


\vspace{-2mm}
\section*{Conclusion}
\vspace{-2mm}
We investigate the problem of learning a deep, non-linear, extensive-width random neural network. We propose asymptotic expressions for the Bayes-optimal error and the test error of linear / kernel ERM methods, for classification and regression. The technical backbone of the derivations is the deep Bayes GEP conjecture \ref{prop:BGP_ML}, and novel closed-form formulae for second-order population statistics of network post-activations. The  conclusion is that kernel methods are optimal in this regime. We showed, however, that the situation is drastically different in the quadratic sample regime, and evidence the onset of feature learning leading to a vanishing test error for neural nets, while kernel methods can learn only quadratic approximation and thus become suboptimal. This marks a clear separation in the power of neural nets with respect to kernels as soon as $n\sim d^2$.\\

From a theoretical standpoint, a quantitative analysis of the learning curve for the quadratic regime is challenging. 
In particular, Gaussian universality results such as \cite{Goldt2021TheGE,Hu2020UniversalityLF,Mei2019TheGE}, 
cease to hold outside of the proportional regime and would need to be extended. Building a theory for extensive-width networks in these superlinear sample regimes could unveil rich behavior and properties, and constitutes in the authors's point of view a crucial challenge in machine learning theory.

\section*{Acknowledgements}
We thank Haim Sompolinsky for his wonderful lecture in Les Houches Summer School July 2022 that inspired parts of this work. 
We also thank Emanuele Troiani, Yatin Dandi, Sebastian Goldt, Antoine Maillard, Marylou Gabri\'e and Bruno Loureiro for insightful discussions at the early stages of this project, which originated at the Les Houches workshop on Statistical Physics and Machine Learning in 2020. We acknowledge funding from the ERC under the European Union’s Horizon 2020 Research and Innovation Program Grant Agreement 714608-SMiLe as well as from by the Swiss National
Science Foundation grant SNFS OperaGOST, $200021\_200390$.

\newpage
\hfill

\bibliography{biblio}
\bibliographystyle{icml2022}

\newpage
\onecolumngrid
\appendix

\section{Deep Gaussian Equivalence Principles}
\label{App:GET}
\input{Appendix/GET.tex}

\section{Replica computation for shallow networks}
\label{App:2layer}
\input{Appendix/2layers.tex}

\section{Replica computation for deep networks}
\label{App:multilayer}
\input{Appendix/Multilayers.tex}

\section{Equivalent shallow network}
\label{App:Shallow}
\input{Appendix/Shallow.tex}

\section{ERM: the shallow case}
\label{App:ERM}
\input{Appendix/ERM_G3M.tex}

\section{Optimality of kernel ERM on shallow targets}
\label{App:regu_ERM}
\input{Appendix/ERM_regu.tex}

\section{ERM: the deep case}
\label{App:Multi_ERM}
\input{Appendix/Multi_ERM.tex}

\section{Classification}
\label{App:Classification}
\input{Appendix/Classification.tex}

\end{document}

%% file: Appendix/GET.tex
In this Appendix, we provide a discussion of the deep GEP \ref{prop:BGP_ML}. We first provide a derivation for the closed form expressions of the population covariances of the network activations, from which \eqref{eq:Omega_Psi_multilayer} can be deduced as a particular case. We then provide strong numerical support for the 1dCLT \ref{prop:BGP_2l}.

\subsection{Setting}
In this section, we provide a reminder of the notation. Since the matrix identities are applied also in section \ref{sec:ERM_regression}, where the learner network does not necessarily possess the same architecture as the target \eqref{eq:teacher}, we generically study the correlations between any pair of activations of \textit{two} networks. For clarity, in this appendix, we will denote all target hyperparameters by a star $\star$, to distinguish them with the parameters of the learner.

We consider a $L^\star$ layers target (teacher) function
\begin{align}
    y(x)=\frac{1}{\sqrt{k^\star_{L^\star}}}a_\star^\top  \underbrace{\left(
    \varphi^\star_{L^\star}\circ \varphi^\star_{L^\star-1}\circ \dots\circ \varphi^\star_2\circ\varphi_1\right)}_{L^\star}
    (x) 
\end{align}
where the $\ell$-th layer is
\begin{equation}
    \varphi^\star_\ell(x)=\sigma^\star_\ell\left(
    \frac{1}{\sqrt{k^\star_{\ell-1}}}W^\star_\ell \cdot x
    \right)
\end{equation}
$(\sigma^\star_\ell)_{\ell=1}^{L^\star}$ is a sequence of \textit{\textbf{odd}} activation functions. The width of the $\ell$-th layer is denoted $k_\ell$, and the associated weight matrix is $W^\star_\ell\in\mathbb{R}^{k_\ell\times k_{\ell-1}}$. In contrast to the main text, for clarity, we use a star $\star$ superscript for all the parameters related to the target function. Unstarred variables shall conversely refer to the learner (hereafter referred to as the \textit{student}) network.

The student network is also taken to be a random network, with trainable last layer:

\begin{align}
    y(x)=\frac{1}{\sqrt{k_{L}}}a^\top  \underbrace{\left(
    \varphi_{L}\circ \varphi_{L-1}\circ \dots\circ \varphi_2\circ\varphi_1\right)}_{L}
    (x) ,
\end{align}
with layers
\begin{equation}
    \varphi_\ell(x)=\sigma_\ell\left(
    \frac{1}{\sqrt{k_{\ell-1}}}W_\ell \cdot x
    \right).
\end{equation}
In the main text, we consider a student network with the same architecture as the target in section \ref{sec:Bayes_regression} ($L=L^\star$ and $\forall \ell,~~k_\ell=k^\star_\ell$), where the Bayes errors are derived. In section \ref{sec:ERM_regression} on the other hand, the student shall correspond to a linear single layer ($L=0$) network (for ridge regression, logistic regression and ridge classification) or a one-hidden-layer ($L=1$) network with frozen Gaussian first layer (for random features).
We denote the $\ell$-th layer post-activation $h_\ell(x)=(\varphi_\ell\circ...\circ\varphi_1)(x)$ (for student) $h^\star_\ell(x)=(\varphi^\star_\ell\circ...\circ\varphi^\star_1)(x)$ (for the teacher), and introduce the covariances 
\begin{align}
\label{eq:App:GET:cov_def}
    \Omega_{\ell}=\langle h_\ell(x)h_\ell^\top(x)\rangle_x, && \Psi_{\ell}=\langle h^\star_\ell(x)h_\ell^{\star\top}(x)\rangle_x
    &&
    \Phi_{\ell^\star \ell}=\langle h^\star_{\ell^\star}(x)h_\ell^\top(x)\rangle_x,
\end{align}
which we shall characterize for any $\ell$. Note that we could also compute correlations of the form $\langle h_\ell(x){h_{{\ell^\prime}}}^\top(x)\rangle_x$,$\langle h^\star_\ell(x){h^\star_{{\ell^\prime}}}^\top(x)\rangle_x$ for generic $\ell\ne \ell^\prime$, i.e. population covariances between activations of the same network, but at different layers. Since these matrices however never enter our main discussion, and are furthermore a straightforward extension away from the results presented here, we choose to skip the discussion thereof for clarity purposes.

We further suppose the data $x$ is i.i.d Gaussian
\begin{equation}
    x\sim \mathcal{N}(0,\Sigma),
\end{equation}
with a covariance $\Sigma$ of extensive Frobenius norm and trace, i.e. there exists constant $c,c^\prime$ so that asymptotically (noting $k_0=d$)
\begin{align}
    c<\frac{1}{d}\tr \Sigma^2=\frac{1}{d}||\Sigma||_F^2<c^\prime<\infty, && c<\frac{1}{d}\tr \Sigma<c^\prime<\infty.
\end{align}
In terms of the limiting spectral density $\mu$, these assumptions imply that the first and second moments are finite and non zero.
Finally, all weights are assumed to be Gaussian:
\begin{align}
    W_\ell^\star\overset{i.i.d.}{\sim}\mathcal{N}(0,\Delta_\ell^\star),
    &&
    W_\ell\overset{i.i.d.}{\sim}\mathcal{N}(0,\Delta_\ell).
\end{align}
Note that this last assumption is made for simplicity and definiteness. An exhaustive study of the most generic distribution for the weights under which \eqref{eq:Omega_Psi_multilayer} holds is a research line of independent interest, and falls out of the scope of the present manuscript.

\subsection{Closed-form formulas for second order population statistics}

We now re-state the recursion relation allowing to compute the covariances \eqref{eq:App:GET:cov_def}. Consider the sequence of variances defined by the recurrence
\begin{align}
\label{eq:App:GET:r_multilayer}
r^{(\star)}_{\ell+1}=\Delta_{\ell+1}^{(\star)}\mathbb{E}_z^{\mathcal{N}(0,r_\ell)}\left[\sigma_\ell^{(\star)}(z)^2\right],
\end{align}
with the initial condition
\begin{align}
    r_1^{(\star)}=\Delta_1^{(\star)}\frac{1}{d}\tr \Sigma.
\end{align}
$r_\ell^{(\star)}$ physically corresponds to the average squared norm of the post-activations after layer $\ell$. Then introduce the coefficients 
\begin{align}
\label{eq:App:GET:kappa_multilayer}
    \kappa_1^{\ell(\star)}=\frac{1}{r_\ell^{(\star)}}\mathbb{E}_z^{\mathcal{N}(0,r_\ell)}\left[z\sigma_\ell^{(\star)}(z)\right],
    &&\kappa_*^{\ell (\star)}=\sqrt{\mathbb{E}_z^{\mathcal{N}(0,r_\ell)}\left[\sigma_\ell^{(\star)}(z)^2\right]-r_\ell^{(\star)}\left(\kappa_1^{\ell(\star)}\right)^2}.
\end{align}
For the first layer ($\ell=1$), these coefficient correspond to the coefficients introduced by the seminal works of \cite{Pennington2019NonlinearRM,Louart2017ARM}, sometimes referred to as GET coefficients \cite{Goldt2020ModellingTI, Gerace2020GeneralisationEI}. \camera{These coefficients also appear in the characterization of the spectrum of the conjugate kernel of random multi-layer networks in \cite{Fan2020SpectraOT}.}
The layer-wise covariances $\Omega, \Psi$ \eqref{eq:App:GET:cov_def} are then defined by the following recursion
\begin{align}
\label{eq:App:GET:Omega_Psi_multilayer}
    \Omega_{\ell+1}=\kappa_1^{\ell 2}\frac{W_{\ell+1} \Omega_\ell W_{\ell+1}^\top}{k_\ell}+\kappa_*^{\ell 2}I_{k_{\ell+1}},\\
    \Psi_{\ell+1}=\kappa_1^{\star\ell 2}\frac{W^\star_{\ell+1} \Psi_\ell W_{\ell+1}^{\star\top}}{k^\star_\ell}+\kappa_*^{\star\ell 2}I_{k^\star_{\ell+1}},
\end{align}
with initialization
\begin{align}
    \Omega_0=\Psi_0=\Sigma.
\end{align}
The teacher/teacher student correlation $\Phi$ admits the closed form expression
\begin{align}
\label{eq:App:GET:Phi_multilayer}
    \Phi_{\ell^\star \ell}=\prod\limits_{r=1}^\ell\prod\limits_{s=1}^{\ell^\star}\kappa_1^r\kappa_1^{s\star}
    \times
    \frac{W_{\ell^\star}\cdot...\cdot W_1^\star\cdot \Sigma \cdot W_1^\top \cdot ... \cdot W_\ell^\top}{\prod\limits_{r=0}^{\ell-1}\prod\limits_{s=0}^{\ell^\star-1}\sqrt{k_rk^\star_s}}.
\end{align}

\paragraph{Example for $L=2$}
For definiteness, let us provide an example for $L^\star=1,  L=2$ (one hidden layer teacher, two hidden layers student). The recursions \eqref{eq:App:GET:Omega_Psi_multilayer}\eqref{eq:App:GET:Phi_multilayer} then translate to the compact formulas
\begin{align}
\label{eq:App:GET:two_layer_covariance}
    &\Omega_2=(\kappa_1^1)^2(\kappa_1^2)^2\frac{W_2W_1\Sigma W_1^\top W_2^\top}{k_1d}+(\kappa_1^2)^2(\kappa_*^1)^2\frac{W_2W_2^\top}{k_1}+(\kappa_*^2)^2I_{k_1},\\
    &\Psi_1=(\kappa_1^{1\star})^2\frac{W_1^\star\Sigma W_1^{\star \top}}{d}+(\kappa_*^{1\star})^2I_{k^\star_1},\\
    &\Phi_{1,2}=\kappa_1^1\kappa_1^2\kappa_1^{1\star}\frac{W_1^\star\Sigma W_2^\top W_1^\top}{d\sqrt{k_1}}.
\end{align}

\subsection{Derivation for $\Omega, \Psi$}
In this subsection, we detail the derivation of the recursion \eqref{eq:Omega_Psi_multilayer}. We first derive a preliminary results, describing the evolution of the data covariance through propagation of a single layer, and subsequently iterate thereupon to reach \eqref{eq:Omega_Psi_multilayer}. Note that because the student and teacher share the same architecture and model definition, the treatment of the teacher/teacher covariance $\Psi$ and the student/student covariance $\Omega$ is identical. In the following, we therefore only discuss the student/student covariance $\Omega$.

\paragraph{Preliminary: propagating through one layer}
Consider the auxiliary single-layer problem
\begin{equation}
    h(x)=\sigma\left(\frac{1}{ \sqrt{d}}W\cdot x\right)
\end{equation}
with $x\sim\mathcal{N}(0,\Sigma)$. The covariance of the post-activation $h(x)$ reads
\begin{align}
    \Omega_{ij}=\langle h_i(x)h_j(x)\rangle_x=\int \frac{e^{-\frac{1}{2}\begin{pmatrix}
    u&v
    \end{pmatrix}
    \begin{pmatrix}
    \frac{w_i^\top \Sigma w_i}{d} & \frac{w_i^\top \Sigma w_j}{d}\\
    \frac{w_i^\top \Sigma w_j}{d}& \frac{w_j^\top \Sigma w_j}{d}
    \end{pmatrix}^{-1}
    \begin{pmatrix}
    u\\v
    \end{pmatrix}
    }}{\sqrt{\det 2\pi \begin{pmatrix}
    \frac{w_i^\top \Sigma w_i}{d} & \frac{w_i^\top \Sigma w_j}{d}\\
    \frac{w_i^\top \Sigma w_j}{d}& \frac{w_j^\top \Sigma w_j}{d}
    \end{pmatrix}}}\sigma(u)\sigma(v).
\end{align}
Note that the random variable $\frac{w^\top \Sigma w}{d}$ concentrates, as $w$ is i.i.d Gaussian of variance $\Delta$. We indeed have that
\begin{equation}
\label{eq:App:GET:r_singlelayer}
    \mathbb{E}_w \frac{w^\top \Sigma w}{d}=\frac{\Delta}{d}\tr \Sigma \equiv r
\end{equation}
and (diagonalizing $\Sigma=U\Lambda U^\top$) and noting that $U^\top w$ is still Gaussian with independent entries
\begin{align}
    \mathbb{V}_w\left[\frac{w^\top \Sigma w}{d}\right]=\frac{1}{d^2}\sum\limits_{i=1}^d \lambda_i^2 \mathbb{V}_w\left[(U^\top w)_i^2\right]=\frac{2\Delta}{d^2}\tr \Sigma^2=\frac{2\Delta}{d}\frac{||\Sigma||_F^2}{d}=\mathcal{O}\left(\frac{1}{d}\right)
\end{align}
provided $\sfrac{||\Sigma||_F^2}{d}$ is finite. We used the fact that the variance of a $1-$degree of freedom $\chi^2$ variable is $2$. Plugging the definition of $r$ into the above yields, for an off-diagonal entry $i\ne j$:
\begin{align}
    \Omega_{ij}&=\int \frac{e^{-\frac{1}{2}\frac{1}{r^2-\mathcal{O}\left(\frac{1}{d}\right)}(r u^2+rv^2)}e^{\frac{1}{r^2-\mathcal{O}\left(\frac{1}{d}\right)}\frac{w_i^\top\Sigma w_j}{d}uv}}{2\pi\sqrt{r^2-\mathcal{O}\left(\frac{1}{d}\right)}}\sigma(u)\sigma(v)\notag\\
    &=\left(\int \frac{e^{-\frac{1}{2r}z^2}}{\sqrt{2\pi r}}\sigma(z)\right)^2+\frac{1}{r}
    \frac{w_i^\top \Sigma w_j}{d}
    \left(\int \frac{e^{-\frac{1}{2r}z^2}}{\sqrt{2\pi r}}z\sigma(z)\right)^2+\mathcal{O}\left(\frac{1}{d}\right)\notag\\
    &=\kappa_1^2\times \frac{w_i^\top \Sigma w_j}{d}.
\end{align}

On the diagonal,
\begin{align}
    \Omega_{ii}=\int \frac{e^{-\frac{1}{2r}z^2}}{\sqrt{2\pi r}}\sigma(z)^2=\kappa_*^2+r\kappa_1^2.
\end{align}
Therefore, 
\begin{align}
\label{eq:App:GET:linearization_single_layer}
    \Omega=\kappa_1^2\frac{W\Sigma W^\top}{d}+\kappa_*^2I_k,
\end{align}
where
\begin{align}
\label{eq:App:GET:kappa_single_layer}
    \kappa_1=\frac{1}{r}\mathbb{E}_{z}^{\mathcal{N}(0,r)}\left[z\sigma(z)\right]&&\kappa_*^2=\mathbb{E}_{z}^{\mathcal{N}(0,r)}\left[\sigma(z)^2\right]-r\times \kappa_1^2
\end{align}
This extends the single-layer GET \cite{Gerace2020GeneralisationEI} to arbitrary input covariances. A similar generalization was reported in \cite{dAscoli2021OnTI} in the special case of $\Sigma$ which spectral density is equal to a finite sum of Dirac atoms.

\paragraph{Iterating layer to layer}
\eqref{eq:kappa_multilayer} and \eqref{eq:Omega_Psi_multilayer} follow by straightforward recursion from the single-layer results \eqref{eq:App:GET:kappa_single_layer} and \eqref{eq:App:GET:linearization_single_layer}, by making the strong assumption \textit{that all post-activations can be treated as Gaussian variables}.  \camera{This assumption is supported, at a physics level of rigor, by the work of \cite{Fischer2022DecomposingNN}, which establishes that cumulants of order $\ge 4$ are asymptotically suppressed at least as $\sfrac{1}{k_\ell^2}$}. One then just needs to connect \eqref{eq:App:GET:r_multilayer} to the single-layer variance $r$ \eqref{eq:App:GET:r_singlelayer}. 
\begin{align}
    r_{\ell+1}&=\Delta_{\ell}\frac{1}{k_\ell}\tr \Omega_\ell\notag\\
    &=\Delta_{\ell+1}\left(
    \frac{1}{k_\ell}\left(\kappa_1^{\ell}\right)^2\tr[\frac{W_\ell\Omega_{\ell-1} W_\ell^\top}{k_{\ell-1}}]+\left(\kappa_*^{\ell}\right)^2
    \right)\notag\\
    &=\Delta_{\ell}\left(
    \left(\kappa_1^{\ell}\right)^2r_\ell+\left(\kappa_*^{\ell}\right)^2
    \right)\notag\\
    &=\Delta_{\ell}\mathbb{E}_{z}^{\mathcal{N}(0,r_\ell)}\left[\sigma(z)^2\right].
\end{align}
We used
\begin{align}
    \frac{1}{k_\ell}\tr[\frac{W_\ell\Omega_{\ell-1} W_\ell^\top}{k_{\ell-1}}]&=\frac{1}{k_{\ell-1}}\sum\limits_{i=1}^{k_{\ell-1}}\lambda_i^{\ell-1}\frac{1}{k_\ell}\left(U^\top W_\ell^\top W_\ell U\right)_{ii}\notag\\
    &=\frac{1}{k_{\ell-1}}\sum\limits_{i=1}^{k_{\ell-1}}\lambda_i^{\ell-1}\Delta_\ell
\notag\\
&=\Delta_\ell\frac{1}{k_{\ell-1}}\tr\Omega_{\ell-1}=r_\ell
\end{align}
Finally, one must check that the assumption on $\Sigma$ that $\sfrac{||\Sigma||_F^2}{d},\sfrac{\tr\Sigma}{d}=\mathcal{O}(1)$ carries over to $\Omega$. Because $W\Sigma W^\top$ is positive semi definite it is straightforward that 
\begin{equation}
    \frac{1}{k}\left\lVert\kappa_1^2\frac{W\Sigma W^\top}{d}+\kappa_*^2I_k\right\lVert_F^2 \ge \kappa_*^2 >0.
\end{equation}
The upper bound can be established using the triangle inequality and the submultiplicativity of the Frobenius norm, as
\begin{align}
    \frac{1}{k}\left\lVert\kappa_1^2\frac{W\Sigma W^\top}{d}+\kappa_*^2I_k\right\lVert_F^2 &\le \frac{1}{k}\left\lVert\kappa_1^2\frac{W\Sigma W^\top}{d}\right\lVert_F^2+\kappa_*^2\notag\\
    &\le \kappa_*^2+\frac{||W||_F^4}{d^2}\frac{||\Sigma||_F^2}{k}\notag \\
    &\le \kappa_*^2 + c^\prime<\infty.
\end{align}
We used that $\sfrac{||W||_F^2}{dk}=1$ almost surely asymptotically. Moving on to the trace,
\begin{equation}
    \frac{1}{k}\Tr[\kappa_1^2\frac{W\Sigma W^\top}{d}+\kappa_*^2I_k]=\kappa_*^2+\frac{\kappa_1^2}{kd}\Tr[\Sigma W^\top W].
\end{equation}
Bounding
\begin{equation}
    0\le \frac{\kappa_1^2}{kd}\Tr[\Sigma W^\top W]=\frac{\kappa_1^2}{kd}\sum\limits_{i=1}^k w_i^\top \Sigma w_i=\kappa_1^2\frac{1}{d}\Tr{\Sigma}\le \kappa_1^2 c^\prime,
\end{equation}
where the last bound holds asymptotically almost surely. As discussed in the main text, this derivation carries over to the case where $W$ is sampled from the Bayes posterior \eqref{eq:Bayes_classif}, with the Nishimori identities \cite{Nishimori2001StatisticalPO,Iba1998TheNL} ensuring its statistics are Gaussian.

\subsection{Derivation sketch for $\Phi$}
This subsection deals with the cross-covariance $\Phi$, which is needed to connect with the results of \cite{Loureiro2021CapturingTL} to compute closed-form sharp asymptotics for the test error of linear ERM procedures. We first derive two preliminary results, corresponding to propagations through single layers, and conclude by constructing the straightforward recursion.
\paragraph{Two Gaussians propagating through two layers}
Consider two jointly Gaussian variables $u\in\mathbb{R}^{d},~v\in\mathbb{R}^k$
\begin{equation}
   (u,v)\sim \mathcal{N}\left(
\begin{array}{cc}
     \Psi &\Phi  \\
     \Phi^\top &\Omega
\end{array}\right)
\end{equation}
each independently propagated through a non-linear layer
\begin{align}
    h^\star(u)=\sigma_\star\left(\frac{1}{ \sqrt{d_\star}}W_\star\cdot u\right),
    &&
    h(v)=\sigma\left(\frac{1}{ \sqrt{d}}W\cdot v\right).
\end{align}
The weights $W_\star\in\mathbb{R}^{k_\star\times d_\star}$ and $W\in\mathbb{R}^{k\times d}$ have independently sampled Gaussian entries, with respective variance $\Delta_\star$ and $\Delta$. The $i,j-$th element of the cross-covariance $\Phi^h$ can be expressed as

\begin{align}
    \Phi^h_{ij}=\langle h^\star_i(u)h_j(v)\rangle_{u,v}=\int \frac{e^{-\frac{1}{2}\begin{pmatrix}
    x&y
    \end{pmatrix}
    \begin{pmatrix}
    \frac{w_i^{\star\top} \Sigma w^\star_i}{d_\star} & \frac{w_i^{\star\top} \Sigma w_j}{\sqrt{d_\star d}}\\
    \frac{w_i^{\star\top} \Sigma w_j}{\sqrt{d_\star d}}& \frac{w_j^\top \Sigma w_j}{d}
    \end{pmatrix}^{-1}
    \begin{pmatrix}
    x\\y
    \end{pmatrix}
    }}{\sqrt{\det 2\pi \begin{pmatrix}
    \frac{w_i^{\star\top} \Sigma w^\star_i}{d_\star} & \frac{w_i^{\star\top} \Sigma w_j}{\sqrt{d_\star d}}\\
    \frac{w_i^{\star\top} \Sigma w_j}{\sqrt{d_\star d}}& \frac{w_j^\top \Sigma w_j}{d}
    \end{pmatrix}}}\sigma_\star(x)\sigma(y).
\end{align}
As before, the random variables $\sfrac{w_i^{\star\top} \Sigma w^\star_i}{d_\star}$ and $\sfrac{w_j^\top \Sigma w_j}{d}$ concentrate around their mean value 
\begin{align}
    r_\star\equiv\frac{\Delta_\star}{d_\star}\tr \Psi,
    &&
    r\equiv \frac{\Delta}{d}\tr \Omega.
\end{align}
Plugging these definitions into the above:
\begin{align}
    \Phi^h_{ij}&=\int \frac{e^{-\frac{1}{2}\frac{1}{r_\star r-\mathcal{O}\left(\frac{1}{d}\right)}(r x^2+r_\star y^2)}e^{\frac{1}{r_\star r-\mathcal{O}\left(\frac{1}{d}\right)}\frac{w_i^{\star\top}\Phi w_j}{\sqrt{d_\star d}}xy}}{2\pi\sqrt{r_\star r-\mathcal{O}\left(\frac{1}{d}\right)}}\sigma_\star(x)\sigma(y)\notag\\
    &=\left(\int \frac{e^{-\frac{1}{2r_\star}z^2}}{\sqrt{2\pi r_\star}}\sigma_\star(z)\right)\left(\int \frac{e^{-\frac{1}{2r}z^2}}{\sqrt{2\pi r}}\sigma(z)\right)+\frac{1}{r_\star r}
    \frac{w_i^{\star\top}\Phi w_j}{\sqrt{d_\star d}}
    \left(\int \frac{e^{-\frac{1}{2r_\star}z^2}}{\sqrt{2\pi r_\star}}z\sigma_\star(z)\right)\left(\int \frac{e^{-\frac{1}{2r}z^2}}{\sqrt{2\pi r}}z\sigma(z)\right)+\mathcal{O}\left(\frac{1}{d}\right)\notag\\
    &:=\kappa_1\kappa_1^\star\times  \frac{w_i^{\star\top}\Phi w_j}{\sqrt{d_\star d}},
\end{align}

yielding
\begin{align}
\label{App:eq:propgation_Phi_DRM}
    \Phi^h=\kappa_1\kappa_1^\star\frac{W_\star\Phi W^\top}{\sqrt{d_\star d}},
\end{align}
with
\begin{align}
    \kappa_1=\frac{1}{r}\mathbb{E}_{z}^{\mathcal{N}(0,r)}\left[z\sigma(z)\right]&&\kappa_1=\frac{1}{r_\star}\mathbb{E}_{z}^{\mathcal{N}(0,r_\star)}\left[z\sigma_\star(z)\right].
\end{align}

\paragraph{One Gaussian propagating through one layer}
Consider two jointly Gaussian variables $u\in\mathbb{R}^{d_\star},~v\in\mathbb{R}^d$
\begin{equation}
   (u,v)\sim \mathcal{N}\left(
\begin{array}{cc}
     \Psi &\Phi  \\
     \Phi^\top &\Omega
\end{array}\right)
\end{equation}
with \textit{only} $v$
being propagated through a non linear layer \begin{align}
    h(v)=\sigma\left(\frac{1}{ \sqrt{k}}W\cdot v\right).
\end{align}
The entries $W\in\mathbb{R}^{k\times d}$ are independently sampled from a Gaussian distribution with variance $\Delta$. The $i,j-$th element of the cross-covariance $\Phi$ between $h(v)$ and $u$ can be expressed as
\begin{align}
    \Phi^h_{ij}=\langle u_ih_j(v)\rangle_{u,v}=\int \frac{e^{-\frac{1}{2}\begin{pmatrix}
    x&y
    \end{pmatrix}
    \begin{pmatrix}
    \Psi_{ii} & \frac{\Phi_i w_j}{\sqrt{k}}\\
    \frac{\Phi_i w_j}{\sqrt{k}}& \frac{w_j^\top \Sigma w_j}{k}
    \end{pmatrix}^{-1}
    \begin{pmatrix}
    x\\y
    \end{pmatrix}
    }}{\sqrt{\det 2\pi \begin{pmatrix}
    \Psi_{ii} & \frac{\Phi_i w_j}{\sqrt{k}}\\
    \frac{\Phi_i w_j}{\sqrt{k}}& \frac{w_j^\top \Sigma w_j}{k}
    \end{pmatrix}}}x\sigma(y).
\end{align}
Again, the random variable $\sfrac{w_j^\top \Sigma w_j}{k}$ concentrate around its mean value 
\begin{align}
    r\equiv \frac{\Delta}{k}\tr \Omega.
\end{align}
Plugging this definition into the above:
\begin{align}
    \Phi^h_{ij}&=\int \frac{e^{-\frac{1}{2}\frac{1}{\Psi_{ii} r-\mathcal{O}\left(\frac{1}{d}\right)}(rx^2+\Psi_{ii}y^2)}e^{\frac{1}{\Psi_{ii}  r-\mathcal{O}\left(\frac{1}{d}\right)}\frac{\Phi_i w_j}{\sqrt{k}}xy}}{2\pi\sqrt{\Psi_{ii}  r-\mathcal{O}\left(\frac{1}{d}\right)}}x\sigma(y)\notag\\
    &=\left(\int \frac{e^{-\frac{1}{2\Psi_{ii} }z^2}}{\sqrt{2\pi \Psi_{ii} }}z\right)\left(\int \frac{e^{-\frac{1}{2r}z^2}}{\sqrt{2\pi r}}\sigma(z)\right)+\frac{1}{\Psi_{ii}   r}
    \frac{\Phi_i w_j}{\sqrt{k}}
    \left(\int \frac{e^{-\frac{1}{2\Psi_{ii}}z^2}}{\sqrt{2\pi \Psi_{ii}}}z^2\right)\left(\int \frac{e^{-\frac{1}{2r}z^2}}{\sqrt{2\pi r}}z\sigma(z)\right)+\mathcal{O}\left(\frac{1}{d}\right)\notag\\
    &:=\kappa_1\times  \frac{\Phi_i w_j}{\sqrt{k}}
\end{align}

yielding
\begin{align}
\label{App:eq:propgation_Phi_HMM}
    \Phi^h=\kappa_1\frac{\Phi W^\top}{\sqrt{k}}
\end{align}
with
\begin{align}
    \kappa_1=\frac{1}{r}\mathbb{E}_{z}^{\mathcal{N}(0,r)}\left[z\sigma(z)\right].
\end{align}

\paragraph{Iterating}
Iterating \eqref{App:eq:propgation_Phi_HMM} $L$ times under Gaussianity assumptions on the post-activations yields the cross-covariance between the last layer post-activations of the teacher and the input of a linear ERM (in the present work, ridge regression/classification, and logistic regression), which can in turn be plugged into \cite{Loureiro2021CapturingTL} to access sharp error asymptotics. The exact expressions for the covariances $\Psi,\Phi$ and $\Omega$ used are detailed on a case-per-case basis in Appendix \ref{App:ERM} and \ref{App:Classification}.

Applying \eqref{App:eq:propgation_Phi_DRM} and then \eqref{App:eq:propgation_Phi_HMM} $L-1$ times yields the cross covariance between the last layer post-activations of the teacher and the random features in the framework of random features regression. This extends the results of \cite{Gerace2020GeneralisationEI} to a multilayer target function.

This result carries over to the case where $W,W_\star$ are sampled from the Bayes posterior (or, because of the Nishimori identites \cite{Nishimori2001StatisticalPO,Iba1998TheNL}, when $W_\star$ is a the target weight), assuming non-specialization \ref{conj:non-spec}. The latter ensures that the term $\sfrac{w_i^{\star\top}\Phi w_j}{\sqrt{dd_\star}}$ stay of order $\sfrac{1}{\sqrt{d}}$ for all $i,j$.

\subsection{Numerical evidence of the closed-form recursions}
\begin{figure}
    \centering
    \includegraphics[scale=0.4]{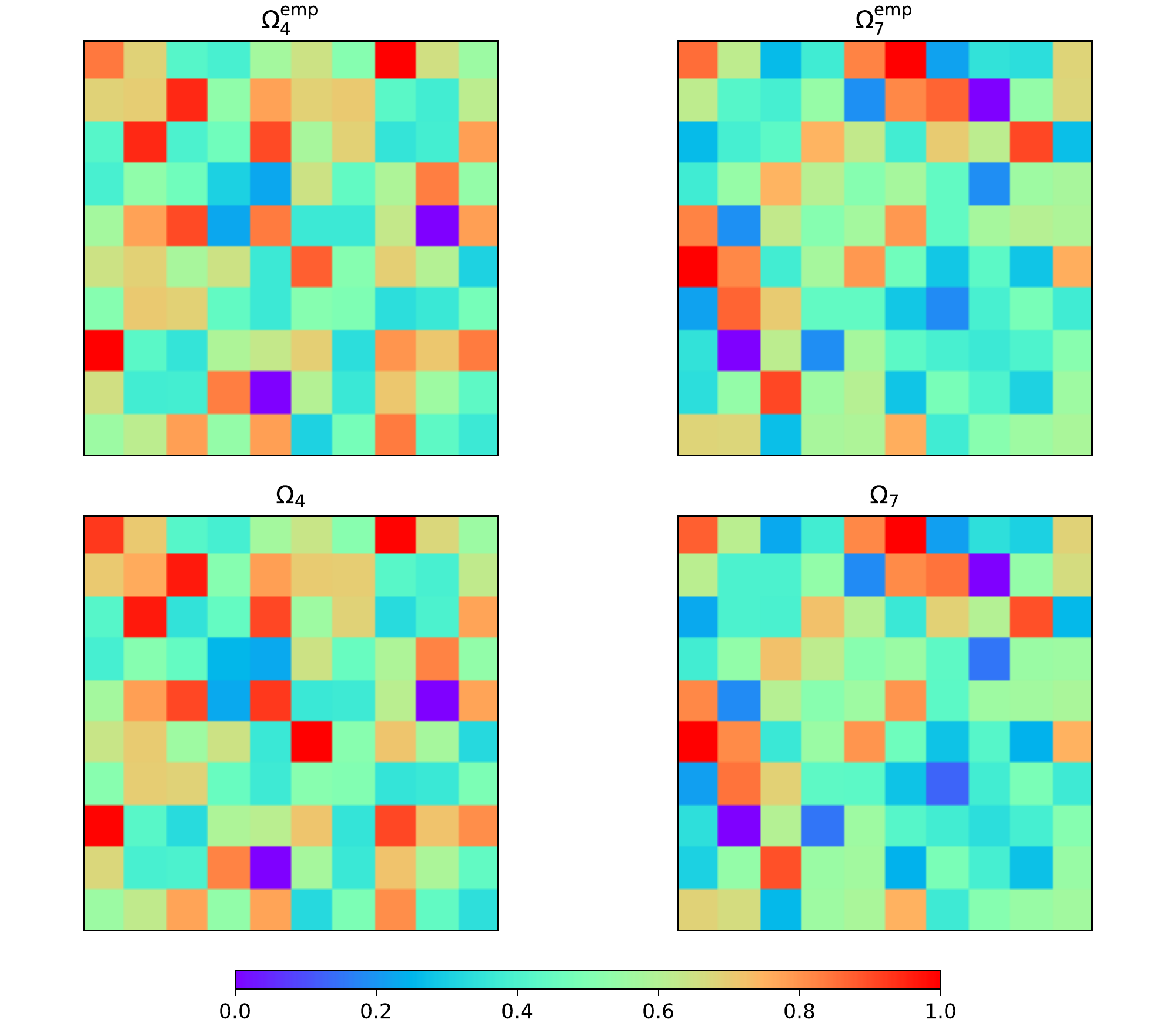}
    \caption{First 10 rows of the population covariance matrix of activations at layers $\ell=4$ (left) and $\ell=7$(right), for a random network with tanh activation and architecture $\forall\ell,~\gamma_\ell=1$, in dimension $d=500$.Top: numerically computed population covariance $\Omega^{\mathrm{emp.}}_\ell$ estimated from $N=10^5$ samples. Bottom: theoretical $\Omega_\ell$, computed from the closed-form formula \eqref{eq:Omega_Psi_multilayer}.}
    \label{fig:check_COv_tanh}
\end{figure}

\begin{figure}
    \centering
    \includegraphics[scale=0.4]{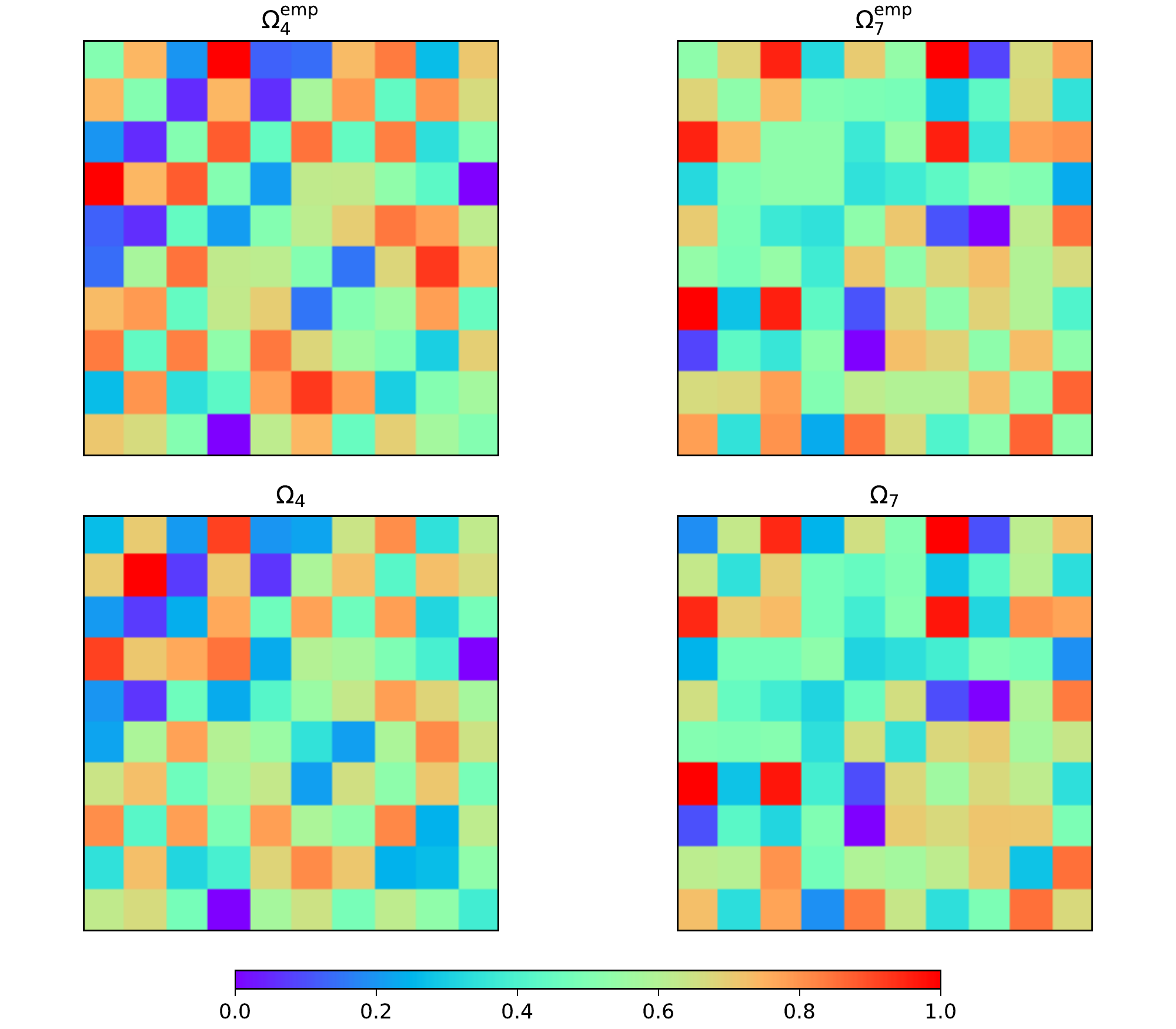}
    \caption{First 10 rows of the population covariance matrix of activations at layers $\ell=4$ (left) and $\ell=7$(right), for a random network with sign activation and architecture $\forall\ell,~\gamma_\ell=1$, in dimension $d=500$.Top: numerically computed population covariance $\Omega^{\mathrm{emp.}}_\ell$ estimated from $N=10^5$ samples. Bottom: theoretical $\Omega_\ell$, computed from the closed-form formula \eqref{eq:Omega_Psi_multilayer}. Due to the nature of the sign activation, the finite size effects are slightly stronger than for the tanh activation (Fig.\ref{fig:check_COv_tanh}).}
    \label{fig:check_COv_sign}
\end{figure}

\begin{figure}
    \centering
    \includegraphics[scale=0.4]{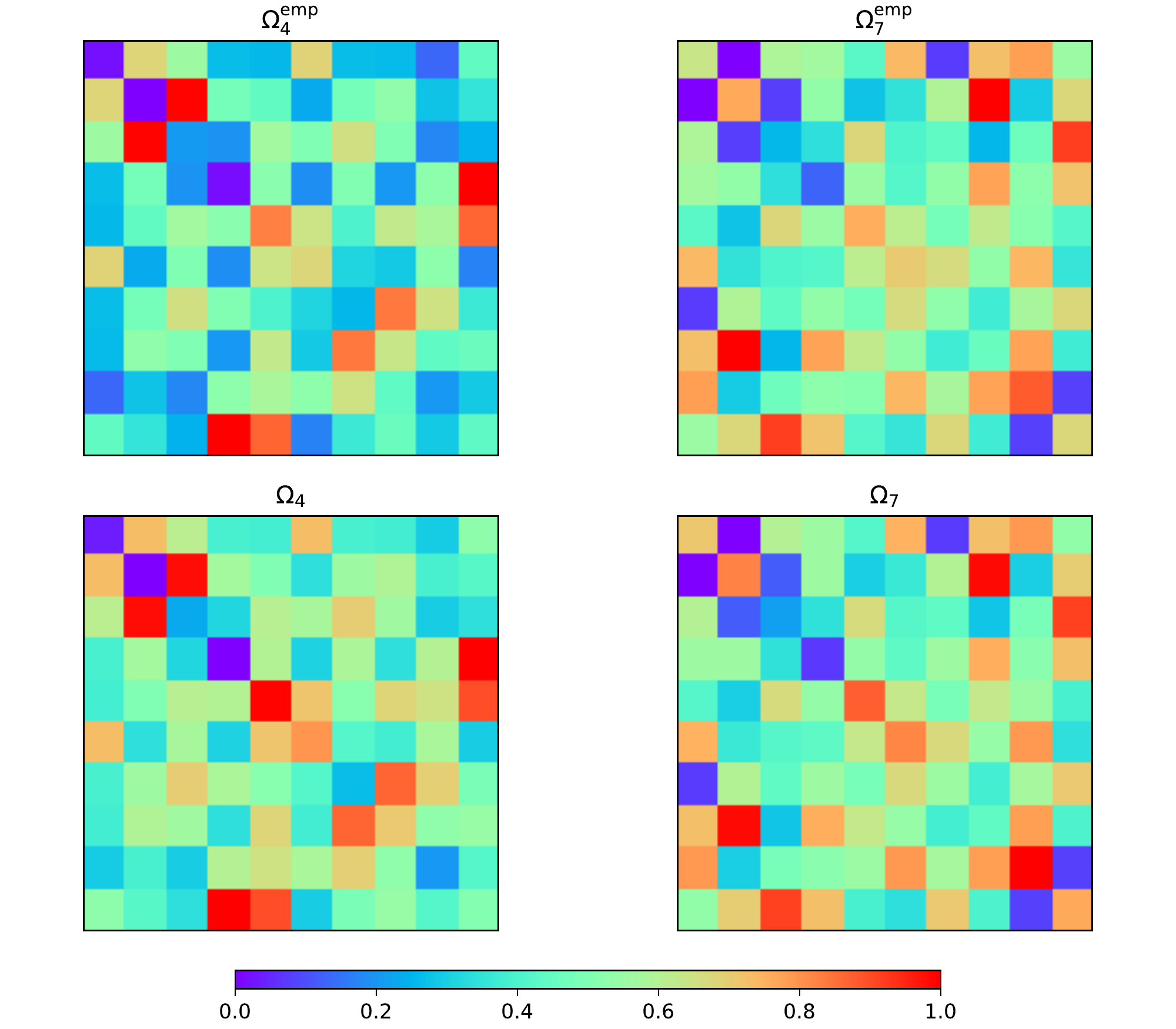}
    \caption{First 10 rows of the population covariance matrix of activations at layers $\ell=4$ (left) and $\ell=7$(right), for a random network with erf activation and architecture $\forall\ell,~\gamma_\ell=1$, in dimension $d=500$.Top: numerically computed population covariance $\Omega^{\mathrm{emp.}}_\ell$ estimated from $N=10^5$ samples. Bottom: theoretical $\Omega_\ell$, computed from the closed-form formula \eqref{eq:Omega_Psi_multilayer}.}
    \label{fig:check_COv_erf}
\end{figure}

We close the discussion on the closed-form formulae for the population covariances by providing numerical evidence of their validity. Figures \ref{fig:check_COv_tanh}, \ref{fig:check_COv_sign} and \ref{fig:check_COv_erf} show the first $10$ rows and columns of the numerically estimated population covariance $\Omega^{\mathrm{emp.}}$, estimated from the sample covariance of $N=10^5$ Gaussian samples $x$, and the closed-form formula \eqref{eq:Omega_Psi_multilayer}, for a rectangular network in dimension $d=500$ and tanh/sign/erf activations. These covariances concern the activations at layers $\ell=4$ and $\ell=7$. In all these plots, for clarity, a matrix $t_\ell \mathbb{I}_{k_\ell}$ was substracted, with
\begin{equation}
    t_\ell=\sum\limits_{\ell^0=1}^{\ell-1}\left(\kappa_*^{\ell^0}\right)^2\prod\limits_{\ell^\prime=\ell^0+1}^L \left(\kappa_1^{\ell^\prime}\right)^2\Delta_{\ell^\prime}
    +\left(\kappa_*^L\right)^2+\int zd\mu(z)\prod\limits_{\ell^0=1}^\ell \left(\kappa_1^{\ell^0}\right)^2\Delta_{\ell^0}.
\end{equation}
$t_\ell$ is the theoretical average value of $\sfrac{\tr\Omega_\ell}{d}$. The visual agreement between the numerically estimated $\Omega^{\mathrm{emp.}}_\ell$ and the theoretical $\Omega_\ell$ is very good, with small relative distances $\sfrac{||\Omega^{\mathrm{emp.}}_\ell-\Omega_\ell||_F^2}{||\Omega^{\mathrm{emp.}}||_F^2}\approx 0.005$ for tanh, $\sfrac{||\Omega^{\mathrm{emp.}}_\ell-\Omega_\ell||_F^2}{||\Omega^{\mathrm{emp.}}||_F^2}\approx 0.008$ for sign, and $\sfrac{||\Omega^{\mathrm{emp.}}_\ell-\Omega_\ell||_F^2}{||\Omega^{\mathrm{emp.}}||_F^2}\approx 0.004$ for erf.

\subsection{Numerical test of the 1dCLT}

We provide numerical evidence of the 1dCLT \ref{prop:dGP}, namely that the output $\hat{y}(x)$ of a deep random Gaussian network is Gaussian. Note that from the closed-form formula \eqref{eq:Omega_Psi_multilayer}, the theoretical variance of the output reads
\begin{equation}
\label{eq:App:GET:traceomega}
\check{q}=\sum\limits_{\ell=1}^{L-1}\left(\kappa_*^\ell\right)^2\prod\limits_{\ell^\prime=\ell+1}^L \left(\kappa_1^{\ell^\prime}\right)^2\Delta_{\ell^\prime}
    +\left(\kappa_*^L\right)^2+\int zd\mu(z)\prod\limits_{\ell=1}^L \left(\kappa_1^{\ell}\right)^2\Delta_\ell.
\end{equation}

\begin{figure}
    \centering
    \includegraphics[scale=.45]{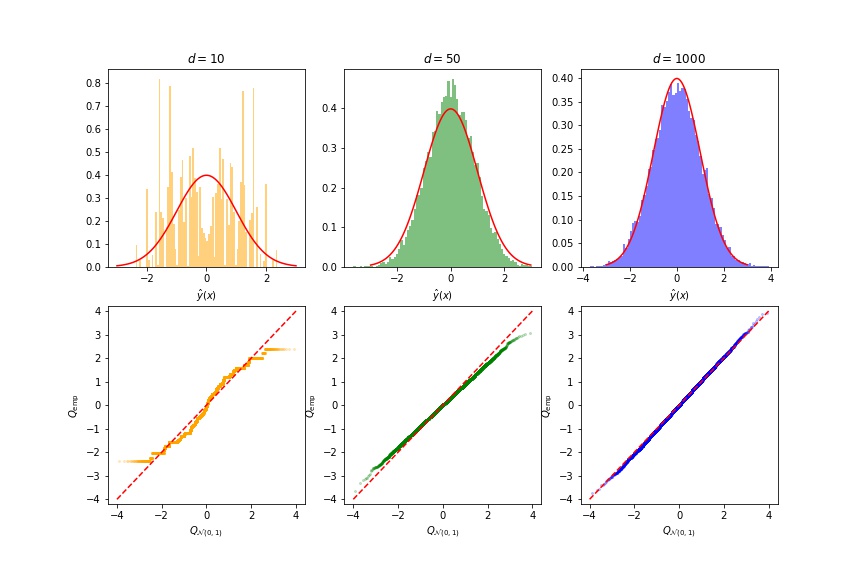}
    \includegraphics[scale=.45]{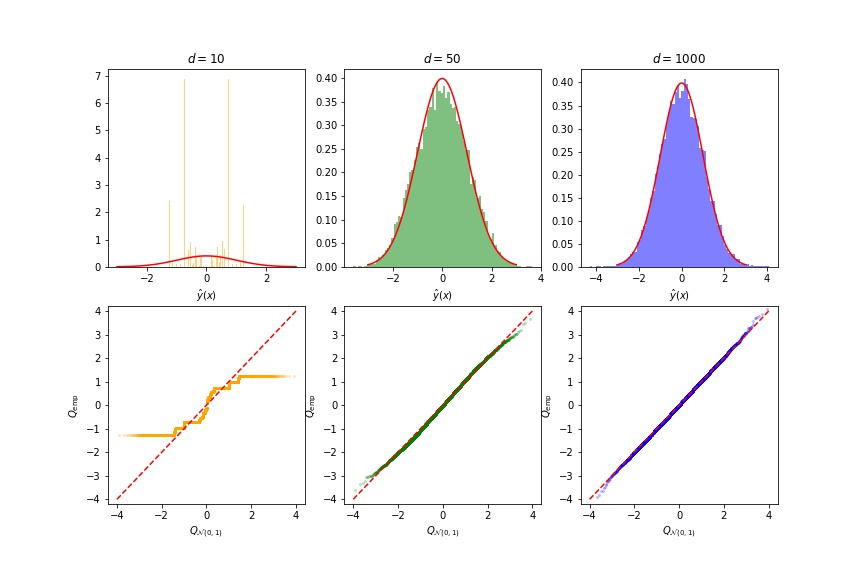}
    \caption{We study the rescaled output $\sfrac{\hat{y}(x)}{\sqrt{\check{q}}}$of a random Gaussian network, $L=3$ (top) and $L=9$ (bottom) hidden layers, sign activations, and widths $k_1=...=k_L=d$. From left to right, the dimension $d$ is varied from $d=10$ to $d\sim 1000$. (top) Histogram of the output. The red line represents a normal distribution. (bottom) Quantile-Quantile plot. The x axis represent the theoretical quantiles of a normal distribution $Q_{\mathcal{N}(0,1)}$, while the empirical quantiles are plotted on the y axis.}
    \label{fig:test_gaussian_sign}
\end{figure}

\begin{figure}
    \centering
    \includegraphics[scale=.45]{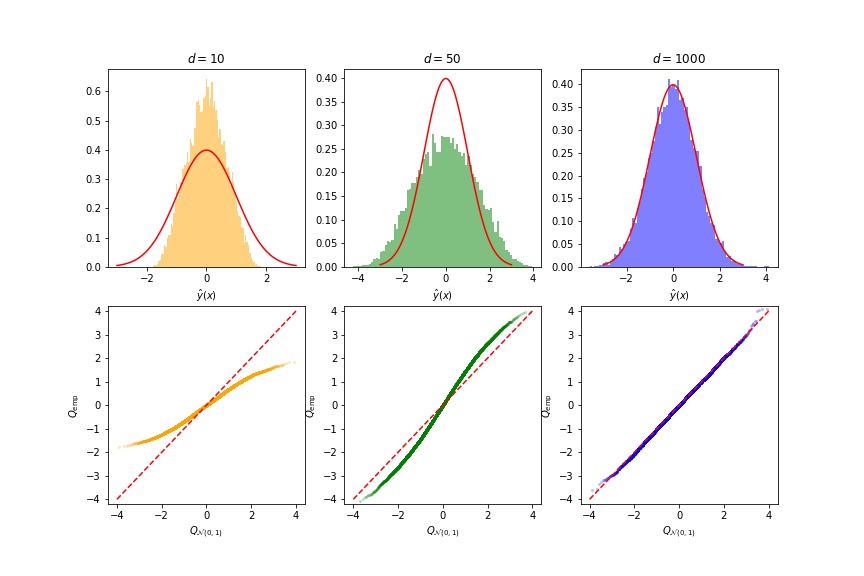}
    \includegraphics[scale=.45]{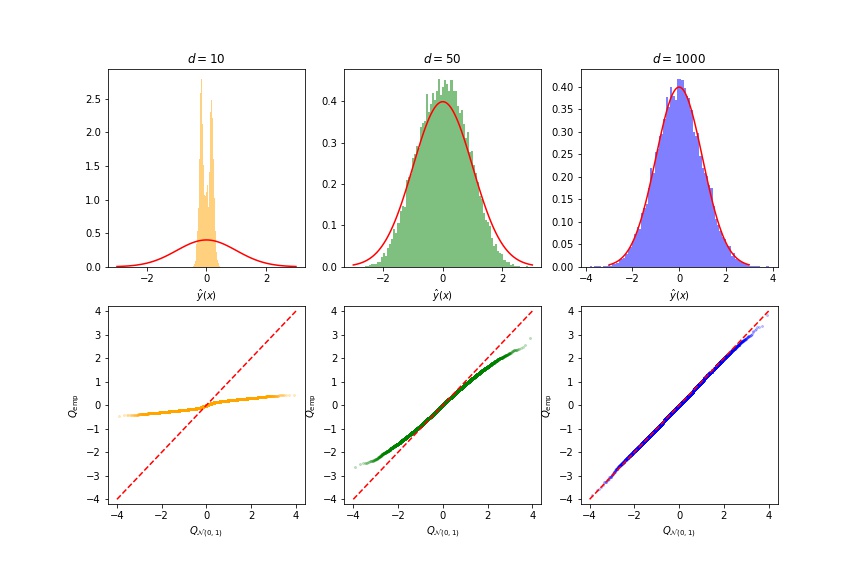}
    \caption{We study the output $\sfrac{\hat{y}(x)}{\sqrt{\check{q}}}$ of a random Gaussian network, $L=3$ (top) and $L=9$ (bottom) hidden layers, tanh activations, and widths $k_1=...=k_L=d$. From left to right, the dimension $d$ is varied from $d=10$ to $d\sim 1000$. (top) Histogram of the output. The red line represents a normal distribution. (bottom) Quantile-Quantile plot. The x axis represent the theoretical quantiles of a normal distribution $Q_{\mathcal{N}(0,1)}$, while the empirical quantiles are plotted on the y axis.}
    \label{fig:test_gaussian_tanh}
\end{figure}

\begin{figure}
    \centering
    \includegraphics[scale=.55]{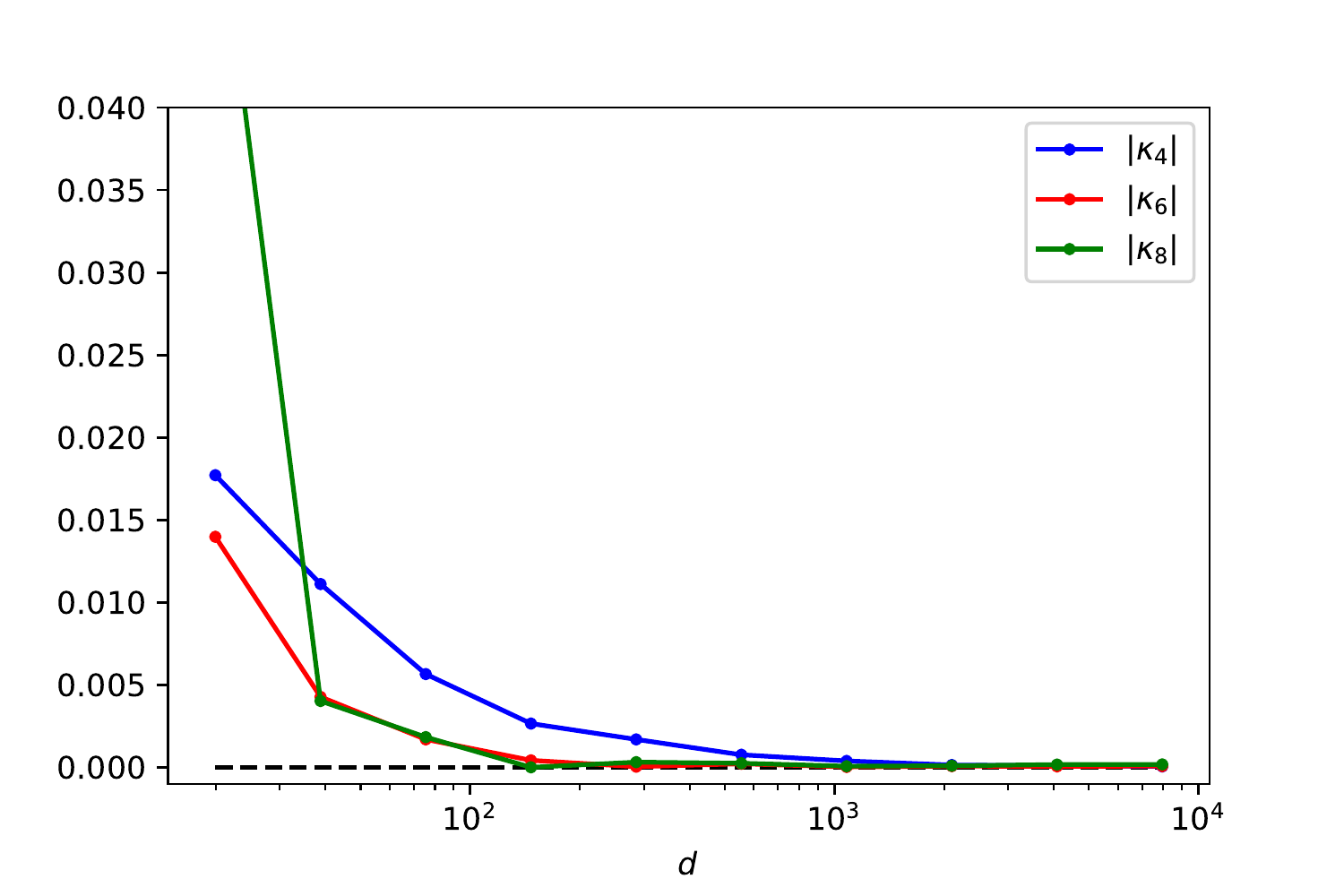}
    \caption{Order $4,6,8$ of the output $\sfrac{\hat{y}(x)}{\sqrt{\check{q}}}$ of a random Gaussian network of depth $L=3$ with tanh activations, and widths $k_1=...=k_L=d$. The cumulants were evaluated from $N=5.10^4$ samples, for one realization of the weights. Note that because of the fact that the activation is odd, all odd cumulants are vanishing.}
    \label{fig:cumulants}
\end{figure}

The empirical histograms of the scaled output $\sfrac{\hat{y}(x)}{\sqrt{\check{q}}}$ of networks with $L=3$ and $L=9$ hidden layers are plotted in Fig.\,\ref{fig:test_gaussian_tanh} (for tanh activation) and Fig.\,\ref{fig:test_gaussian_sign} (for sign activation). In all cases, the widths are taken equal to the input dimension $d$. Several sizes $d$ are plotted. As supplementary confirmation, quantile-quantile (QQ) plots are also included. \camera{Fig.\,\ref{fig:cumulants} further shows, for a depth $L=3$ network with tanh activation, that the higher order cumulants of the output distribution is asymptotically suppressed.}. These numerical simulation provide compelling evidence that as $d\rightarrow\infty$ the distribution of $\sfrac{\hat{y}(x)}{\sqrt{\check{q}}}$ converges to a normal distribution.

\subsection{Towards rigorous proofs}
It would be desirable to rigorously prove the conjectured formulae discussed in this paper. There are a few difficulties that we discuss now briefly.

Concerning the single layer conjecture \ref{prop:BGP_2l}(\textbf{Shallow Bayes GEP}), a full theorem can probably be within reach of current techniques with, however, one difficulty, summarized in conjecture \ref{conj:non-spec}. Indeed, the Gaussian equivalence theorem applied to two neural nets with weights $W_1$ and $W_2$ implies that both of them are effectively linear so that, in terms of second-order statistics, $\sigma(W_1 x) \approx \kappa_1^{(1)}W_1 X + \kappa_*^{(1)}Z_1$ and $\sigma(W_2 x) \approx \kappa_1^{(1)} W_2 X + \kappa_*^{(1)}Z_2$, with $Z_{1,2}$ Gaussian-distributed. When $W_1$ and $W_2$ are independent, so are $Z_1$ and $Z_2$. However, when $W_1$ and $W_2$ are correlated then $Z_1$ and $Z_2$ becomes correlated Gaussians as well, thus complicating the rigorous analysis. As noted in the main text, the work of \cite{Aubin2018TheCM} shows that this is not expected to happens, but a rigorous control remains so far elusive, and is left for future work.

The deep conjecture requires significantly more work, as an additional difficulty arises. While conjecture \ref{prop:BGP_ML} should also be amenable to a rigorous treatment similar to the one of conjecture \ref{prop:BGP_2l}, we are still lacking a control of a deep GET as expressed in \ref{prop:dGP} \eqref{eq:Omega_Psi_multilayer}. While much progress has been attained on the topic, giving very strong indication for its validity (see e.g. \cite{Fan2020SpectraOT,DRF2023,Bosch2023PreciseAA}) this is still an open problem mathematically. 


%% file: Appendix/2layers.tex
This appendix provides a detailed derivation of the asymptotic formula for the test error of the Bayes optimal MSE using the replica method \cite{Replica, Zdeborov2015StatisticalPO, Gabrie2019MeanfieldIM}. The replica method has been leveraged in a very large body of works \cite{Aubin2018TheCM,Aubin2020GeneralizationEI,Gerace2020GeneralisationEI,Maillard2020PhaseRI,Barbier2017OptimalEA} to access exact asymptotic characterizations of accuracy metrics in simple learning tasks, for both Bayesian learning and ERM.\\

In this appendix, we consider the case of a two-layer target
$$
y^\mu\sim P_{\mathrm{out}}\left(\cdot |\frac{1}{k_\star}a_\star^\top \sigma_\star\left(
\frac{1}{ \sqrt{d}}W_\star^\top x^\mu
\right)\right)
$$
and study Bayesian learning using a two-layer network
$$
\hat{y}(x)=\frac{1}{k}a_\star^\top \sigma\left(
\frac{1}{ \sqrt{d}}W^\top x^\mu
\right).
$$
Note that the large part of the computation is detailed for the generic case where the student architecture is not required to match that of the target, i.e. $k\ne k_\star$. We denote by $P_{\mathrm{out}}$ the Gaussian measure of variance $\Delta$, corresponding to the Gaussian additive channel \eqref{eq:teacher}. Similarly, we adopt the notation $P_w(\cdot)$ (resp. $P_a(\cdot)$) for the Gaussian distribution of variance $\Delta_w$ (resp. $\Delta_a$). In the last sections, we specialize to the Bayes optimal setting $\sigma=\sigma_\star$ and $k=k_\star$ to derive the Bayes-optimal MSE \eqref{eq:Bayes_regression}. The general case with deep target and student is discussed in Appendix \ref{App:multilayer}.\\

The exact computation of the Bayes posterior measure \eqref{eq:Bayes_post_classif} is generically intractable. However, it is possible to compute the associated free energy
$$
f\equiv -\mathbb{E}_{a_\star,\{W_\ell^\star\}}\mathbb{E}_{\mathcal{D}}\ln Z
$$
where $Z$ is the partition function associated to the measure \eqref{eq:Bayes_post_classif}, i.e.
\begin{align}
Z=\int \dd a \int \dd W P_w(W)P_a(a)
P_{\mathrm{out}}(y^\mu|\hat{y}(x^\mu)).
\end{align}
 We introduced the priors
\begin{align}
    P_w(W)\propto e^{-\frac{1}{2\Delta_w}||W_\ell||_F^2},&&
    P_a(a)\propto e^{-\frac{1}{2\Delta_a}||a||^2},
\end{align}
and the posteriors 
\begin{align}
    P_{\mathrm{out}}(y|\hat{y})\propto e^{-\frac{
1}{2\Delta}\left(y-\hat{y}\right)^2
}
\end{align}
The computation of the free energy yields asymptotic characterizations of the overlaps
\begin{align*}
q_w=\mathbb{E} \left \lVert\left\langle \frac{a^\top W}{\sqrt{kd}}\right\rangle\right\lVert^2,
&&
q_a=\mathbb{E} \left\langle
\left\lVert \frac{a}{\sqrt{k}}\right\lVert^2
\right\rangle,
&&
m_w=\mathbb{E} \frac{\left\langle a^\top W\right\rangle W_\star^\top a_\star}{\sqrt{kk_\star }d},
\end{align*}
which fully describe the asymptotic average test error. In order to compute the free energy, the \textit{replica trick} \cite{Replica,Replica2}
$$
\ln Z=\underset{s\rightarrow 0}{\lim}\frac{Z^s-1}{s}
$$
can be employed.

\subsection{Replica trick}
The replicated partition function reads
\begin{align}
    Z^s=\int \prod_{a=1}^sdW^ada^aP_w(W^a)P_a(a^a)
    \underbrace{
    \mathbb{E}_{\{x^\mu\}_\mu}\scriptstyle \int \prod\limits_{\mu=1}^nP_{\mathrm{out}}^\star\left(y^\mu|\frac{1}{\sqrt{k_\star}}a_\star^\top \sigma_\star\left(\frac{1}{\sqrt{d}}
    W_\star x^\mu
    \right)\right)\prod\limits_{\mu=1}^n\prod_{a=1}^s
    P_{\mathrm{out}}\left(y^\mu|\frac{1}{\sqrt{k}}a_a^\top \sigma_\star\left(\frac{1}{\sqrt{d}}
    W_a x^\mu
    \right)\right)
    }_{*}.
\end{align}
The expectation with respect to the train set in the energetic part can be carried out as:
\begin{align}
    *=\prod_{\mu}\left[
    \mathbb{E}_{\lambda_\star,\{\lambda_a\}_{a=1}^s}\int dy P_{\mathrm{out}}^\star(y|\lambda_\star)\prod\limits_{a=1}^sP_{\mathrm{out}}(y |\lambda_a)
    \right].
\end{align}
We have introduced the local fields
\begin{align}
    \lambda_\star\equiv \frac{1}{\sqrt{k_\star}}a_\star^\top \sigma_\star\left(\frac{1}{\sqrt{d}}
    W_\star x^\mu
    \right),
    &&\lambda_a=\frac{1}{\sqrt{k}}a_a^\top \sigma_\star\left(\frac{1}{\sqrt{d}}
    W_a x^\mu
    \right).
\end{align}

\subsection{GET-linearization of the hidden layer}
The shallow Bayes GEP \eqref{prop:BGP_2l} implies that these $s+1$ fields are jointly Gaussian and have statistics
\begin{align}
\label{eq:App:2L:local_fields_stats}
\begin{cases}
    \langle \lambda_a\lambda_b\rangle  =\kappa_1^2 \frac{1}{kd}a_a^\top W_a\Sigma W_b^\top a_b+\delta_{ab}\kappa_*^2 \frac{1}{k}a_a^\top a_b\\
    \langle \lambda_a\lambda_\star\rangle  =\kappa_1\kappa_1^\star \frac{1}{dk}a_a^\top W_a\Sigma W_\star^\top a_\star
   \\
    \langle \lambda_\star\lambda_\star\rangle  =\kappa_1^{\star2} \frac{1}{k d}a_\star^\top W_\star\Sigma W_\star^\top a_\star+\kappa_*^{\star 2} \frac{1}{k}a_\star^\top a_\star
    \end{cases},
\end{align}
where the Kronecker delta $\delta_{ab}$ arises because of the non-specialization conjecture \ref{conj:non-spec}. We defined

\begin{align}
\begin{cases}
\kappa_1=\frac{1}{\frac{\Tr\Sigma}{\Delta_w^{-1}d}}\mathbb{E}_z^{\mathcal{N}\left(0,\frac{\Tr\Sigma}{\Delta_w^{-1}d}\right)}[z\sigma(z)]
 \\
    \kappa_*=\sqrt{
    \mathbb{E}_z^{\mathcal{N}\left(0,\frac{\Tr\Sigma}{\Delta_w^{-1}d}\right)}[\sigma(z)^2]-\kappa_1^2 \frac{\Tr\Sigma}{\Delta_w^{-1}d}
    }
\end{cases}
\begin{cases}
\kappa_1^\star=\frac{1}{\Delta_w \frac{\Tr\Sigma}{d}}\mathbb{E}_z^{\mathcal{N}\left(0,\Delta_w \frac{\Tr\Sigma}{d}\right)}[z\sigma(z)]
   \\
    \kappa_*^\star=\sqrt{
    \mathbb{E}_z^{\mathcal{N}\left(0,\frac{\Tr\Sigma}{\Delta_w^{-1}d}\right)}[\sigma(z)^2]-\kappa_1^2 \Delta_w \frac{\Tr\Sigma}{d}
    }
\end{cases},
\end{align}
see \eqref{eq:kappa_multilayer}.

We introduce the order parameters:
\begin{align}
    q^w_{ab}=\frac{1}{kd}a_a^\top W_a\Sigma W_b^\top a_b &&m^w_a=\frac{1}{k d}a_\star^\top W_\star\Sigma W_\star^\top a_\star && q_{ab}^a=\delta_{ab}\frac{1}{k}a_a^\top a_b\\
    \rho_w =\frac{1}{k d}a_\star^\top W_\star\Sigma W_\star^\top a_\star &&\rho_a=\frac{1}{k}a_\star^\top a_\star
\end{align}

Introducing Diracs in their Fourier transform form to enforce those definitions, the replicated partition function reads
\begin{align}
    Z^s=&\int \prod\limits_{a\le b }\dd q^w_{ab}\dd q^a_{ab}\prod\limits_a \dd m^w_a\underbrace{ e^{-\sum\limits_a \hat{m}^w_a m_a^w +\sum\limits_{a\le b} (\hat{q}^w_{ab}q^w_{ab}+\hat{q}^a_{ab}q^a_{ab})}}_{e^{sd\Psi_t}}\notag\\
    &
    \underbrace{\int \prod\limits_{a=1}^s dW_ada_aP_w(W_a)P_a(a_a)e^{\sum\limits_{a\le b}\hat{q}^w_{ab}\frac{a_a^\top W_a\Sigma W_b^\top a_b}{k}+\sum\limits_{a\le b}\hat{q}^a_{ab}\frac{a_a^\top a_b}{\gamma}+\sum\limits_a \hat{m}^w_a\frac{a_a^\top W_a \Sigma W_\star^\top a_\star}{\sqrt{kk^\star}}
    }}_{e^{sd\Psi_w}}\notag\\
    &\underbrace{\prod_{\mu}\left[
    \mathbb{E}_{\lambda_\star,\{\lambda_a\}_{a=1}^s}\int dy P_{\mathrm{out}}^\star(y|\lambda_\star)\prod\limits_{a=1}^sP_{\mathrm{out}}(y |\lambda_a)
    \right]}_{e^{sd\Psi_y}}
\end{align}

\subsection{Entropic potential}
We first compute the entropic potential $\Psi_w$.
\begin{align}
    e^{sd \Psi_w}&=\int \prod\limits_{a=1}^s dW_ada_aP_w(W_a)P_a(a_a)e^{\sum\limits_{a\le b}\hat{q}^w_{ab}\frac{a_a^\top W_a\Sigma W_b^\top a_b}{k}+\sum\limits_{a\le b}\hat{q}^a_{ab}\frac{a_a^\top a_b}{\gamma}+\sum\limits_a \hat{m}^w_a\frac{a_a^\top W_a \Sigma W_\star^\top a_\star}{\sqrt{kk^\star}}
    }
    \\
    &=\underbrace{\int \prod\limits_{a=1}^s da_aP_a(a_a)e^{\sum\limits_{a\le b}\hat{q}^a_{ab}\frac{a_a^\top a_b}{\gamma}
    }}_{(a)}\prod\limits_{j=1}^d\underbrace{\left[
    \int \prod\limits_{a=1}^s dw_aP_w(w_a)e^{\sum\limits_{a\le b}\sigma_j\hat{q}^w_{ab}\frac{(a_a^\top w_a)(a_b^\top w_b) }{k}+\sum\limits_a \sigma_j\hat{m}_a\frac{(a_a^\top W_a) (a_\star\top w^\star_j)}{\sqrt{kk^\star}}}
    \right]}_{(b)}
\end{align}
We have used the fact that the prior $P_w(\cdot)$, being Gaussian, factorizes over the $d$ columns $\{w_j\}_{j=1}^d$  of the first layer weights $W, W^\star$. Also, $\Sigma$ has been without loss of generality supposed to be diagonal, i.e. $\Sigma=\mathrm{diag}(\sigma_1,...,\sigma_d)$. Note that all the final expression hold for generic $\Sigma$, and that one can carry out the computation in full detail without this shortcut by introducing $\Sigma=U\mathrm{diag}(\sigma_1,...,\sigma_d)U^\top$ and $U-$ rotating $W,W^\star$. Defining
\begin{equation}
    \eta_a=\frac{a_a^\top w_a}{\sqrt{k}}
\end{equation}
and remembering $P_w(w)=e^{-\frac{\Delta_w^{-1}}{2}w\top w}$, it follows that the variables $\{\eta_a\}_{a=1}^s$ are jointly Gaussian with statistics
\begin{align}
\label{eq:Psi_w_before_RS}
    \begin{cases}
        \langle \eta_a\rangle=0\\
        \langle \eta_a\eta_b\rangle=\delta_{ab}\frac{q^a_{aa}}{\Delta_w^{-1}}.
    \end{cases}
\end{align}
It then follows that
\begin{align}
(b)&=\int \left( \prod\limits_{a=1}^s\frac{d\eta_a}{\sqrt{2\pi\frac{q^a_{aa}}{\Delta_w^{-1}} }} e^{-\frac{1}{2}\frac{\Delta_w^{-1}}{q^a_{aa}}\eta_a^2} 
\right)
e^{\sigma_j\sum\limits_{a\le b}\hat{q}^w_{ab}\eta_a\eta_b+\sigma_j\frac{a_\star\top w^\star_j}{\sqrt{k^\star}}\sum\limits_a \hat{m}_a\eta_a}\\
&=e^{\frac{1}{2}\sigma_j^2\left(\frac{a_\star\top w^\star_j}{\sqrt{k^\star}}\right)^2 \hat{m}^\top \left[
\mathrm{diag}\left(\frac{\Delta_w^{-1}}{q^a_{aa}}\right) -\sigma_j \times (2I_s)\odot \hat{Q}_w 
\right]^{-1}\hat{m}
-\frac{1}{2}\ln \det \left[
I_s-\sigma_j\mathrm{diag}\left( \frac{\Delta_w^{-1}}{q_{aa}^a}\right)(2I_s)\odot \hat{Q}_w
\right]
}.
\end{align}

\subsection{Replica-symmetric ansatz}
Since the integral for $Z^s$ involves exponentials with arguments of order $d$, Laplace method suggests that the integral concentrates over the maximizer of the exponent $\Psi_t+\Psi_w+\Psi_y$. Since explicitly carrying out the optimization over the order parameters $q^w_{ab},q^a_{ab},m^w_{ab}$ is hard, we look for solutions of the form
\begin{align}
    &\hat{q}^w_{ab}=\delta_{ab}\left(-\frac{\hat{r}_w}{2}-\hat{q}_w\right)+\hat{q}_w\\
    &\hat{q}^a_{ab}=-\delta_{ab}\frac{\hat{r}_a}{2}\\
    &\hat{m}^w_a=\hat{m}_w
\end{align}
and
\begin{align}
    &q^w_{ab}=\delta_{ab}(r_w-q_w)+q_w\\
    &q^a_{ab}=\delta_{ab}r_a\\
    & m^w_a=m_w.
\end{align}
This ansatz is standardly known as the \textit{replica symmetric} ansatz.
For convenience, we further define the variances
\begin{align}
    V_w=r_w-q_w, && V_a=r_a, && \hat{V}_w=\hat{r}_w+\hat{q}_w, && \hat{V}_a=\hat{r}_a.
\end{align}

\subsection{RS free energy}
\paragraph{Entropy potential}

We first recall as a preliminary statement two linear algebra results for matrices of the form $(\delta_{ab}(r-q)+q)_{1\le a,b\le s}$:
\begin{align}
    (\delta_{ab}(r-q)+q)^{-1}_{ab}=\left(
    \delta_{ab}\left( \frac{r+(s-2)q}{(r-q)^2+sq(r-q)}+ \frac{q}{(r-q)^2+sq(r-q)}\right)-\frac{q}{(r-q)^2+sq(r-q)}
    \right)_{ab}
\end{align}
and
\begin{align}
    \det (\delta_{ab}(r-q)+q)^{-1}_{ab} = (r+(s-1)q)(r-q)^{s-1}
\end{align}

These identities can be leveraged to simplify the two terms in \eqref{eq:Psi_w_before_RS} as

\begin{align}
    \hat{m}^\top \left[
\mathrm{diag}\left(\frac{\Delta_w^{-1}}{q^a_{aa}}\right) -\sigma_j \times (2I_s)\odot \hat{Q}_w 
\right]^{-1}\hat{m}=\sigma_j^2 s \hat{m}^2\frac{1}{\sigma_j\hat{V}_w+\frac{\Delta_w^{-1}}{r_a}}
\end{align}
and 
\begin{align}
   \ln \det \left[
I_s-\sigma_j\mathrm{diag}\left( \frac{\Delta_w^{-1}}{q_{aa}^a}\right)(2I_s)\odot \hat{Q}_w
\right]=s \ln\left(
\frac{r_a}{\Delta_w^{-1}}\hat{V}_w\sigma_j+1
\right)-s\frac{\sigma_j\hat{q}_w}{\sigma_j\hat{V}_w+\frac{\Delta_w^{-1}}{r_a}}.
\end{align}
Thus 
\begin{align}
    \prod\limits_{j=1}^d (b)=e^{\frac{s\hat{m}^2}{2}\Tr\left[
    \frac{\Sigma^2 \frac{W_\star^\top a_\star a_\star^\top W_\star}{k_\star}}{\hat{V}_w\Sigma+\frac{\Delta_w^{-1}}{r_a}I_d}
    \right]
    -\frac{s}{2}\ln\det\left[
    \frac{r_a}{\Delta_w^{-1}}\hat{V}_w\Sigma+I_d
    \right]
    +\frac{s}{2}\Tr\left[     
    \frac{\hat{q}_w\Sigma}{\hat{V}_w\Sigma+\frac{\Delta_w^{-1}}{r_a}I_d}
    \right]
        }
\end{align}
Computing now the $(a)$ term:
\begin{align}
    (a)&=\prod\limits_{j=1}^k\left[
    \int \prod \limits_{a=1}^s \frac{da_a}{\sqrt{2\pi\frac{1}{\Delta_a^{-1}}}}e^{-\frac{\Delta_a^{-1}}{2}\sum\limits_{a=1}^s a_a^2
    -\frac{
    \hat{r}_a}{2\gamma}\sum\limits_{a=1}^s a_a^2
    }
    \right]
    =e^{
    -\frac{ks}{2}\ln \left(1+\frac{\hat{V}_a}{\gamma\Delta_a^{-1}}\right)}
\end{align}
The entropic potential thus reads
\begin{align}
    \Psi_w=\frac{1}{ds}\ln\left[ (a)\prod\limits_{j=1}^d\right] (b)&=
    -\frac{\gamma}{2}\ln \left(1+\frac{\hat{V}_a}{\Delta_a^{-1}}\right)
    \\
    &+\frac{\hat{m_w}^2}{2d}\Tr\left[
    \frac{\Sigma^2 \frac{W_\star^\top a_\star a_\star^\top W_\star}{k_\star}}{\hat{V}_w\Sigma+\frac{\Delta_w^{-1}}{r_a}I_d}
    \right]
    -\frac{1}{2d}\ln\det\left[
    \frac{r_a}{\Delta_w^{-1}}\hat{V}_w\Sigma+I_d
    \right]
    +\frac{1}{2d}\Tr\left[     
    \frac{\hat{q}_w\Sigma}{\hat{V}_w\Sigma+\frac{\Delta_w^{-1}}{r_a}I_d}
    \right]
\end{align}
where for readability we redefined $\hat{V}_a\leftarrow \hat{V}_a/\gamma$.
\paragraph{Trace terms}
It is straightforward to obtain
\begin{align}
    \Psi_t=\frac{1}{2}(\hat{r}_wr_w+\hat{q}_wq_w)+\frac{\gamma}{2}(\hat{r}_ar_a)-\hat{m}_wm_w
\end{align}

\paragraph{Loss potential}
The loss potential is identical to \cite{Loureiro2021CapturingTL}, provided the relevant overlap are plugged in:
\begin{align}
\label{eq:2l_loss_pot_firstexp}
    \Psi_y=\int D\xi \int dy\mathcal{Z}_\star\left({\scriptstyle 
    y,\frac{\kappa_1\kappa_1^\star m_w}{\sqrt{\kappa_1^2q_w}}\xi, \kappa_1^{\star 2}\rho_w+\kappa_*^{\star2}\rho_a - \frac{\left(\kappa_1\kappa_1^\star m_w\right)^2}{\kappa_1^2q_w}}
    \right)\ln \mathcal{Z}_\ell\left({\scriptstyle
    y,
\sqrt{\kappa_1^2q_w}\xi,\kappa_1^2V_w+\kappa_*^2V_a}
    \right)
\end{align}
with
\begin{align}
\label{eq:intro_Z}
    \mathcal{Z}_{\ell/\star}(y,\omega,V)=\mathbb{E}_x^{\mathcal{N}(\omega,V)}P_{\mathrm{out}}^{/\star}(y|x).
\end{align}

\subsection{Finite temperature free energy}
The free energy thus reads
\begin{align}
\label{eq:finite_T_f}
    f=\underset{V_w,q_w,m_w,V_a,\hat{V}_w,\hat{q}_w,\hat{m}_w,\hat{V}_a}{\mathrm{extr}}&-\frac{1}{2}(\hat{V}_wV_w+\hat{V}_wq_w-\hat{q}_wV_w)-\frac{\gamma}{2}\hat{V}_aV_a+\hat{m}_wm_w+\frac{\gamma}{2}\ln \left(1+\frac{\hat{V}_a}{\Delta_a^{-1}}\right)\notag\\
    &-\frac{1}{2d}\mathbb{E}_{a_\star,W_\star}\Tr\left[
    \frac{\hat{m}^2_w\Sigma^2 \frac{W_\star^\top a_\star a_\star^\top W_\star}{k_\star}+\hat{q}_w\Sigma}{\hat{V}_w\Sigma+\frac{\Delta_w^{-1}}{V_a}I_d}
    \right]
    +\frac{1}{2d}\ln\det\left[
    \frac{V_a}{\Delta_w^{-1}}\hat{V}_w\Sigma+I_d
    \right]\notag\\
    &-\alpha \int D\xi \int dy\scriptstyle\mathcal{Z}_\star\left( 
    y,\frac{\kappa_1\kappa_1^\star m_w}{\sqrt{\kappa_1^2q_w}}\xi, \kappa_1^{\star 2}\rho_w+\kappa_*^{\star2}\rho_a - \frac{(\kappa_1\kappa_1^\star m_w)^2}{\kappa_1^2q_w+\kappa_*^2q_a}
    \right)\ln \mathcal{Z}_\ell\left(
    y,
    \sqrt{\kappa_1^2q_w}\xi,\kappa_1^2V_w+\kappa_*^2V_a
    \right).
\end{align}
In the next section, we specialize this expression to the Bayes optimal setting, for regression. Appendix \ref{App:Classification} provides in parallel the specialization for the Bayes optimal classification case.

\subsection{Bayes-Optimal setting}
\label{sec:BO}

For the linear readout $f_\star=id$ and Gaussian additive noise considered in the regression case, $\mathcal{Z}(y,\omega,V)$ admits a compact expression
\begin{equation}
    \mathcal{Z}(y,\omega,V)=\int \frac{dx}{\sqrt{2\pi V}\sqrt{2\pi\Delta}}e^{-\frac{(x-\omega)^2}{2V}-\frac{1}{2\Delta}(y-x)^2}=\frac{1}{\sqrt{2\pi(\Delta+V)}}e^{-\frac{(\omega-y)^2}{2(\Delta+V)}}.
\end{equation}

\subsection{Nishimori identities}
Since we consider the Bayes optimal case, the Nishimori identities hold, yielding
\begin{align}
    r_{a}=\rho_{a}=\Delta_a, &&r_w=\rho_w=\frac{\Delta_w \Delta_a }{d}\Tr\Sigma,  && \hat{r}_{a,w}=0 
    \\ q_{a,w}=m_{a,w}, &&\hat{q}_{a,w}=\hat{m}_{a,w},
\end{align}
causing the free energy to simplify to 

\begin{align}
\label{eq:f_BO_1}
    f=\underset{q_w\hat{q}_w}{\mathrm{extr}} &\left[\frac{1}{2}q_w\hat{q}_w-\frac{1}{2d}\Tr\left[
    \frac{\Delta_w \Delta_a \hat{q}^2_w\Sigma^2 +\hat{q}_w\Sigma}{\hat{q}_w\Sigma+\frac{1}{\Delta_w \Delta_a }I_d}
    \right]+\frac{1}{2d}\ln\det\left[
    \Delta_w  \Delta_a \hat{q}_w\Sigma+I_d
    \right]
    +\frac{\alpha}{2}\ln \left(
    \Delta+\kappa_1^2(\rho_w-q_w)+\kappa_*^2\rho_a
    \right)\right]
\end{align}
i.e.
\begin{align}
\label{eq:f_BO}
    f=\underset{q_w\hat{q}_w}{\mathrm{extr}}&\left[\frac{1}{2}q_w\hat{q}_w
    -\frac{1}{2}\hat{q}_w\rho_w+\frac{1}{2d}\ln\det\left[
    \Delta_w  \Delta_a \hat{q}_w\Sigma+I_d
    \right]
    +\frac{\alpha}{2}\ln \left(
    \Delta+\kappa_1^2(\rho_w-q_w)+\kappa_*^2\rho_a
    \right)\right]
\end{align}

\subsection{Bayes-optimal saddle points}
The extremization in \eqref{eq:f_BO} can be carried out by imposing that the gradient with respect to each optimization parameter be vanishing, yielding
\begin{align}
\label{eq:BO_SP}
    \begin{cases}
        \hat{q}_w=\frac{\alpha\kappa_1^2}{\Delta+\kappa_1^2(\rho_w-q_w)+\kappa_*^2\rho_a}\\
         q_w=\frac{1}{d}\Tr[\frac{\Delta_w^2\Delta_a^2\hat{q}_w\Sigma^2}{I_d+\Delta_w \Delta_a \hat{q}_w\Sigma}]
    \end{cases}.
\end{align}
The equation \eqref{eq:BO_SP} can be recovered by plugging the first line into the second, and remembering that the excess prediction error can be evaluated as
\begin{equation}
    \epsilon_g-\Delta=\kappa_1^2(\rho_w-q_w)+\kappa_*^2\rho_a.
\end{equation}
Indeed, the Bayes optimal estimator is given by the averaged output of the student network over the posterior measure \eqref{eq:Bayes_post_classif}
$$
\left\langle \frac{1}{\sqrt{k}}a^\top \sigma\left(\frac{1}{\sqrt{d}}W x\right)\right\rangle_{a,W\sim \mathbb{P}}
$$
and the corresponding test error reads
\begin{align}
    \epsilon_g&=\mathbb
{E}_{a_\star,W_\star,\mathcal{D}}\mathbb{E}_{x,\xi}\left( 
\frac{1}{\sqrt{k}}a_\star^\top \sigma\left(\frac{1}{\sqrt{d}}W_\star x\right)+\sqrt{\Delta}\xi-\left\langle \frac{1}{\sqrt{k}}a^\top \sigma\left(\frac{1}{\sqrt{d}}W x\right)\right\rangle \right)^2\notag\\
&={E}_{\mathcal{D}}{E}_{a_\star,W_\star}\frac{1}{k}a_\star^\top \mathbb{E}_{x}\left[\sigma\left(\frac{1}{\sqrt{d}}W_\star x\right)   \sigma\left(\frac{1}{\sqrt{d}}W_\star x\right)   ^\top   \right] a_\star \notag\\
&\qquad+
{E}_{\mathcal{D}}{E}_{a_\star,W_\star}\frac{1}{k}\left\langle a^\top \mathbb{E}_{x}\left[\sigma\left(\frac{1}{\sqrt{d}}W x\right)   \sigma\left(\frac{1}{\sqrt{d}}W^\prime x\right)   ^\top   \right] a^\prime\right\rangle_{a,a^\prime,W,W^\prime \overset{iid}{\sim} \mathbb{P}}\notag\\
&\qquad -2{E}_{\mathcal{D}}{E}_{a_\star,W_\star}\frac{1}{k}\left\langle a_\star^\top \mathbb{E}_{x}\left[\sigma\left(\frac{1}{\sqrt{d}}W_\star x\right)   \sigma\left(\frac{1}{\sqrt{d}}W x\right)   ^\top   \right] a^\prime\right\rangle +\Delta\notag\\
&={E}_{\mathcal{D}}{E}_{a_\star,W_\star}\frac{1}{k}a_\star^\top \mathbb{E}_{x}\left[ \kappa_1^2\frac{W_\star \Sigma W_\star^\top}{d}  +\kappa_*^2 I_d \right] a_\star \notag\\
&\qquad+
{E}_{\mathcal{D}}{E}_{a_\star,W_\star}\frac{1}{k}\left\langle a^\top \mathbb{E}_{x}\left[\kappa_1^2\frac{W \Sigma W^{\prime\top}}{d}  \right] a^\prime\right\rangle_{a,a^\prime,W,W^\prime \overset{iid}{\sim} \mathbb{P}}\notag\\
&\qquad -2{E}_{\mathcal{D}}{E}_{a_\star,W_\star}\frac{1}{k}\left\langle a_\star^\top \mathbb{E}_{x}\left[\kappa_1^2\frac{W_\star \Sigma W^\top}{d}   \right] a^\prime\right\rangle +\Delta\notag\\
&=\kappa_1^2\rho_w+\kappa_*^2\rho_a+\kappa_1^2 q_w-2\kappa_1^2m_w+\Delta=\kappa_1^2(\rho_w-q_w)+\kappa_*^2 \rho_a.
\end{align}
In going from the second line to the third, we used the covariance identities \eqref{eq:Omega_Psi_multilayer}.

%% file: Appendix/Multilayers.tex
\subsection{Generalization to multi-layer nets}

In this section we generalize the discussion of Appendix \ref{App:2layer} to deep networks with $L\ge 2$ hidden layers. We use the same notations as the main text \eqref{eq:teacher}\eqref{eq:student}. To treat simultaneously regression and classification, it is useful to introduce generic notations for the priors
\begin{align}
    P_w(W_\ell)\propto e^{-\frac{1}{2\Delta_\ell}||W_\ell||_F^2},&&
    P_a(a)\propto e^{-\frac{1}{2\Delta_a}||a||^2},
\end{align}
and the posteriors 
\begin{align}
    P_{\mathrm{out}}(y|\hat{y})\propto e^{-\frac{
1}{2\Delta}\left(y-\hat{y}\right)^2
}
\end{align}
for regression. In Appendix \ref{App:Classification}, for classification, the output channel will be 
\begin{equation}
     P_{\mathrm{out}}(y|\hat{y})\propto\int \frac{d\xi e^{-\frac{1}{2\Delta}\xi^2}}{\sqrt{2\pi\Delta}}\Theta\left(
y\times\mathrm{sign}(\hat{y}+\xi)
\right).
\end{equation}
With a slight abuse of notations, we omit for readability the layer index $\ell$ in the prior $P_w$, with the correct variance $\Delta_\ell$ of the Gaussian distribution being implied by the argument $W_\ell$ thereof. Besides, we keep the notation $P_w$ also for the marginal distribution of the rows of the weight matrices.

\subsection{Tower of order parameters}

\paragraph{Replicated partition function}
By the same token, we employ the replica trick, and write the replicated multi-layer partition function as 
\begin{align}
     Z^s=\int \prod\limits_{a=1}^s\prod\limits_{\ell=1}^L dW_\ell^aP_w(W_\ell^a)da^aP_a(a^a)
    \underbrace{ 
    \mathbb{E}_{\{x^\mu\}_\mu}\int \prod\limits_{\mu=1}^nP_{\mathrm{out}}^\star\left(y^\mu|\hat{y}^a_\star(x^\mu)\right)\prod\limits_{\mu=1}^n\prod_{a=1}^s
    P_{\mathrm{out}}\left(y^\mu|\hat{y}^a(x^\mu)\right)
    }_{*}.
\end{align}
Carrying out the expectation with respect to the train set $\mathcal{D}$ :
\begin{align}
    *=\prod_{\mu}\left[
    \mathbb{E}_{\lambda_\star,\{\lambda_a\}_{a=1}^s}\int dy P_{\mathrm{out}}(y|\lambda_\star)\prod\limits_{a=1}^sP_{\mathrm{out}}(y |\lambda_a)
    \right].
\end{align}
We have introduced the local fields
\begin{align}
    \lambda_\star\equiv \frac{1}{\sqrt{k_{L^\star}}}a_\star^\top  \underbrace{\left(
    \varphi^{\star}_{L^\star}\circ \varphi^{\star}_{L^\star-1}\circ \dots\circ \varphi^{\star}_2\circ\varphi_1^{\star}\right)}_{L^\star}
    (x),
    &&\lambda_a=\frac{1}{\sqrt{k_{L}}}a^\top  \underbrace{\left(
    \varphi^a_{L}\circ \varphi^a_{L-1}\circ \dots\circ \varphi^a_2\circ\varphi^a_1\right)}_{L}
    (x) .
\end{align}
$\varphi_\ell^{a}$ denotes the $a-$th replica of the $\ell-$th layer, i.e.
\begin{equation}
    \varphi^{a}_\ell(x)=\sigma_\ell\left(
    \frac{1}{\sqrt{k_{\ell-1}}}W^a_\ell \cdot x
    \right)
\end{equation}

\paragraph{Local fields statistics}
The second order statistics are given by an application of the deep Bayes GEP \eqref{prop:BGP_ML} under the non-specialization assumption as
\begin{align}
\label{eq:App:Multi:appli_GEP}
    \langle \lambda_a\lambda_b\rangle
    =&\prod\limits_{\ell=1}^L\left(\kappa_1^{(\ell)}\right)^2\frac{a_a^\top\left(\prod\limits_{\ell=L}^1 W^a_\ell\right) \Sigma \left(\prod\limits_{\ell=1}^L W^{b\top}_\ell\right)a_b}{\prod\limits_{\ell=0}^{L}k_\ell}\notag\\
    &+\delta_{ab}\left[\sum\limits_{\ell_0=1}^{L-1}\left(\kappa_*^{(\ell_0)}\right)^2\prod\limits_{\ell=\ell_0+1}^L\left(\kappa_1^{(\ell)}\right)^2\frac{a_a^\top\left(\prod\limits_{\ell=L}^{\ell_0+1} W^a_\ell\right) \Sigma \left(\prod\limits_{\ell=\ell_0+1}^L W^{b\top}_\ell\right)a_b}{\prod\limits_{\ell=\ell_0}^{L}k_\ell}
    +\left(\kappa_*^{(L)}\right)^2\frac{a_a^\top a_b}{k_L}
    \right],
\end{align}
and 
\begin{align}
    \langle \lambda_a\lambda_\star\rangle
    =&\prod\limits_{\ell=1}^L\kappa_1^{(\ell)}\prod\limits_{\ell=1}^{L^\star}\kappa_1^{(\ell)\star}\frac{a_a^\top\left(\prod\limits_{\ell=L}^1 W^a_\ell\right) \Sigma \left(\prod\limits_{\ell=1}^{L^\star} W^{\star\top}_\ell\right)a_\star}{\prod\limits_{\ell=0}^{L}\sqrt{k_\ell}\prod\limits_{\ell=0}^{L}\sqrt{k_\ell}^\star}.
\end{align}
By the same token the teacher variance reads
\begin{align}
    \langle \lambda_\star^2\rangle
    =&\prod\limits_{\ell=1}^{L^\star}\left(\kappa_1^{\star(\ell)}\right)^2\frac{a_\star^\top\left(\prod\limits_{\ell=L^\star}^1 W^\star_\ell\right) \Sigma \left(\prod\limits_{\ell=1}^{L^\star} W^{\star\top}_\ell\right)a_\star}{\prod\limits_{\ell=0}^{L}k^\star_\ell}\notag\\
    &+\sum\limits_{\ell_0=1}^{L^\star-1}\left(\kappa_*^{\star(\ell_0)}\right)^2\prod\limits_{\ell=\ell_0+1}^{L^\star}\left(\kappa_1^{\star(\ell)}\right)^2\frac{a_\star^\top\left(\prod\limits_{\ell=L^\star}^{\ell_0+1} W^\star_\ell\right) \Sigma \left(\prod\limits_{\ell=\ell_0+1}^{L^\star} W^{\star\top}_\ell\right)a_\star}{\prod\limits_{\ell=\ell_0}^{L^\star}k^\star_\ell}
    +\left(\kappa_*^{\star(L)}\right)^2\frac{a_\star^\top a_\star}{k_{L^\star}^\star}
\end{align}
These statistics warrant the introduction of the order parameters $m_a,\{\left(q^L_\ell\right)_{ab}\}_{\ell=1}^{L},q^a_{ab},\{\rho^{L^\star}_\ell\}_{\ell=1}^{L^\star},\rho_a$ as
\begin{align}
    \begin{cases}
    m_a=\frac{a_a^\top\left(\prod\limits_{\ell=L}^1 W^a_\ell\right) \Sigma \left(\prod\limits_{\ell=1}^{L^\star} W^{\star\top}_\ell\right)a_\star}{\prod\limits_{\ell=0}^{L}\sqrt{k_\ell}\prod\limits_{\ell=0}^{L}\sqrt{k_\ell}^\star}\\
   \left(q^L_{1}\right)_{ab}=\frac{a_a^\top\left(\prod\limits_{\ell=L}^{\ell_0} W^a_\ell\right) \Sigma \left(\prod\limits_{\ell=\ell_0}^L W^{b\top}_\ell\right)a_b}{\prod\limits_{\ell=\ell_0-1}^{L}k_\ell}\\
   \forall \ell_0\ge 2,~\left(q^L_{\ell_0}\right)_{ab}=\delta_{ab}\frac{a_a^\top\left(\prod\limits_{\ell=L}^{\ell_0} W^a_\ell\right) \Sigma \left(\prod\limits_{\ell=\ell_0}^L W^{b\top}_\ell\right)a_b}{\prod\limits_{\ell=\ell_0-1}^{L}k_\ell}\\
    q^a_{ab}=\delta_{ab}\frac{a_a^\top a_b}{k_L}
    \end{cases}
    &&
    \begin{cases}
    \rho^{L^\star}_{\ell_0}=\frac{a_\star^\top\left(\prod\limits_{\ell=L^\star}^{\ell_0} W^\star_\ell\right) \Sigma \left(\prod\limits_{\ell=\ell_0}^{L^\star} W^{\star\top}_\ell\right)a_\star}{\prod\limits_{\ell=\ell_0-1}^{L^\star}k^\star_\ell}\\
    \rho_a=\frac{a_\star^\top a_\star}{k_{L^\star}^\star}
    \end{cases}
\end{align}
Note that $q^L_{\ell_0}$s corresponds to the overlap between the products of weights from layer $\ell_0$ (included) to layer $L$ (included, i.e. the last layer before readout), with similar intuition holding for the $\rho^{L^\star}_{\ell_0}$s.

\subsection{Replica Symmetry}
Similarly to the two-layers case, we assume replica symmetry. The order parameters are taken of the form
\begin{align}
    &\left(\hat{q}^L_{1}\right)_{ab}=\delta_{ab}\left(-\frac{\hat{r}^L_1}{2}-\hat{q}^L_1\right)+\hat{q}^L_1,\\
    &\forall \ell\ge 2, ~\left(\hat{q}^L_{\ell}\right)_{ab}=-\delta_{ab}\frac{\hat{r}^L_\ell}{2},\\
    &\hat{q}^a_{ab}=-\delta_{ab}\frac{\hat{r}_a}{2},  \\
    &\hat{m}_a=\hat{m},
\end{align}
while the non-hatted overlap read
\begin{align}
&\left(q^L_1\right)_{ab}=\delta_{ab}\left(r^L_1-q^L_1\right)+q^L_1,\\
    &\forall \ell\ge 2, ~\left(q^L_\ell\right)_{ab}=\delta_{ab}r^L_\ell,\\
    &q^a_{ab}=\delta_{ab}r_a,\\
    & m_a=m.\\
\end{align}

\subsection{Multilayer Loss potential}
The computation of the loss potential $\Psi_y$ follows rather tightly the one detailed in \cite{Loureiro2021CapturingTL} and leads to
\begin{align}
    \Psi_y=\int D\xi \int dy&\mathcal{Z}_\star\scriptstyle\left( 
    y,\frac{\prod\limits_{\ell=1}^L\kappa_1^{(\ell)}\prod\limits_{\ell=1}^{L^\star}\kappa_1^{\star(\ell)} m}{\sqrt{\prod\limits_{\ell=1}^L\left(\kappa_1^{(\ell)}\right)^2q^L_1}}\xi,\prod\limits_{\ell=1}^{L^\star}\left(\kappa_1^{\star(\ell)}\right)^2\rho^L_1+
    \sum\limits_{\ell_0=1}^{L^\star-1}\kappa_*^{\star(\ell_0)}\prod\limits_{\ell=\ell_0+1}^{L^\star}\left(\kappa_1^{\star(\ell)}\right)^2\rho^{L^\star}_{\ell_0+1}+\left(\kappa_*^{\star(L^\star)}\right)^2\rho_a - \frac{\left(\prod\limits_{\ell=1}^L\kappa_1^{(\ell)}\prod\limits_{\ell=1}^{L^\star}\kappa_1^{\star(\ell)} m\right)^2}{\prod\limits_{\ell=1}^L\left(\kappa_1^{(\ell)}\right)^2q^L_1}
    \right)\notag\\
    &\times\ln \mathcal{Z}_\ell\scriptstyle\left(y,\sqrt{\prod\limits_{\ell=1}^L\left(\kappa_1^{(\ell)}\right)^2q^L_1}\xi,\prod\limits_{\ell=1}^L\left(\kappa_1^{(\ell)}\right)^2V^L_1+\sum\limits_{\ell_0=1}^{L-1}\kappa_*^{(\ell_0)}\prod\limits_{\ell=\ell_0+1}^L\left(\kappa_1^{(\ell)}\right)^2r^L_{\ell_0+1}+\left(\kappa_*^{L}\right)^2r_a
    \right).
\end{align}

\subsection{Layer-wise entropy}
The computation of the entropy potential $\Psi_w$ follows the same lines as for the two-layer case, see Appendix \ref{App:2layer}. In practice, one has to first integrate the first layer, then the middle layers and the readout. For clarity purposes, we decompose $\Psi_w$ into the following series of terms
\begin{equation}
    \Psi_w=\Psi_1(\hat{q}^L_1,\hat{r}^L_1,q^L_2)+\sum\limits_{\ell=2}^L\Psi_\ell(\hat{r}^L_\ell,q^L_{\ell+1})+\Psi_a(\hat{r}_a)
\end{equation}
and proceed to evaluate each sequentially.
\paragraph{First layer} We focus on $\ell=1$ first:
\begin{align}
    e^{sd\Psi_1}=\prod\limits_{j=1}^d
    \int \prod\limits_{a=1}^s dw^1_aP_w(w^1_a)e^{\sum\limits_{a\le b}\sigma_j\left(\hat{q}^L_1\right)_{ab}\frac{\left(w_a^{1\top}\prod\limits_{\ell=2}^LW_a^\ell a_a\right)\times\left(w_b^{1\top}\prod\limits_{\ell=2}^LW_b^\ell a_b\right) }{\prod\limits_{\ell=1}^Lk_\ell}+\sum\limits_a \sigma_j\hat{m}\frac{\left(w_a^{1\top}\prod\limits_{\ell=2}^LW_a^\ell a_a\right)\times\left(w_\star^{1\top}\prod\limits_{\ell=2}^{L^\star}W_\star^\ell a_\star\right)}{\prod\limits_{\ell=1}^L\sqrt{k_\ell}\prod\limits_{\ell=1}^{L^\star}\sqrt{k^\star_\ell}}}.
\end{align}
Introducing the local fields
\begin{equation}
    \eta_a\equiv \frac{w_a^{1\top}\prod\limits_{\ell=2}^LW_a^\ell a_a}{\prod\limits_{\ell=1}^L\sqrt{k_\ell}},
\end{equation}
which inherit from $P_w(\cdot)$ the statistics
\begin{align}
    \langle \eta_a\eta_b\rangle=\delta_{ab}\frac{\left(q^L_2\right)_{aa}}{\Delta_1^{-1}},
\end{align}
it follows that
\begin{align}
    e^{sd\Psi_1}=\prod\limits_{j=1}^d e^{\frac{1}{2}\sigma_j^2\left(\frac{a_\star\top \prod\limits_{\ell=L^\star}^2 W_\ell^{\star\top} (w_1^\star)_j}{\prod \limits_{\ell=1}^{L^\star}\sqrt{k_\ell^\star}}\right)^2 \hat{m}^\top \left[
\mathrm{diag}\left(\frac{\Delta_1^{-1}}{\left(q^L_2\right)_{aa}}\right) -\sigma_j \times (2I_s)\odot \hat{Q}^L_1 
\right]^{-1}\hat{m}
-\frac{1}{2}\ln \det \left[
I_s-\sigma_j\mathrm{diag}\left( \frac{\Delta_1^{-1}}{\left(q^L_2\right)_{aa}}\right)(2I_s)\odot \hat{Q}^L_1
\right]
}.
\end{align}
Finally,
\begin{align}
    \Psi_1=\frac{1}{d}\Tr[
    \frac{\hat{q}^L_1\Sigma +\hat{m}^2\Sigma^2 \frac{\prod\limits_{\ell=L^\star}^1W_\ell^{\star\top} a_\star a_\star^\top\prod\limits_{\ell=1}^{L^\star}W_\ell^{\star}}{\prod\limits_{\ell=1}^{L^\star} k^\star_\ell}}{\hat{V}^L_1\Sigma+\frac{\Delta_1^{-1}}{r^L_2}}
    ]
    -\frac{1}{2d}\ln\det[\frac{r^L_2}{\Delta_1^{-1}}\hat{V}^L_1\Sigma+I_d]
\end{align}

\paragraph{Middle layers} We know provide the derivation for $\ell\ge 2$ (middle layers). The corresponding potential reads
\begin{align}
    e^{sd\Psi_{\ell_0}}&=\int \prod\limits_{a=1}^sdW^a_{\ell_0} P_w(W^a_{\ell_0})e^{\sum\limits_{a=1}^s\left(\hat{q}^L_\ell\right)_{aa}\frac{a_a^\top \prod\limits_{\ell=L}^{\ell_0}W_\ell^{a\top}\prod\limits_{\ell=\ell_0}^{L}W_\ell^{a} a_a}{\prod\limits_{\ell=\ell_0}^Lk_\ell}}\\
    &=\prod\limits_{a=1}^s\left[
    \int dw e^{-\frac{\Delta_{\ell_0}^{-1}}{2}w^\top w -\frac{1}{2}\hat{r}^L_{\ell_0}\left(
    \frac{w^\top\prod\limits_{\ell=\ell_0+1}^LW_\ell a_a}{\prod\limits_{\ell=\ell_0}^L\sqrt{k_\ell}}
    \right)^2  }
    \right]^{k_{\ell_0-1}}.
\end{align}
Introducing the local fields 
\begin{align}
    \eta_a=\frac{w^\top\prod\limits_{\ell=\ell_0+1}^LW_\ell a_a}{\prod\limits_{\ell=\ell_0}^L\sqrt{k_\ell}},
\end{align}
it follows that
\begin{equation}
    \Psi_\ell=- \frac{\gamma_{\ell-1}}{2}\ln\left(1+\frac{r^L_{\ell+1}\hat{r}^L_{\ell}}{\Delta_\ell^{-1}}\right).
\end{equation}

\paragraph{Readout} The readout layer entropy can be computed as in the two layer case yielding
\begin{equation}
    \Psi_a=-\frac{\gamma_L}{2}\ln\left(
    1+\frac{\hat{r}_a}{\Delta_a^{-1}}
    \right),
\end{equation}
see Appendix \ref{App:2layer}

\paragraph{Summary: multilayer entropic potential}
Assembling these results, the total entropic action reads
\begin{align}
    \Psi_w=&-\frac{\gamma_L}{2}\ln\left(
    1+\frac{\hat{r}_a}{\Delta_a^{-1}}
    \right)-\sum\limits_{\ell=2}^L\frac{\gamma_{\ell-1}}{2}\ln\left(1+\frac{r^L_{\ell+1}\hat{r}^L_{\ell}}{\Delta_\ell^{-1}}\right)+\frac{1}{d}\Tr[{\scriptstyle
    \frac{\hat{q}^L_1\Sigma +\hat{m}^2\Sigma^2 \frac{\prod\limits_{\ell=L^\star}^1W_\ell^{\star\top} a_\star a_\star^\top\prod\limits_{\ell=1}^{L^\star}W_\ell^{\star}}{\prod\limits_{\ell=1}^{L^\star} k^\star_\ell}}{\hat{V}^L_1\Sigma+\frac{\Delta_1^{-1}}{r^L_2}}
    }
    ]
    \notag\\
    &
    -\frac{1}{2d}\ln\det[\frac{r^L_2}{\Delta_1^{-1}}\hat{V}^L_1\Sigma+I_d].
\end{align}

\subsection{Finite T multilayer free energy}
Finally, the complete free energy at finite temperature reads
\begin{align}
    f=&\underset{q_1^L,\hat{q}_1^L,V_1^L,\hat{V}_1^L,m,\hat{m},\{r_\ell^L,\hat{r}_\ell^L\},r_a,\hat{r}_a}{\mathrm{extr}}   -\frac{1}{2}(\hat{V}^L_1V^L_1+\hat{V}^L_1q^L_1-\hat{q}^L_1V^L_1)-\frac{1}{2}\sum\limits_{\ell=2}^L\gamma_{\ell-1}\hat{r}^L_\ell r^L_\ell -\frac{\gamma_L}{2}\hat{r}_ar_a+\hat{m}m+\frac{\gamma_L}{2}\ln\left(
    1+\frac{\hat{r}_a}{\Delta_a^{-1}}
    \right)\notag\\
    &+\sum\limits_{\ell=2}^L\frac{\gamma_{\ell-1}}{2}\ln\left(1+\frac{r^L_{\ell+1}\hat{r}^L_{\ell}}{\Delta_\ell^{-1}}\right)-\frac{1}{d}\Tr[{\scriptstyle
    \frac{\hat{q}^L_1\Sigma +\hat{m}^2\Sigma^2 \frac{\prod\limits_{\ell=L^\star}^1W_\ell^{\star\top} a_\star a_\star^\top\prod\limits_{\ell=1}^{L^\star}W_\ell^{\star}}{\prod\limits_{\ell=1}^{L^\star} k^\star_\ell}}{\hat{V}^L_1\Sigma+\frac{\Delta_1^{-1}}{r^L_2}}
    }
    ]+\frac{1}{2d}\ln\det[\frac{r^L_2}{\Delta_1^{-1}}\hat{V}^L_1\Sigma+I_d]
    \notag
    \\
    &-\alpha
    \int D\xi \int dy\mathcal{Z}_\star\scriptstyle\left( y,\frac{\prod\limits_{\ell=1}^L\kappa_1^{(\ell)}\prod\limits_{\ell=1}^{L^\star}\kappa_1^{\star(\ell)} m}{\sqrt{\prod\limits_{\ell=1}^L\left(\kappa_1^{(\ell)}\right)^2q^L_1}}\xi,\prod\limits_{\ell=1}^{L^\star}\left(\kappa_1^{\star(\ell)}\right)^2\rho^L_1+
    \sum\limits_{\ell_0=1}^{L^\star-1}\kappa_*^{\star(\ell_0)}\prod\limits_{\ell=\ell_0+1}^{L^\star}\left(\kappa_1^{\star(\ell)}\right)^2\rho^{L^\star}_{\ell_0+1}+\left(\kappa_*^{\star(L^\star)}\right)^2\rho_a - \frac{\left(\prod\limits_{\ell=1}^L\kappa_1^{(\ell)}\prod\limits_{\ell=1}^{L^\star}\kappa_1^{\star(\ell)} m\right)^2}{\prod\limits_{\ell=1}^L\left(\kappa_1^{(\ell)}\right)^2q^L_1}
    \right)\notag\\
    &\times\ln \mathcal{Z}_\ell\scriptstyle\left(
y,\sqrt{\prod\limits_{\ell=1}^L\left(\kappa_1^{(\ell)}\right)^2q^L_1}\xi,\prod\limits_{\ell=1}^L\left(\kappa_1^{(\ell)}\right)^2V^L_1+\sum\limits_{\ell_0=1}^{L-1}\kappa_*^{(\ell_0)}\prod\limits_{\ell=\ell_0+1}^L\left(\kappa_1^{(\ell)}\right)^2r^L_{\ell_0+1}+\left(\kappa_*^{L}\right)^2r_a
    \right).
\end{align}

\subsection{Bayes-Optimal setting}

\paragraph{Nishimori identities} For matching teacher/student the Nishimori identities guarantee that
\begin{align}
\begin{cases}
    q^L_1=m\\
    \hat{q}^L_1=\hat{m}
\end{cases}
&&
\forall \ell,~
\begin{cases}
    r^L_\ell=\rho^L_\ell=\Delta_a\prod\limits_{\ell}^L \Delta_\ell \times \left(\frac{\tr\Sigma}{d}\right)^{\delta_{\ell,1}}\\
    \hat{r}^L_\ell=0.
\end{cases}
\end{align}

The Bayes-Optimal free energy accordingly simplifies to
\begin{align}
\label{eq:App:multi:f_Bayes}
    f=&\underset{q,\hat{q}}{\mathrm{extr}} \frac{1}{2}\hat{q}q-\frac{1}{d}\Tr[{\scriptstyle
    \frac{\hat{q}\Sigma +\hat{q}^2\Sigma^2 \Delta_a\prod\limits_{\ell=1}^L\Delta_\ell}{\hat{q}\Sigma+\frac{1}{\Delta_a\prod\limits_{\ell=1}^L\Delta_\ell}}
    }
    ]+\frac{1}{2d}\ln\det[\Delta_a\prod\limits_{\ell=1}^L\Delta_\ell\hat{q}\Sigma+I_d]
    \notag
    \\
    &-\alpha
    \int D\xi \int dy\mathcal{Z}_\star\scriptstyle\left( 
    y,\sqrt{\prod\limits_{\ell=1}^L\left(\kappa_1^{(\ell)}\right)^2q
    }\xi,
    \prod\limits_{\ell=1}^{L}\left(\kappa_1^{(\ell)}\right)^2V+
    \sum\limits_{\ell_0=1}^{L-1}\kappa_*^{(\ell_0)}\prod\limits_{\ell=\ell_0+1}^{L^\star}\left(\kappa_1^{(\ell)}\right)^2\rho^{L}_{\ell_0+1}+\left(\kappa_*^{\star(L)}\right)^2\rho_a 
    \right)\notag\\
    &\times\ln \mathcal{Z}_\ell\scriptstyle\left(
    y,
    \sqrt{\prod\limits_{\ell=1}^L\left(\kappa_1^{(\ell)}\right)^2q}\xi,\prod\limits_{\ell=1}^L\left(\kappa_1^{(\ell)}\right)^2V+\sum\limits_{\ell_0=1}^{L-1}\kappa_*^{(\ell_0)}\prod\limits_{\ell=\ell_0+1}^L\left(\kappa_1^{(\ell)}\right)^2\rho^L_{\ell_0+1}+\left(\kappa_*^{L}\right)^2\rho_a
    \right).
\end{align}
For a linear readout, and a noisy teacher channel with Gaussian additive noise of variance $\Delta$ \eqref{eq:teacher}:
\begin{align}
\label{eq:App:Multi:f_BO}
    f=&\underset{q,\hat{q}}{\mathrm{extr}}\frac{1}{2}\hat{q}q-\frac{1}{d}\Tr[{\scriptstyle
    \frac{\hat{q}\Sigma +\hat{q}^2\Sigma^2 \Delta_a\prod\limits_{\ell=1}^L\Delta_\ell}{\hat{q}\Sigma+\frac{1}{\Delta_a\prod\limits_{\ell=1}^L\Delta_\ell}}
    }
    ]+\frac{1}{2d}\ln\det[\Delta_a\prod\limits_{\ell=1}^L\Delta_\ell\hat{q}\Sigma+I_d]
    \notag
    \\
    &
    +\frac{\alpha}{2}\ln\left(\Delta+
    \prod\limits_{\ell=1}^L\left(\kappa_1^{(\ell)}\right)^2(\rho^L_1-q)+\sum\limits_{\ell_0=1}^{L-1}\kappa_*^{(\ell_0)}\prod\limits_{\ell=\ell_0+1}^L\left(\kappa_1^{(\ell)}\right)^2\rho^L_{\ell_0+1}+\left(\kappa_*^{L}\right)^2\rho_a
    \right)
\end{align}

\subsection{Bayes-Optimal Saddle point equations}
The saddle-point equations corresponding to the extremization \eqref{eq:App:Multi:f_BO} read
\begin{align}
\begin{cases}
    \hat{q}=\alpha\frac{\prod\limits_{\ell=1}^L\left(\kappa_1^{(\ell)}\right)^2}{\Delta+
    \prod\limits_{\ell=1}^L\left(\kappa_1^{(\ell)}\right)^2(\rho^L_1-q)+\sum\limits_{\ell_0=1}^{L-1}\kappa_*^{(\ell_0)}\prod\limits_{\ell=\ell_0+1}^L\left(\kappa_1^{(\ell)}\right)^2\rho^L_{\ell_0+1}+\left(\kappa_*^{L}\right)^2\rho_a}\\
    q=\frac{1}{2d}\Tr[
    \frac{\hat{q}\Sigma^2\prod\limits_{\ell=1}^L \Delta_\ell^2}{1+\hat{q}\Sigma\prod\limits_{\ell=1}^L \Delta_\ell}
    ].
\end{cases}
\end{align}
The Bayes optimal MSE can be computed along the same lines as the two-layers case. Recall that $h_L^\star(x)$ and $h_L(x)$ designate the post activations at the last layer for an input $x$, respectively for the teacher \eqref{eq:teacher} and student \eqref{eq:student} networks. The Bayes optimal estimator is then $\langle a^\top h_L(x)\rangle /\sqrt{k_L} $, where the expectation is carried out over the Bayes posterior measure \eqref{eq:Bayes_post_classif}. The MSE then reads
\begin{align}
    \epsilon_g&=\mathbb{E}_{\mathcal{D},a_\star,\{W_\ell^\star\}_{\ell}}\mathbb{E}_{x,\xi}\left(\frac{1}{\sqrt{k_L}}a_\star^\top h_L^\star(x)+\xi-\left\langle \frac{1}{\sqrt{k_L}}a^\top h_L(x)
    \right\rangle
    \right)^2\notag\\
    &=\mathbb{E}_{\mathcal{D},a_\star,\{W_\ell^\star\}_{\ell}}\left[{\scriptstyle
    \frac{1}{k_L} a_\star^\top \mathbb{E}_x\left[
    h_L^\star(x)h_L^{\star\top}(x)
    \right]a_\star+\frac{1}{k_L}\left\langle
a^\top \mathbb{E}_x\left[ 
h_L(x)h_L^\prime(x)^\top
\right] a^\prime
    \right\rangle_{a,\{W_\ell\},a^\prime,\{W^\prime_\ell\}\overset{i.i.d}{\sim}\mathbb{P}}
    -2 \frac{1}{k_L} \left\langle a_\star^\top \mathbb{E}_x\left[
    h_L^\star(x)h_L^{\top}(x)
    \right]a \right\rangle}
    \right]+\Delta \notag\\
&=\frac{1}{k_L}\mathbb{E}_{\mathcal{D},a_\star,\{W_\ell^\star\}_{\ell}}\Bigg[\prod\limits_{\ell=1}^L\left(\kappa_1^{(\ell)}\right)^2
a_\star^\top  \left(\prod\limits_{\ell=1}^L \frac{W_\ell^{\star\top}}{\sqrt{k_{\ell-1}}}\right)^\top \Sigma \left(\prod\limits_{\ell=1}^L \frac{W_\ell^{\star}}{\sqrt{k_{\ell-1}}}\right)a_\star\notag\\&\qquad
+\sum\limits_{\ell_0=1}^{L-1}\left(\kappa_*^{(\ell_0)}\right)^2\prod\limits_{\ell=\ell_0+1}^L\left(\kappa_1^{(\ell)}\right)^2 a_\star \left(\prod\limits_{\ell=\ell_0+1}^L \frac{W_\ell^{\star\top}}{\sqrt{k_{\ell-1}}}\right)^\top \left(\prod\limits_{\ell=\ell_0+1}^L \frac{W_\ell^{\star}}{\sqrt{k_{\ell-1}}}\right)a_\star
\Bigg]\notag\\
&~~~~+\frac{1}{k_L}\mathbb{E}_{\mathcal{D},a_\star,\{W_\ell^\star\}_{\ell}}\left[\left\langle
\prod\limits_{\ell=1}^L\left(\kappa_1^{(\ell)}\right)^2
a^\top  \left(\prod\limits_{\ell=1}^L \frac{W_\ell^{\top}}{\sqrt{k_{\ell-1}}}\right)^\top \Sigma\left(\prod\limits_{\ell=1}^L \frac{W_\ell^{\prime\top}}{\sqrt{k_{\ell-1}}}\right)a^\prime\right\rangle_{a,\{W_\ell\},a^\prime,\{W^\prime_\ell\}\overset{i.i.d}{\sim}\mathbb{P}}
\right]\notag\\
&~~~~-2\frac{1}{k_L}\mathbb{E}_{\mathcal{D},a_\star,\{W_\ell^\star\}_{\ell}}\left[\left\langle
\prod\limits_{\ell=1}^L\left(\kappa_1^{(\ell)}\right)^2
a_\star^\top  \left(\prod\limits_{\ell=1}^L \frac{W_\ell^{\star\top}}{\sqrt{k_{\ell-1}}}\right)^\top\Sigma \left(\prod\limits_{\ell=1}^L \frac{W_\ell}{\sqrt{k_{\ell-1}}}\right)a\right\rangle\right]+\Delta\notag\\
&=\prod\limits_{\ell=1}^L\left(\kappa_1^{(\ell)}\right)^2(\rho^L_1-q)+\sum\limits_{\ell_0=1}^{L-1}\left(\kappa_*^{(\ell_0)}\right)^2\prod\limits_{\ell=\ell_0+1}^L\left(\kappa_1^{(\ell)}\right)^2\rho^L_{\ell_0+1}+\left(\kappa_*^{L}\right)^2\rho_a
\end{align}
Replacing the $\rho_\ell^L$ order parameters by their definitions, one finally reaches
\begin{equation}
   \epsilon_g^{\mathrm{Bayes}}-\Delta=\prod\limits_{\ell=1}^L\left(\kappa_1^{(\ell)}\right)^2\left(\Delta_a\left(\int 
 z\dd \mu(z)\right)\prod\limits_{\ell=1}^L \Delta_\ell-q\right)\notag\\
    +\sum\limits_{\ell_0=1}^{L-1}\left(\kappa_*^{(\ell_0)}\right)^2\Delta_a\prod\limits_{\ell=\ell_0+1}^L\left(\kappa_1^{(\ell)}\right)^2\Delta_\ell+\left(\kappa_*^{(L)}\right)^2\Delta_a
\end{equation}
which is equation \eqref{eq:Bayes_regression}.

%% file: Appendix/Shallow.tex
In this appendix, we show that the shallow network \eqref{eq:linear_model}
\begin{equation}
    y^{\mathrm{lin}}(x)=f_\star\left(
    \frac{\sqrt{\rho} \theta^\top x}{\sqrt{d}}  +\sqrt{\epsilon_r} \xi
    \right),
\end{equation}
possesses the same Bayes free energy as the deep network \eqref{eq:teacher}, implying in particular that they are characterized by the same Bayes error. Again, let us defined the notations
\begin{align}
    P_w(w)\propto e^{-\frac{1}{2}||w||^2}
\end{align}
and the posterior
\begin{align}
    P^\prime_{\mathrm{out}}(y|z)\propto \int \frac{d\xi}{\sqrt{2\pi \epsilon_r}}e^{-\frac{1}{2\epsilon_r}\xi^2}\delta(y-f_\star(z+\xi))
\end{align}
. We will again resort to the replica trick. While this exact replica computation has not, to the authors' knowledge, been detailed in previous work, it is nevertheless very standard and close to existing studies, e.g. \cite{Barbier2017OptimalEA, Loureiro2021CapturingTL}. For these reasons, we shall frequently point to related computations and be concise.

The replicated partition function reads
\begin{align}
    Z^s=\int \prod\limits_{a=1}^s dw_a P_w(w_a) d\theta P_w(\theta) \prod\limits_{\mu=1}^n \mathbb{E}_{y^\mu, x^\mu} \int dy^\mu
    P^\prime_{\mathrm{out}}\left(
    y^\mu |\frac{\theta^\top x^\mu}{\sqrt{d}}
    \right)    \prod\limits_{a=1}^sP^\prime_{\mathrm{out}}\left(
    y^\mu |\frac{w_a^\top x^\mu}{\sqrt{d}}
    \right)
\end{align}
Define the overlap matrix
\begin{align}
    q_{ab}=\frac{w_a^\top \Sigma w_b}{d}&&
    m_a=\frac{\theta^\top\Sigma w_a}{d}
\end{align}
and the associated Dirac conjugates $\{\hat{q}_{ab}\},\{\hat{m}_a\}$, the partition function becomes
\begin{align}
    Z^s=\int \prod\limits_{a\le b}dq_{ab}d\hat{q}_{ab}\prod\limits_{a=1}^s dm_ad\hat{m}_{a} e^{-\sum\limits_{a\le b} d q_{ab}\hat{q}_{ab}-\sum\limits_{a} dm_a\hat{m}_a}e^{d s\Psi_w+\alpha ds\Psi_y} 
\end{align}
where we introduced the entropic potential $\Psi_w$
\begin{align}
    e^{ds\Psi_w}=\int \prod\limits_{a=1}^s dw_a P_w(w_a) d\theta P_w(\theta) e^{\sum\limits_{a\le b}\hat{q}_{ab}w_a^\top \Sigma w_b +\sum \limits_{a}\hat{m}_a\theta^\top \Sigma w_a}
\end{align}
and the loss potential $\Psi_y$
\begin{align}
    e^{\alpha ds \Psi_y}=\mathbb{E}_{\{\lambda_a,\mu}\int dy P^\prime_{\mathrm{out}}(y|\mu)\prod\limits_{a=1}^sP^\prime_{\mathrm{out}}(y|\lambda_a)
\end{align}
where the local fields $\{\lambda_a\}, \mu$ are Gaussian with statistics
\begin{align}
    \langle \mu^2\rangle=1,&&
    \langle \mu \lambda_a\rangle=m_a,&&
    \langle \lambda_a\lambda_b\rangle=q_{ab}.
\end{align}
We assume the replica symmetric ansatz with the Nishimori identities
\begin{align}
    &q_{ab}=(\rho-q)\delta_{ab}+q\\
    &m_a=q\\
    &\hat{m}_a=\hat{q}\\
    &\hat{q}_{ab}=\hat{q}(1-\delta_{ab})   
\end{align}

The computation of $\Psi_y$ is standard and leads to (see e.g. \cite{Loureiro2021CapturingTL})
\begin{align}
    \Psi_y=\int D\xi \int dy\mathcal{Z}^\prime\left(
    y,\sqrt{q}\xi,\rho\int zd\mu(z)-q
    \right)\ln \mathcal{Z}^\prime\left(
    y,\sqrt{q}\xi,\rho\int zd\mu(z)-q
    \right)
\end{align}
with
\begin{align}
    \mathcal{Z}^\prime(y,\omega,V)=\mathbb{E}_x^{\mathcal{N}(\omega,V)}P^\prime_{\mathrm{out}}(y|x)&=\int DxD\xi \delta(y-f_\star(\sqrt{\epsilon_r}\xi+\sqrt{V}x+\omega))\notag\\
    &=\int DxD\xi\delta(y-f_\star(\sqrt{\Delta}\xi+\sqrt{V+\epsilon_r-\Delta}x+\omega))\notag\\
    &=\mathcal{Z}_\ell(y,\omega,V+\epsilon_r-\Delta)
\end{align}
$\mathcal{Z}_\ell$ \eqref{eq:intro_Z} has been introduced in Appendix \ref{App:multilayer}.
As for the entropic potential, introducing the Hubbard Stratonovitch field $\eta$, and using the notation $D$ for integrands with normal distributions,
\begin{align}
    e^{sd\Psi_w}&=\int D\theta D\eta \left[
    \int dw \frac{1}{(2\pi)^{\frac{d}{2}}}e^{-\frac{1+ \rho\hat{q}\Sigma}{2}w^\top w+\rho\hat{q}\theta^\top \Sigma w+\sqrt{\rho\hat{q}}w^\top \Sigma^{\frac{1}{2}}\eta}
    \right]^s\notag\\
    &=\int D\theta D\eta e^{
    -\frac{s}{2}\ln\det (1+\hat{q}\Sigma)+
    \frac{s}{2}(\sqrt{\rho\hat{q}}\Sigma^{\frac{1}{2}}\eta+ \rho\hat{q}\Sigma \theta)^\top(1+\rho\hat{q}\Sigma)^{-1} (\sqrt{\rho\hat{q}}\Sigma^{\frac{1}{2}}\eta+ \rho\hat{q}\Sigma \theta)
    }\notag\\
    &=e^{-\frac{s}{2}\ln\det (1+\rho\hat{q}\Sigma)}\int \frac{dz}{\sqrt{\det 2\pi (\rho\hat{q}\Sigma+\rho^2\hat{q}^2\Sigma^2)}} e^{-\frac{1}{2}z^\top (\rho\hat{q}\Sigma+\rho^2\hat{q}^2\Sigma^2)^{-1}z+\frac{s}{2}z^\top(1+\rho\hat{q}\Sigma)^{-1} z }\notag\\
    &=e^{-\frac{s}{2}\ln\det (1+\rho\hat{q}\Sigma)-\frac{1}{2}\ln \det\left[\mathbb{I}-s(\rho\hat{q}\Sigma+\rho^2\hat{q}^2\Sigma^2)(1+\rho\hat{q}\Sigma)^{-1}\right]}.
\end{align}
Finally
\begin{align}
    \Psi_w=\frac{1}{2}\ln\det (1+\rho\hat{q}\Sigma)-\frac{1}{2}\Tr[(\rho\hat{q}\Sigma+\rho^2\hat{q}^2\Sigma^2)(1+\rho\hat{q}\Sigma)^{-1}]
\end{align}
Putting everything together
\begin{align}
    f&=\underset{q,\hat{q}}{\mathrm{extr}} -\frac{1}{2}\hat{q}q+\frac{1}{2d}\ln\det\left[1+\Delta_a\prod\limits_\ell \Delta_\ell \prod\limits_\ell \left(\kappa_1^{(\ell)}\right)^2\hat{q}\Sigma\right]-\frac{1}{2d}\Tr[\frac{\scriptstyle\Delta_a\prod\limits_\ell \Delta_\ell \prod\limits_\ell \left(\kappa_1^{(\ell)}\right)^2\hat{q}\Sigma+\left(\Delta_a\prod\limits_\ell \Delta_\ell \prod\limits_\ell \left(\kappa_1^{(\ell)}\right)^2\right)^2\hat{q}^2\Sigma^2}{1+\Delta_a\prod\limits_\ell \Delta_\ell \prod\limits_\ell \left(\kappa_1^{(\ell)}\right)^2\hat{q}\Sigma}]\notag\\
    &\qquad +\alpha\int D\xi \int dy\mathcal{Z}\left(
    y,\sqrt{q}\xi,\rho\int zd\mu(z)-q
    \right)\ln \mathcal{Z}\left(
    y,\sqrt{q}\xi,\rho\int zd\mu(z)-q
    \right)\notag\\
    &=\underset{q,\hat{q}}{\mathrm{extr}} -\frac{1}{2}\hat{q}q+\frac{1}{2d}\ln\det\left[1+\Delta_a\prod\limits_\ell \Delta_\ell \hat{q}\Sigma\right]-\frac{1}{2d}\Tr[\frac{\scriptstyle \hat{q}\Sigma+\Delta_a\prod\limits_\ell \Delta_\ell \hat{q}^2\Sigma^2}{\frac{1}{\prod\limits_\ell \left(\kappa_1^{(\ell)}\right)^2}+\hat{q}\Sigma}]\notag\\
    &\qquad +\alpha\int D\xi \int dy\mathcal{Z}\left({\scriptstyle
    y,\sqrt{\prod\limits_\ell \left(\kappa_1^{(\ell)}\right)^2q}\xi,\prod\limits_\ell \left(\kappa_1^{(\ell)}\right)^2(r_1^L-q)+\epsilon_r-\Delta}
    \right)\ln \mathcal{Z}_\ell\left({\scriptstyle
    y,\sqrt{\prod\limits_\ell \left(\kappa_1^{(\ell)}\right)^2q}\xi,\prod\limits_\ell \left(\kappa_1^{(\ell)}\right)^2(r_1^L-q)+\epsilon_r-\Delta}
    \right)\notag\\
    &=\underset{q,\hat{q}}{\mathrm{extr}} -\frac{1}{2}\hat{q}q+\frac{1}{2d}\ln\det\left[1+\Delta_a\prod\limits_\ell \Delta_\ell \hat{q}\Sigma\right]-\frac{1}{2d}\Tr[\frac{\scriptstyle \hat{q}\Sigma+\Delta_a\prod\limits_\ell \Delta_\ell \hat{q}^2\Sigma^2}{\frac{1}{\prod\limits_\ell \left(\kappa_1^{(\ell)}\right)^2}+\hat{q}\Sigma}]\notag\\
    &\qquad +\alpha\int D\xi \int dy\mathcal{Z}\left({\scriptstyle
    y,\sqrt{\prod\limits_\ell \left(\kappa_1^{(\ell)}\right)^2q}\xi,\prod\limits_\ell \left(\kappa_1^{(\ell)}\right)^2(r_1^L-q)+\sum\limits_{\ell_0=1}^{L-1}\left(\kappa_*^{(\ell_0)}\right)^2\Delta_a\prod\limits_{\ell=\ell_0+1}^L\left(\kappa_1^{(\ell)}\right)^2\Delta_\ell+\left(\kappa_*^{(L)}\right)^2\Delta_a}
    \right)\notag\\
    &\qquad\ln \mathcal{Z}_\ell\left({\scriptstyle
    y,\sqrt{\prod\limits_\ell \left(\kappa_1^{(\ell)}\right)^2q}\xi,\prod\limits_\ell \left(\kappa_1^{(\ell)}\right)^2(r_1^L-q)+\sum\limits_{\ell_0=1}^{L-1}\left(\kappa_*^{(\ell_0)}\right)^2\Delta_a\prod\limits_{\ell=\ell_0+1}^L\left(\kappa_1^{(\ell)}\right)^2\Delta_\ell+\left(\kappa_*^{(L)}\right)^2\Delta_a}
    \right)
\end{align}
which is exactly the Bayes free energy \eqref{eq:App:multi:f_Bayes} for the multi-layer target. The expressions of the Bayes MSE and classification errors for the single layer network are known \cite{Loureiro2021CapturingTL} and exactly recover the expressions \eqref{eq:Bayes_regression} and \eqref{eq:Bayes_classif}.

%% file: Appendix/ERM_G3M.tex
In this appendix, we contrast the Bayes optimal MSE for regression \eqref{eq:Bayes_regression} obtained in Appendix \ref{App:2layer} and \ref{App:multilayer} to the test error achieved by linear ERM methods. We successively describe ridge regression, and random features regression \cite{Rahimi2007RandomFF}, alongside its infinite width limit, kernel regression.  {\color{black} Sharp asymptotics for the test error of those ERM methods were given by the concomitant work of \cite{DRF2023}. In this appendix, we provide a complementary view of these characterization, starting from Theorem $1$ of \cite{Loureiro2021CapturingTL} applied to the equivalent model \eqref{eq:linear_model}. We heuristically show that this route offers an intuitive way to recover the formulae of \cite{DRF2023}. }\\

We start by reviewing the results of \cite{Loureiro2021CapturingTL}. They consider learning from the dataset $\{v^\mu,y^\mu\}_{\mu=1}^n$, where $v^\mu \in\mathbb{R}^p$ represent the student features, assuming that the labels are generated from a generalized linear model as
$$
y^\mu=f_\star\left(
\frac{1}{k_L}a_\star^\top h_L^{\star\mu}
\right)
$$
with $h_L^{\star\mu} \in\mathbb{R}^{k_L}$. Finally, suppose that $h_L^{\star\mu},v^\mu$ are jointly Gaussian
$$
\left(
\begin{array}{c}
     h_L^{\star\mu}\\
     v^\mu
\end{array}
\right)
\sim
\mathcal{N}\left(
0,
\left[
\begin{array}{cc}
    \Psi & \Phi  \\
    \Phi^\top  & \Omega
\end{array}
\right]
\right).
$$
The trains set can be learnt resorting to ERM with loss function $g(\cdot)$
$$
\hat{w}=\underset{w}{\mathrm{argmin}} \sum\limits_{\mu=1}^n
g\left(
y^\mu, \frac{w^\top v^\mu}{\sqrt{p}}
\right)+\frac{\lambda}{2}||w||^2
$$
with the test error being measured as 
$$
\epsilon_g=\mathbb{E}_{\mathcal{D}}\mathbb{E}_{v,h_L^\star}\left[
\hat{g}\left(
\hat{f}\left(\frac{\hat{w}^\top v^\mu}{\sqrt{p}}\right), f_\star\left(
\frac{1}{k_L}a_\star^\top h_L^{\star\mu}
\right)
\right)
\right]
$$
for some metric $\hat{g}$ and estimator function $\hat{f}$. Finally, the proportional limit $p, k_L, n\rightarrow \infty$ with finite ratios $\alpha:=\sfrac{n}{p}, \gamma:=\sfrac{k_L}{p}$. Then Theorem $1$ of \cite{Loureiro2021CapturingTL} states that
$$
\epsilon_g\rightarrow \mathbb{E}_{(\nu,\lambda)}\left[\hat{g}(\hat{f}(\lambda),f_\star(\nu) \right]
$$
where
$$
\left(
\begin{array}{c}
    \nu\\
     \lambda
\end{array}
\right)
\sim
\mathcal{N}\left(
0,
\left[
\begin{array}{cc}
    \frac{a_\star^\top \Psi a_\star}{k_L} & m  \\
    m  & q
\end{array}
\right]
\right).
$$
The \textit{self-overlap} $q$ and the \textit{magnetization} $m$ are two scalar values characterized as the solutions of a system of scalar \textit{saddle-point} equations, depending only on $\Omega, \Phi$, the loss function $g$ and the output channel $f_\star$. \cite{Loureiro2021CapturingTL} provide detailed examples of these saddle-point equations for standard loss functions, which we do not replicate here for conciseness. \\

This provide a very versatile framework to evaluate the sharp asymptotics of the test error of linear ERM methods, provided one crucially assumes that the student features $v$ and the last layer post-activations $h^\star_L$ are jointly Gaussian. \\


\paragraph{Linear regression}
{\color{black} Applying Theorem $1$ from \cite{Loureiro2021CapturingTL} to the equivalent shallow network \eqref{eq:linear_model}} using, in the notations of \cite{Loureiro2021CapturingTL},
\begin{align}
\begin{cases}
     &\rho=\kappa_1^{\star 2}\rho_w+\kappa_*^{\star2}\rho_a+\Delta,\\
    &\Omega=\Sigma\notag\\
    &\Phi=\sqrt{\Delta_a\Delta_w}\kappa_1^\star \Sigma,
\end{cases}   
\end{align}
yields the saddle point equations

\begin{align}
\label{eq:App:ERM:SP_ridge_teacher2l}
\begin{cases}
\hat{V}=\frac{\alpha}{1+V}\\
\hat{q}=\alpha\frac{\kappa_1^{\star 2}\rho_w+\kappa_*^{\star2}\rho_a+q-2\kappa_1^\star m+\Delta}{(1+V)^2}\\
\hat{m}=\frac{\kappa_1^\star\alpha}{1+V}
\end{cases}
&&\begin{cases}
V= \int \dd\mu(z)
\frac{z}{\lambda I_d+\hat{
V}z}
\\
q=\int \dd \mu(z)
\frac{\Delta_a \Delta_w \hat{m}^2z^3+\hat{q}\Sigma^2}{(\lambda I_d+\hat{
V}z)^2}
\\
m=\Delta_a \Delta_w \hat{m}\int \dd\mu(z)
\frac{z^2}{\lambda I_d+\hat{
V}z}
\end{cases}
\end{align}
The generalization error is 
\begin{equation}
\label{eq:App:ERM:eg_ridge_teacher2l}
    \epsilon_g=\kappa_1^{\star 2}\rho_w+\kappa_*^{\star 2}\rho_a+q-2\kappa_1^\star m
\end{equation}

\paragraph{Random Features}
Denoting $F\in \mathbb{R}^{k\times d}$ the RF matrix, which we will take $\mathcal{N}(0,\Delta_f)$, and denoting 
\begin{align}
    \kappa_1=\frac{1}{\Delta_f \frac{\Tr\Sigma}{d}}\mathbb{E}_z^{\mathcal{N}\left(0,\Delta_f \frac{\Tr\Sigma}{d}\right)}[z\sigma(z)]
    &&
    \kappa_*=\sqrt{
    \mathbb{E}_z^{\mathcal{N}\left(0,\Delta_f \frac{\Tr\Sigma}{d}\right)}[\sigma(z)^2]-\kappa_1^2 \Delta_f \frac{\Tr\Sigma}{d}
    }
\end{align}
the GET coefficients associated to the RF non linearity $\sigma(\cdot)$ \eqref{eq:kappa_multilayer}, the covariances read \eqref{eq:Omega_Psi_multilayer} (see Appendix \ref{App:GET})
\begin{align}
\begin{cases}
     &\rho=\kappa_1^{\star 2}\rho_w+\kappa_*^{\star2}\rho_a+\Delta,\\
    &\Omega=\kappa_1^2 \frac{F\Sigma F^\top}{d} +\kappa_*^2I_k\notag\\    &\Phi=\sqrt{\Delta_a\Delta_w}\kappa_1\kappa_1^\star \Sigma \frac{F^\top}{\sqrt{d}}. 
\end{cases}   
\end{align}
From which it follows, leveraging the asymptotic results of Theorem $1$ from  \cite{Loureiro2021CapturingTL},
\begin{align}
\label{eq:App:ERM:SP_RF_teacher2l}
\begin{cases}
\hat{V}=\frac{\frac{\alpha}{\gamma}}{1+V}\\
\hat{q}=\frac{\alpha}{\gamma}\frac{\kappa_1^{\star 2}\rho_w+\kappa_*^{\star2}\rho_a+q-2\kappa_1^\star m+\Delta}{(1+V)^2}\\
\hat{m}=\sqrt{\gamma}\frac{\kappa_1^\star\frac{\alpha}{\gamma}}{1+V}
\end{cases}
&&\begin{cases}
V=\frac{1}{\gamma d}\Tr[
\frac{\kappa_1^2 \frac{F\Sigma F^\top}{d} +\kappa_*^2I_k}{\lambda I_k+\hat{
V}(\kappa_1^2 \frac{F\Sigma F^\top}{d} +\kappa_*^2I_k)}
]\\
q=\frac{1}{\gamma d}\Tr[
\frac{\Delta_a \Delta_w \hat{m}^2\kappa_1^2 \frac{F\Sigma^2 F^\top}{d}(\kappa_1^2 \frac{F\Sigma F^\top}{d} +\kappa_*^2I_k)+\hat{q}(\kappa_1^2 F\Sigma F^\top +\kappa_*^2I_k)^2}{(\lambda I_k+\hat{
V}(\kappa_1^2 \frac{F\Sigma F^\top}{d} +\kappa_*^2I_k))^2}
]\\
m=\sqrt{\gamma}\frac{\Delta_a \Delta_w \hat{m}}{\gamma d}\Tr[
\frac{\kappa_1^2 \frac{F\Sigma^2 F^\top}{d} }{\lambda I_k+\hat{
V}(\kappa_1^2 \frac{F\Sigma F^\top}{d} +\kappa_*^2I_k)}
]
\end{cases}
\end{align}
The error is the same as for ridge regression \eqref{eq:App:ERM:eg_ridge_teacher2l}. In the case where $\Sigma=I_d$ (which we assume from now on), $F\Sigma F^\top$ and $F\Sigma^2 F^\top$ a jointly diagonalizable and introducing the spectral density $\rho(s)$ of $FF^\top/d$:
\begin{align}
\label{eq:App:ERM:SP_RF_teacher2l_spec}
\begin{cases}
\hat{V}=\frac{\frac{\alpha}{\gamma}}{1+V}\\
\hat{q}=\frac{\alpha}{\gamma}\frac{\kappa_1^{\star 2}\rho_w+\kappa_*^{\star2}\rho_a+q-2\kappa_1^\star m+\Delta}{(1+V)^2}\\
\hat{m}=\sqrt{\gamma}\frac{\kappa_1^\star\frac{\alpha}{\gamma}}{1+V}
\end{cases}
&&\begin{cases}
V=\int d\rho(s)
\frac{\kappa_1^2 s +\kappa_*^2}{\lambda +\hat{
V}(\kappa_1^2 s +\kappa_*^2)}\\
q=\int d\rho(s)
\frac{\Delta_a \Delta_w \hat{m}^2\kappa_1^2 s(\kappa_1^2 s +\kappa_*^2)+\hat{q}(\kappa_1^2 s +\kappa_*^2)^2}{(\lambda +\hat{
V}(\kappa_1^2 s +\kappa_*^2))^2}
\\
m=\sqrt{\gamma}\Delta_a \Delta_w \hat{m}\int d\rho(s)
\frac{\kappa_1^2 s }{\lambda +\hat{
V}(\kappa_1^2 s +\kappa_*^2)}
\end{cases}.
\end{align}
Introducing the Stieljes transform
\begin{equation}
    g(z)=\int d\rho(s)\frac{1}{s-z},
\end{equation}
Then the saddle point equations can be expressed as 
\begin{align}
\label{eq:App:ERM:SP_RF_teacher2l_Steiljes}
\begin{cases}
\hat{V}=\frac{\frac{\alpha}{\gamma}}{1+V}\\
\hat{q}=\frac{\alpha}{\gamma}\frac{\kappa_1^{\star 2}\rho_w+\kappa_*^{\star2}\rho_a+q-2\kappa_1^\star m\sqrt{\Delta_a \Delta_w}+\Delta}{(1+V)^2}\\
\hat{m}=\sqrt{\Delta_a \Delta_w}\sqrt{\gamma}\frac{\kappa_1^\star\frac{\alpha}{\gamma}}{1+V}
\end{cases}
&&\begin{cases}
&V=\frac{1}{\hat{V}}
-\frac{\lambda}{\hat{V}^2\kappa_1^2}g\left(
-\frac{\lambda+\hat{V}\kappa_*^2}{\hat{V}\kappa_1^2}
\right)
\\
&q=\frac{\hat{m}^2+\hat{q}}{\hat{V}^2}-\frac{1}{\kappa_1^2\hat{V}^2}\left(\frac{2\lambda(\hat{m}^2+\hat{q})}{\hat{V}}+\hat{m}^2\kappa_*^2
\right)g\left(
-\frac{\lambda+\hat{V}\kappa_*^2}{\hat{V}\kappa_1^2}
\right)\notag\\
&\qquad+\frac{\lambda}{\kappa_1^4\hat{V}^3}\left(\frac{\lambda(\hat{m}^2+\hat{q})}{\hat{V}}+\hat{m}^2\kappa_*^2
\right)g^\prime\left(
-\frac{\lambda+\hat{V}\kappa_*^2}{\hat{V}\kappa_1^2}
\right)
\\
&m=\sqrt{\gamma} \frac{\hat{m}}{\hat{V}}\left[
1-\frac{1}{\kappa_1^2}\left(
\frac{\lambda}{\hat{V}}+\kappa_*^2
\right)g\left(
-\frac{\lambda+\hat{V}\kappa_*^2}{\hat{V}\kappa_1^2}
\right)
\right].
\end{cases}
\end{align}

\paragraph{Kernels} 
We know consider the infinite width limit $\gamma\rightarrow\infty$ limit of \eqref{eq:App:ERM:SP_RF_teacher2l_spec}, corresponding to GP kernel regression. For Gaussian projection, $\rho(\cdot)$ is Marcenko-Pastur, and reads
\begin{equation}
    \rho(s)=\left(1-\frac{1}{\gamma}\right)\delta(s)+\nu(s)
\end{equation}
where
\begin{equation}
    \nu(s)=\frac{1}{\gamma 2\pi\Delta_f }\frac{\sqrt{(s-\Delta_f (\sqrt{\gamma}-1)^2)(\Delta_f (\sqrt{\gamma}+1)^2-s)}}{s}\mathbbm{1}_{\Delta_f (\sqrt{\gamma}-1)^2\le s\le \Delta_f (\sqrt{\gamma}+1)^2}.
\end{equation}
The measure $\nu(\cdot)$ simplifies in the $\gamma\rightarrow \infty$ limit. For any test function $h(\cdot)$:
\begin{align}
    \int d\nu(x)h(x)&=\int\limits_{\sigma^2(\gamma-2\sqrt{\gamma}+1)}^{\sigma^2(\gamma+2\sqrt{\gamma}+1)}\frac{dx}{2\pi\gamma\Delta_f }\frac{\sqrt{(x-\Delta_f (\sqrt{\gamma}-1)^2)(\Delta_f (\sqrt{\gamma}+1)^2-x)}}{x}h(x)\notag\\
    &=\int\limits_{-1}^1\frac{2dx}{\pi}\sqrt{(1-x)(1+x)}\frac{h(\gamma\Delta_f +2\sqrt{\gamma}\Delta_f x)}{\gamma+2\sqrt{\gamma}x}=\frac{h( \Delta_f \gamma)}{\gamma}+\mathcal{O}\left(\frac{1}{\sqrt{\gamma}}\right)
\end{align}
meaning 
\begin{equation}
    \rho(s)\approx \left(1-\frac{1}{\gamma}\right)\delta(s)+\frac{1}{\gamma}\delta(s-\gamma\Delta_f ).
\end{equation}
The saddle point equation become approximately
\begin{align}
\label{eq:App:ERM:SP_K_teacher2l_spec}
\begin{cases}
\hat{V}=\frac{\frac{\alpha}{\gamma}}{1+V}\\
\hat{q}=\frac{\alpha}{\gamma}\frac{\kappa_1^{\star 2}\rho_w+\kappa_*^{\star2}\rho_a+q-2\kappa_1^\star m+\Delta}{(1+V)^2}\\
\hat{m}=\sqrt{\gamma}\frac{\kappa_1^\star\frac{\alpha}{\gamma}}{1+V}
\end{cases}
&&\begin{cases}
V&=\left(1-\frac{1}{\gamma}\right)
\frac{ \kappa_*^2}{\lambda +\hat{
V}(\kappa_*^2)}+\frac{1}{\gamma}\frac{\kappa_1^2 \gamma\Delta_f  +\kappa_*^2}{\lambda +\hat{
V}(\kappa_1^2 \gamma\Delta_f  +\kappa_*^2)}\\
q&=\left(1-\frac{1}{\gamma}\right)\frac{\hat{q}\kappa_*^4}{(\lambda +\hat{
V}\kappa_*^2)^2}
+\frac{1}{\gamma}
\frac{\Delta_a \Delta_w \hat{m}^2\kappa_1^2 \gamma\Delta_f (\kappa_1^2 \gamma\Delta_f  +\kappa_*^2)+\hat{q}(\kappa_1^2 \gamma\Delta_f  +\kappa_*^2)^2}{(\lambda +\hat{
V}(\kappa_1^2 \gamma\Delta_f  +\kappa_*^2))^2}
\\
m&=\sqrt{\gamma}\Delta_a \Delta_w \hat{m}
\frac{1}{\gamma}
\frac{\kappa_1^2\gamma\Delta_f  }{\lambda +\hat{
V}(\kappa_1^2 \gamma\Delta_f  +\kappa_*^2)}.
\end{cases}
\end{align}
Rescaling 
\begin{align}
    \hat{V},\hat{q}\leftarrow\gamma \hat{V},\gamma \hat{q}
    &&
    \hat{m}\leftarrow\sqrt{\gamma}\hat{m}
\end{align}
The equations \eqref{eq:App:ERM:SP_K_teacher2l_spec} reduce to
\begin{align}
\label{eq:App:ERM:SP_K_teacher2l_spec_final}
\begin{cases}
\hat{V}=\frac{\alpha}{1+V}\\
\hat{q}=\alpha\frac{\kappa_1^{\star 2}\rho_w+\kappa_*^{\star2}\rho_a+q-2\kappa_1^\star m+\Delta}{(1+V)^2}\\
\hat{m}=\alpha\frac{\kappa_1^\star}{1+V}
\end{cases}
&&\begin{cases}
V&=\frac{\kappa_*^2}{\lambda}+\frac{\kappa_1^2\Delta_f }{\lambda+\hat{V}\kappa_1^2\Delta_f }
\\
q&=\frac{\Delta_a \Delta_w \hat{m}^2\kappa_1^4\Delta_f^2+\hat{q}\kappa_1^4\Delta_f^2}{(\lambda+\hat{V}\kappa_1^2\Delta_f )^2}
\\
m&=\Delta_a \Delta_w \hat{m}
\frac{\kappa_1^2\Delta_f }{\lambda+\hat{V}\kappa_1^2\Delta_f }
\end{cases}.
\end{align}

%% file: Appendix/ERM_regu.tex
In this appendix, we demonstrate that optimally regularized ridge regression and kernel regression achieve the Bayes optimal MSE, and provide a derivation of the explicit formulas for the optimal regularizations \eqref{eq:opt_reg_ridge} and \eqref{eq:Kernel_opt}. Note that these results recover those of \cite{SahraeeArdakan2022KernelMA}, which were derived for a GP target, corresponding to the infinite width limit of \eqref{eq:teacher}. In fact, \eqref{eq:opt_reg_ridge} and \eqref{eq:Kernel_opt} can be heuristically derived by evaluating the formulas of \cite{SahraeeArdakan2022KernelMA} on the equivalent linear teacher \eqref{eq:linear_model}.\\

\subsection{Ridge ERM achieves Bayes optimal error}
We first show that optimally regularized ridge regression achieves the Bayes optimal MSE \eqref{eq:Bayes_regression}. As a side result, the formula \eqref{eq:opt_reg_ridge} is obtained.

\paragraph{Optimal regularization for ridge}
Combining \eqref{eq:App:ERM:eg_ridge_teacher2l} and \eqref{eq:App:ERM:SP_ridge_teacher2l}, one can derive a self-consistent equation for $\epsilon_g$:
\begin{align}
\label{eq:App:ERM:self_cons_eg_ridge}
    \epsilon_g&=\kappa_*^2\rho_a+\int \dd\mu(z)
    \frac{\scriptstyle
    \kappa_1^2\Delta_w \Delta_a \hat{V}^2z^3+\hat{q}z^2-2\kappa_1^2\Delta_w \Delta_a \lambda \hat{V}z^2-2\kappa_1^2\Delta_w \Delta_a \hat{V}^2z^3+\kappa_1^2\Delta_w \Delta_a \lambda^2z+\kappa_1^2\Delta_w \Delta_a \hat{V}^2z^3+2\kappa_1^2\Delta_w \Delta_a \lambda\hat{V}z^2
    }{(\lambda I_d+\hat{V}z)^2}
    \notag\\
    &\overset{\Tilde{\lambda}\equiv \frac{\lambda}{\hat{V}}}{=}\kappa_*^2\rho_a+\int \dd \mu(z)
    \frac{\kappa_1^2\Delta_a \Delta_w \Tilde{\lambda}^2z+\frac{\epsilon_g}{\alpha}z^2}
    {(\Tilde{\lambda} I_d+z)^2}
\end{align}
Requiring $\epsilon_g$ to be minimal with respect to $\lambda$ (equivalently, $\Tilde{\lambda}$):
\begin{align}
    0&=\int \dd\mu(z)
\frac{2\kappa_1^2\Delta_w \Delta_a \Tilde{\lambda}z}{(\Tilde{\lambda} I_d+z)^2}-2\frac{2\kappa_1^2\Delta_w \Delta_a \Tilde{\lambda}^2z+\frac{\epsilon_g}{\alpha}z^2}{(\Tilde{\lambda} I_d+z)^3}
\\
&=\int \dd\mu(z) \frac{z^2}{(\Tilde{\lambda} I_d+z)^3}\times 2\kappa_1^2\Delta_w \Delta_a \left(\Tilde{\lambda}-\frac{ \epsilon_g}{\alpha\kappa_1^2\Delta_w \Delta_a }\right)
\end{align}
It follows the relation
\begin{equation}
\label{eq:App:ERM:opt_lambda}
    \lambda=\frac{\hat{V} \epsilon_g}{\alpha\kappa_1^2\Delta_w \Delta_a }=\frac{ \epsilon_g}{\kappa_1^2\Delta_w \Delta_a (1+V)}.
\end{equation}
\\

\paragraph{Explicit characterization of the optimal regularization}
We now proceed to show \eqref{eq:opt_reg_ridge}. Rewrite \eqref{eq:App:ERM:opt_lambda} as
\begin{equation}
    \epsilon_g=\alpha\kappa_1^2\Delta_w\Delta_a \tilde{\lambda},
\end{equation}
where we remind the auxilary variable $\tilde{\lambda}\equiv\lambda/\hat{V}$. Replacing $\epsilon_g$ in \eqref{eq:App:ERM:self_cons_eg_ridge} for $\tilde{\lambda}$, one reaches
\begin{equation}
\label{eq:App:ERM:ridge_tilde_lamb1}
    \tilde{\lambda}=\frac{1}{\alpha}\left(
    \frac{\kappa_*^2\Delta_a+\Delta}{\Delta_w\Delta_a\kappa_1^2}+\int \dd\mu(z) \frac{\tilde{\lambda}z}{\tilde{\lambda}I_d+z}
    \right)
\end{equation}
On the other hand, the saddle point equations for $\hat{V},V$ \eqref{eq:App:ERM:SP_ridge_teacher2l} can be rewritten as
\begin{align}
    \frac{1}{\hat{V}}=\frac{1}{\alpha}\left(1+\frac{1}{\hat{V}}\int \dd\mu(z) \frac{z}{\tilde{\lambda}I_d+z}\right)
\end{align}
Multiplying by $\lambda$ on both sides:
\begin{align}
\label{eq:App:ERM:ridge_tilde_lamb2}
    \tilde{\lambda}=\frac{1}{\alpha}\left(\lambda+\int \dd\mu(z) \frac{\tilde{\lambda}z}{\tilde{\lambda}I_d+z}\right)
\end{align}
Identifying \eqref{eq:App:ERM:ridge_tilde_lamb1} and \eqref{eq:App:ERM:ridge_tilde_lamb2} yield
\begin{equation}
    \lambda=\frac{\kappa_*^2\Delta_a+\Delta}{\Delta_w\Delta_a\kappa_1^2},
\end{equation}
which is \eqref{eq:opt_reg_ridge} for $L=1$ hidden layer.\\

\paragraph{Connection to \cite{SahraeeArdakan2022KernelMA}} In the two layer case, this result interestingly corresponds to the optimal regularization given by Theorem 4.1 in \cite{SahraeeArdakan2022KernelMA}, provided it is applied to the GP corresponding to the \textit{infinite-width} limit of the target. Following \cite{SahraeeArdakan2022KernelMA}, the optimal regularization for ridge regression over such a GP reads
$$
\lambda=\frac{c_0+\sigma^2}{c_2},
$$
which translating into our notations with 
\begin{align}
    c_2=\Delta_w\Delta_a \kappa_1^2 &&
    c_0=\Delta_a\kappa_*^2
    && \sigma^2=\Delta
\end{align}
leads back to 
\begin{equation}
\lambda=\frac{\kappa_*^2\Delta_a+\Delta}{\Delta_w\Delta_a\kappa_1^2}.
\end{equation}

\paragraph{Equivalence Ridge/Bayes} Finally, we show that ridge regression, regularized with \eqref{eq:opt_reg_ridge}, achieves the Bayes optimal MSE \eqref{eq:Bayes_regression}.
We will show to this end that $-\kappa_1^2q_w=q-2\kappa_1m$, taking inspiration from \eqref{eq:BO_SP} to define $\hat{q}_w=\alpha\kappa_1^2/\epsilon_g=(\hat{V}\Delta_w \Delta_a )/\lambda$. In the following, $\lambda$ is understood as the ridge optimal regularization \eqref{eq:opt_reg_ridge}.

\begin{align}
   -\frac{q-2\kappa_1m}{\kappa_1^2}&=\frac{1}{\kappa_1^2 }\int \dd\mu(z)
\frac{\Delta_w \Delta_a \kappa_1^2\hat{V}^2z^3+\hat{q}z^2-2\lambda\Delta_w \Delta_a \kappa_1^2\hat{V}z^2-2\Delta_w \Delta_a \kappa_1^2\hat{V}^2z^3}{(\lambda I_d+\hat{V}z)^2}
 \notag\\
&=\frac{1}{\kappa_1^2 }\int \dd\mu(z)
\frac{\frac{\hat{V}^2}{\alpha}\epsilon_gz^2+\Delta_w \Delta_a  \kappa_1^2\hat{V}^2z^3}{(\lambda I_d+\hat{V}z)^2}
\notag\\
&=\frac{1}{\kappa_1^2 }\int \dd\mu(z)
\frac{\hat{V}\lambda\Delta_w \Delta_a \kappa_1^2+\Delta_w \Delta_a  \kappa_1^2\hat{V}^2z^3}{(\lambda I_d+\hat{V}z)^2}
\notag\\
&=\frac{1}{\kappa_1^2 }\int \dd\mu(z)
\frac{\hat{V}\lambda\Delta_w \Delta_a \kappa_1^2z^2+\Delta_w \Delta_a  \kappa_1^2\hat{V}^2z^3}{(\lambda I_d+\hat{V}z)^2}
\notag\\
&=\int \dd\mu(z)\frac{\hat{q}_wz^2}{I_d+\Delta_w \Delta_a \hat{q}_wz}\notag\\
&=\int \dd\mu(z)\frac{\alpha \kappa_1^2 z^2}{\epsilon_g+\Delta_w \Delta_a \alpha \kappa_1^2 z}
\end{align}
which corresponds to the equation \eqref{eq:BO_SP}. Renaming $q_w:=-\sfrac{q-2\kappa_1m}{\kappa_1^2}$ allows to recover the two-layer Bayes MSE.

\subsection{Kernels achieve Bayes optimal error}
This subsection similarly addresses the case of kernels.\\
\paragraph{Optimal regularization for RF kernel ERM}
Combining \eqref{eq:App:ERM:eg_ridge_teacher2l} and \eqref{eq:App:ERM:SP_K_teacher2l_spec_final}, one can as in the ridge case derive a self-consistent equation for the generalization error:
\begin{align}
\label{eq:App:Reg_2l:Kernel_eg_opt}
    \epsilon_g-\Delta&=\kappa_*^{\star 2}\rho_a+\frac{\scriptstyle\Delta_w \Delta_a \kappa_1^{\star 2}\lambda^2+\Delta_w \Delta_a \kappa_1^{\star 2}\kappa_1^4\Delta_f^2\hat{V}+2\Delta_w \Delta_a \kappa_1^{\star 2}\lambda\hat{V}\kappa_1\Delta_f -\Delta_w \Delta_a \kappa_1^{\star 2}\kappa_1^4\Delta_f^2\hat{V}+\frac{\hat{V}^2}{\alpha}\kappa_1^4\Delta_f^2\epsilon_g-2\Delta_w \Delta_a \kappa_1^{\star 2}\lambda\hat{V}\kappa_1\Delta_f }{(\lambda+\hat{V}\kappa_1^2\Delta_f )^2}\notag\\
    &\overset{\Tilde{\lambda}\equiv \frac{\lambda}{\hat{V}}}{=}\kappa_*^{\star 2}\rho_a+\frac{\Delta_w \Delta_a \kappa_1^{\star 2}\Tilde{\lambda}^2+\frac{\kappa_1^4\Delta_f^2\epsilon_g}{\alpha}}{(\tilde{\lambda}+\kappa_1^2\Delta_f )^2}
\end{align}
Requiring $\frac{\partial \epsilon_g}{\partial \Tilde{\lambda}}=0$ leads to
\begin{align}
    0=\frac{2\Delta_w \Delta_a \kappa_1^{\star 2}\Tilde{\lambda}^2+2\Delta_w \Delta_a \kappa_1^{\star 2}\kappa_1^2\Delta_f \Tilde{\lambda}-2\Delta_w \Delta_a \kappa_1^{\star 2}\Tilde{\lambda}^2 -2\frac{\kappa_1^4\Delta_f^2\epsilon_g}{\alpha}}{(\Tilde{\lambda}+\kappa_1^2\Delta_f )^3}
    =2\kappa_1^2\Delta_f \frac{\Delta_w \Delta_a \kappa_1^{\star 2}\Tilde{\lambda}-\frac{\epsilon_g\kappa_1^2\Delta_f }{\alpha}}{(\Tilde{\lambda}+\kappa_1^2\Delta_f )^3}
\end{align}
The optimal regularization thus reads
\begin{equation}
\label{eq:App:Reg_2l:Kernel_opt_lambda_formula}
\lambda=\frac{\hat{V}\epsilon_g\kappa_1^2\Delta_f }{\Delta_w \Delta_a \alpha\kappa_1^{\star 2}}=\frac{\epsilon_g\kappa_1^2\Delta_f }{\Delta_w \Delta_a \kappa_1^{\star 2}(1+V)}.
\end{equation}

\paragraph{Explicit characterization for the optimal regularization}
One is now in position to derive \eqref{eq:Kernel_opt}.
 Combining \eqref{eq:App:Reg_2l:Kernel_eg_opt} and \eqref{eq:App:Reg_2l:Kernel_opt_lambda_formula}, the optimal $\lambda$ for kernel regression is the solution on the $\lambda$ variable of the system of equations
\begin{align}
    \begin{cases}
\lambda=\frac{\hat{V}\epsilon_g\kappa_1^2\Delta_f }{\Delta_w \Delta_a \alpha\kappa_1^{\star 2}}\\
   \epsilon_g=\Delta+\kappa_*^{\star 2}\rho_a+\frac{\Delta_w \Delta_a \kappa_1^{\star 2}\Tilde{\lambda}^2+\frac{\kappa_1^4\Delta_f^2\epsilon_g}{\alpha}}{(\tilde{\lambda}+\kappa_1^2\Delta_f )^2}\\
  \frac{1}{\hat{V}}=\frac{1}{\alpha}\left(
  1+\frac{\kappa_*^2}{\lambda}+\frac{\kappa_1^2\Delta_f}{\lambda+\hat{V}\kappa_1^2\Delta_f}
  \right)
    \end{cases}.
\end{align}
Introducing the variable $\tilde{\lambda}\equiv\lambda/\hat{V}$, and plugging the first line into the second, this simplifies to
\begin{align}
    \begin{cases}
        \epsilon_g=\alpha\frac{\Delta_w\Delta_a}{\Delta_f} \frac{\kappa_1^{\star2}}{\kappa_1^2}\tilde{\lambda}\\ \tilde{\lambda}=\frac{\Delta_f\kappa_1^2}{\Delta_w\kappa_1^{\star2}}\frac{1}{\alpha}\left(
        \kappa_*^{\star2}+\frac{\Delta}{\Delta_a}+\frac{\Delta\kappa_1^{\star2}\tilde{\lambda}}{\tilde{\lambda}+\kappa_1^2\Delta_f}
        \right)\\
    \tilde{\lambda}=\frac{1}{\alpha}\left(
    \lambda+\kappa_*^2+\frac{\kappa_1^2\Delta_f\tilde{\lambda}}{\tilde{\lambda}+\kappa_1^2\Delta_f}
    \right)
    \end{cases}.
\end{align}
Substracting the third line from the second yields an expression for $\lambda$:
\begin{equation}
    \frac{\Delta_f\kappa_1^2}{\Delta_w\kappa_1^{\star2}}\left(\kappa_*^{\star2}+\frac{\Delta}{\Delta_a}\right)=\lambda+\kappa_*^2.
\end{equation}
In other words, the optimal regularization is explicitly given by
\begin{equation}
    \lambda=\kappa_1^2\Delta_f\left(
    \frac{\Delta_a\kappa_*^{\star2}+\Delta}{\kappa_1^{\star 2}\Delta_w\Delta_a}-\frac{\kappa_*^2}{\kappa_1^2\Delta_f}
    \right)
\end{equation}

\paragraph{Alternative argument using \cite{SahraeeArdakan2022KernelMA}} Theorem 2.1 of \cite{SahraeeArdakan2022KernelMA} shows that regularizing a kernel method with $\lambda_{\mathrm{GP}}$ yields the same test error as ridge regression with regularization $\lambda_r$, provided these two regularization are related as
\begin{equation}
    \lambda_{\mathrm{GP}}=c_2\times \lambda_r-c_0,
\end{equation}
i.e. in our notations
\begin{equation}
\lambda_{\mathrm{GP}}=\Delta_f\kappa_1^2\times \lambda_r-\kappa_*^2.
\end{equation}
Taking $\lambda_r$ to be the optimal regularizer for ridge regression derived in the previous subsection, this leads to 
\begin{equation}
\lambda_{\mathrm{GP}}=\Delta_f\kappa_1^2\left(
\frac{(\kappa_*^\star)^2\Delta_a+\Delta}{\Delta_w\Delta_a(\kappa_1^\star)^2}
\right)-\kappa_*^2=\kappa_1^2\Delta_f\left(
    \frac{\Delta_a\kappa_*^{\star2}+\Delta}{\kappa_1^{\star 2}\Delta_w\Delta_a}-\frac{\kappa_*^2}{\kappa_1^2\Delta_f}
    \right)
\end{equation}

\paragraph{Equivalence kernel/BO}
We now show the optimally regularized error achieves the Bayes optimal MSE \eqref{eq:Bayes_regression}. More precisely, we show $q_w=-(q-2\kappa_1^\star m)/\kappa_1^{\star 2}$:
\begin{align}
    -\frac{q-2\kappa_1^\star m}{\kappa_1^{\star 2}}&=-\frac{1}{\kappa_1^{\star 2}}\frac{\Delta_w \Delta_a \kappa_1^{\star 2}\kappa_1^4\Delta_f^2\hat{V}^2+\kappa_1^3\Delta_f^2\frac{\hat{V}^2}{\alpha}\epsilon_g-2\Delta_w \Delta_a \kappa_1^{\star 2}\kappa_1^2\Delta_f \hat{V}\lambda-2\Delta_w \Delta_a \kappa_1^{\star 2}\kappa_1^4\Delta_f^2\hat{V}^2}{(\lambda+\hat{V}\kappa_1\Delta_f )^2}\notag\\
    &=\frac{\Delta_w \Delta_a (\Delta_f^2\kappa_1^4\hat{V}^2+2\kappa_1^2\Delta_f \lambda\hat{V})-\frac{\kappa_1^4\Delta_f^2}{\kappa_1^{\star 2}}\frac{\hat{V}}{\alpha}\epsilon_g}{(\lambda+\hat{V}\kappa_1\Delta_f )^2}\notag\\
    &=\frac{\Delta_w \Delta_a  \frac{\alpha\kappa_1^{\star 2}}{\epsilon_g}}{1+\Delta_w \Delta_a  \frac{\alpha\kappa_1^{\star 2}}{\epsilon_g}}\overset{\hat{q}_w=\alpha\kappa_1^{\star 2}/\epsilon_g}{\equiv} \frac{\Delta_w \Delta_a  \hat{q}_w}{1+\Delta_w \Delta_a  \hat{q}_w}
\end{align}
which is exactly the update for $q_w$ \eqref{eq:BO_SP}.

\subsection{A short RMT argument for finite width RF}
We provide in this subsection a short argument, at a physics level of rigor, that RF cannot achieve Bayes optimality away from the infinite width (kernel) limit, for simplicity in the isotropic $\Sigma=I_d$ case. Remember that, in the framework of \cite{Loureiro2021CapturingTL}, the covariance between the student features is  $\Omega=\kappa_1^2FF^\top/d+\kappa_*^2I_p$, while the average effective cross-covariance between teacher and student features can be taken as $\Phi=\kappa_1\kappa_1^\star F^\top /\sqrt{d}$. The labels are generated from the target features (last layer post-activation), and not from the student features, rendering the setting slightly intricate. However, under the assumption that the student features and teacher post-activations are jointly Gaussian, \cite{Clarte2022ASO} provide an equivalent model where the labels are generated directly from the student features. Introduce the effective teacher
\begin{equation}
    \check{\theta}\equiv\Omega^{-\frac{1}{2}}\Phi^\top a^\star
\end{equation}
and the effective noise variance
\begin{align}
    \check{\Delta}=\frac{1}{d}\Tr[\Psi-\Phi\Omega^{-1}\Phi^\top]
    =\left(\kappa_1^\star\right)^2\left[
    1-\gamma\int d\rho(s)\frac{\kappa_*^2s}{\kappa_1^2s+\kappa_*^2}
    \right]=\left(\kappa_1^\star\right)^2\left[
    1-\gamma+\gamma\frac{\kappa_*^2}{\kappa_1^2}g\left(-\frac{\kappa_*^2}{\kappa_1^2}\right)
    \right],
\end{align}
where $g(\cdot)$ is the Stieljes transform of the Marcenko Pastur law. Note that $\check{\Delta}=\mathcal{O}(1)$ even in the $\gamma\rightarrow \infty$ limit. Then random feature regression on a two-layer target is equivalent to the ERM problem
\begin{align}
    &y\sim\frac{1}{\sqrt{p}}\check{\theta}^\top u+\sqrt{\check{\Delta}}\mathcal{N}(0,1),\notag\\
    & \check{\mathcal{R}}(w)=\sum\limits_{\mu=1}^n \frac{1}{2}\left(
    y^\mu-\frac{1}{\sqrt{p}}w^\top u^\mu
    \right)^2+\frac{\lambda}{2}w^\top w, \qquad u\sim \mathcal{N}(0,\Omega).
\end{align}
Absorbing the $\Omega$ dependence into the weights, this problem is further equivalent to 
\begin{align}
\label{eq:equivalent_risk_RF}
&y\sim\frac{1}{\sqrt{d}}a_\star^\top \Phi u+\sqrt{\check{\Delta}}\mathcal{N}(0,1),\notag\\
    &\check{\mathcal{R}}(w)=\frac{1}{2\check{\Delta}}\sum\limits_{\mu=1}^n \left(
    y^\mu-\frac{1}{\sqrt{d}}w^\top u^\mu
    \right)^2+\frac{\check{\Delta}\lambda\gamma}{2}w^\top \Omega^{-1}w, \qquad u\sim \mathcal{N}(0,I_p)
\end{align}
 Note that when $a_\star\sim\mathcal{N}(0,I_d)$, then $\Phi^\top a_\star\sim \mathcal{N}(0,\Phi^\top\Phi/\gamma)$. As discussed in the main text, ERM using a linear model and a square loss, as implied by \eqref{eq:equivalent_risk_RF}, is equivalent to conducting Bayesian inference, with prior $\check{\Delta}$ over the additive noise and prior $\sfrac{1}{\lambda \gamma \check{\Delta}}\Omega$ over the weights. Note that this prior matches the ground truth covariance if and only if $\exists \lambda, ~\sfrac{\Phi^\top\Phi}{\gamma}=\sfrac{1}{\lambda \gamma \check{\Delta}}\Omega$. This requires at least the spectra to match, i.e
\begin{equation}
    \frac{\left(\kappa_1^\star\right)^2}{\gamma}\mathrm{spec}\left(\kappa_1^2FF^\top/d\right)\propto\frac{1}{\gamma\lambda}\mathrm{spec}\left(\kappa_1^2FF^\top/d
    \right)+\frac{\kappa_*^2}{\gamma\lambda},
\end{equation}
which can only be verified asymptotically as $\gamma\rightarrow \infty$, when the spectrum has a finite fraction $1/\gamma$ of $\mathcal{O}(\gamma)$ eigenvalues and $(\gamma-1)/\gamma$ of zero eigenvalues, i.e. in the infinite width (kernel) limit. This implies that doing \textit{finite width} RF regression is equivalent to doing Bayes inference with \textit{mismatched} prior over the effective weights, and that therefore the test error achieved by RF regression cannot be Bayes optimal. This is numerically obersved in Fig.\,\ref{fig:regression}, where the test error of RF is bounded away from the Bayes MSE baseline \eqref{eq:Bayes_regression}.

%% file: Appendix/Multi_ERM.tex
In this section, we extend the discussion of Appendix \ref{App:ERM} and \ref{App:regu_ERM} to the general multi-layer case. {\color{black} We show here how the corresponding formulas are recovered from Theorem $1$ of \cite{Loureiro2021CapturingTL}, applied to the shallow target \eqref{eq:linear_model}, as discussed in section \ref{sec:ERM_regression} in the main text. Another derivation of this sharp asymptotic characterization of ERM methods on the deep target \eqref{eq:teacher} appeared concomitantly in \cite{DRF2023}.}

\subsection{Ridge regression}

Similarly to the two-layer case, these results leverage the asymptotics of \cite{Loureiro2021CapturingTL} on the equivalent shallow network \eqref{eq:linear_model}, using, in their notations,
\begin{align}
\begin{cases}
     &\rho=\scriptstyle \prod\limits_{\ell=1}^{L^\star}\left(\kappa_1^{\star(\ell)}\right)^2\rho^{L^\star}_1+\sum\limits_{\ell_0=1}^{L-1}\kappa_*^{\star(\ell_0)}\prod\limits_{\ell=\ell_0+1}^{L^\star}\left(\kappa_1^{(\ell)}\right)^2\rho^{L^\star}_{\ell_0+1}+\left(\kappa_*^{L^\star}\right)^2\rho_a+\Delta,\\
    &\Omega=\Sigma \notag\\    &\Phi=\Delta_a^{\frac{1}{2}}\prod\limits_{\ell=1}^{L^\star}\Delta_\ell^{\frac{1}{2}}\kappa_1^{\star(\ell)} \Sigma
\end{cases}   
\end{align}

The replica saddle-point equations for a ridge student then read:
\begin{align}
\label{eq:App:Multi_ERMSP_ridge_teacherML}
\begin{cases}
\hat{V}=\frac{\alpha}{1+V}\\
\hat{q}=\alpha\frac{\prod\limits_{\ell=1}^{L^\star}\left(\kappa_1^{\star(\ell)}\right)^2\rho^{L^\star}_1+\sum\limits_{\ell_0=1}^{L-1}\kappa_*^{\star(\ell_0)}\prod\limits_{\ell=\ell_0+1}^{L^\star}\left(\kappa_1^{(\ell)}\right)^2\rho^{L^\star}_{\ell_0+1}+\left(\kappa_*^{L^\star}\right)^2\rho_a+q-2\prod\limits_{\ell=1}^{L^\star}\kappa_1^{\star(\ell)} m+\Delta}{(1+V)^2}\\
\hat{m}=\frac{\prod\limits_{\ell=1}^{L^\star}\kappa_1^{\star(\ell)}\alpha}{1+V}
\end{cases}
&&\begin{cases}
V=\int \dd\mu(z)
\frac{z}{\lambda I_d+\hat{
V}z}
\\
q=\int \dd\mu(z)
\frac{\Delta_a \prod\limits_{\ell=1}^{L^\star}\Delta_\ell \hat{m}^2z^3+\hat{q}z^2}{(\lambda I_d+\hat{
V}z)^2}
\\
m=\Delta_a \prod\limits_{\ell=1}^{L^\star}\Delta_\ell \hat{m}\int \dd\mu(z)
\frac{z^2}{\lambda I_d+\hat{
V}z}
.
\end{cases}
\end{align}
The generalization error is in the multilayer case 
\begin{equation}
\label{eq:App:Multi_ERMeg_ridge_teacherML}
    \epsilon_g=\prod\limits_{\ell=1}^{L^\star}\left(\kappa_1^{\star(\ell)}\right)^2\rho^{L^\star}_1+\sum\limits_{\ell_0=1}^{L^\star-1}\kappa_*^{\star(\ell_0)}\prod\limits_{\ell=\ell_0+1}^{L^\star}\left(\kappa_1^{\star(\ell)}\right)^2\rho^{L^\star}_{\ell_0+1}+\left(\kappa_*^{L^\star}\right)^2\rho_a+q-2\prod\limits_{\ell=1}^{L^\star}\kappa_1^{\star(\ell)} m+\Delta.
\end{equation}

\paragraph{Optimal regularization for ridge}
Combining \eqref{eq:SP_ridge_teacherML} and \eqref{eq:eg_ridge_teacherML}, a self-consistent equation for $\epsilon_g$ can be reached:
\begin{align}
\label{App:ERM_multi:eq:self_cons_eg_ridge}
    \epsilon_g
    &\overset{\Tilde{\lambda}\equiv \frac{\lambda}{\hat{V}}}{=}\epsilon_r+\int \dd\mu(z)
    \frac{\prod\limits_{\ell=1}^{L}\left(\kappa_1^{(\ell)}\right)^2\Delta_a \prod\limits_{\ell=1}^L \Delta_\ell \Tilde{\lambda}^2z+\frac{\epsilon_g}{\alpha}z^2}
    {(\Tilde{\lambda} I_d+z)^2}
\end{align}
Requiring $\epsilon_g$ to be minimal with respect to $\lambda$ (equivalently, $\Tilde{\lambda}$):
\begin{align}
    0&=\int \dd\mu(z)
\frac{2\prod\limits_{\ell=1}^{L}\left(\kappa_1^{(\ell)}\right)^2\Delta_a \prod\limits_{\ell=1}^L \Delta_\ell \Tilde{\lambda}z}{(\Tilde{\lambda} I_d+z)^2}-2\frac{2\prod\limits_{\ell=1}^{L}\left(\kappa_1^{(\ell)}\right)^2\Delta_a \prod\limits_{\ell=1}^L \Delta_\ell \Tilde{\lambda}^2z+\frac{\epsilon_g}{\alpha}z^2}{(\Tilde{\lambda} I_d+z)^3}
\\
&=\int \dd\mu(z) \frac{z^2}{(\Tilde{\lambda} I_d+z)^3}\times 2\prod\limits_{\ell=1}^{L}\left(\kappa_1^{(\ell)}\right)^2\Delta_a \prod\limits_{\ell=1}^L \Delta_\ell \left(\Tilde{\lambda}-\frac{ \epsilon_g}{\alpha\Delta_a \prod\limits_{\ell=1}^{L}\left(\kappa_1^{(\ell)}\right)^2 \Delta_\ell}\right)
\end{align}
It follows the relation
\begin{equation}
\label{App:ERM_multi:eq:opt_lambda}
    \lambda=\frac{\hat{V} \epsilon_g}{\alpha\prod\limits_{\ell=1}^{L}\left(\kappa_1^{(\ell)}\right)^2\Delta_a \prod\limits_{\ell=1}^L \Delta_\ell }=\frac{ \epsilon_g}{(1+V)\Delta_a \prod\limits_{\ell=1}^{L}\left(\kappa_1^{(\ell)}\right)^2 \Delta_\ell }.
\end{equation}

One is now in position to derive the optimal regularization \eqref{eq:opt_reg_ridge}.

Rewrite \eqref{App:ERM_multi:eq:opt_lambda} as
$$
    \epsilon_g=\alpha \tilde{\lambda} \Delta_a \prod\limits_{\ell=1}^{L}\left(\kappa_1^{(\ell)}\right)^2 \Delta_\ell,
$$
where we remind the auxilary variable $\tilde{\lambda}\equiv\lambda/\hat{V}$. Replacing $\epsilon_g$ in \eqref{App:ERM_multi:eq:self_cons_eg_ridge} for $\tilde{\lambda}$, one reaches
\begin{equation}
\label{App:ERM:eq:ridge_tilde_lamb1}
    \tilde{\lambda}=\frac{1}{\alpha}\left(
    \frac{\epsilon_r}{\Delta_a \prod\limits_{\ell=1}^{L}\left(\kappa_1^{(\ell)}\right)^2 \Delta_\ell}+\int \dd\mu(z)\frac{\tilde{\lambda}z}{\tilde{\lambda}I_d+z}
    \right).
\end{equation}
On the other hand, the saddle point equations for $\hat{V},V$ \eqref{eq:App:ERM:SP_ridge_teacher2l} can be rewritten as
\begin{align}
    \frac{1}{\hat{V}}=\frac{1}{\alpha}\left(1+\frac{1}{\hat{V}}\int \dd\mu(z)\frac{z}{\tilde{\lambda}I_d+z}\right)
\end{align}
Multiplying by $\lambda$ on both sides:
\begin{align}
\label{App:ERM:eq:ridge_tilde_lamb2}
    \tilde{\lambda}=\frac{1}{\alpha}\left(\lambda+\int \dd\mu(z)\frac{\tilde{\lambda}z}{\tilde{\lambda}I_d+z}\right).
\end{align}
Identifying \eqref{App:ERM:eq:ridge_tilde_lamb1} and \eqref{App:ERM:eq:ridge_tilde_lamb2} yield
\begin{equation}
    \lambda=\frac{\epsilon_r}{\Delta_a \prod\limits_{\ell=1}^{L}\left(\kappa_1^{(\ell)}\right)^2 \Delta_\ell}
\end{equation}
which is \eqref{eq:opt_reg_ridge}.\\

\paragraph{Equivalence Ridge/BO}
We now show that under the optimal regularization \eqref{eq:opt_reg_ridge}, the saddle point equations for ridge regression \eqref{eq:SP_ridge_teacherML} reduce to \eqref{eq:SP_Bayes}. It then straightforwardly follow that ridge regression achieves the Bayes optimal MSE. In the following, $\lambda$ is understood as the ridge optimal regularization \eqref{eq:opt_reg_ridge}. Define

$$
\hat{q}=\frac{\alpha\prod\limits_{\ell=1}^{L}\left(\kappa_1^{(\ell)}\right)^2}{\epsilon_g}=\frac{\hat{V}\Delta_a \prod\limits_{\ell=1}^{L} \Delta_\ell }{\lambda}
$$
to connect with \eqref{eq:SP_Bayes}. Then

\begin{align}
   -\frac{q-2\prod\limits_{\ell=1}^{L}\kappa_1^{(\ell)}m}{\prod\limits_{\ell=1}^{L}\left(\kappa_1^{(\ell)}\right)^2}&=\frac{1}{\kappa_1^2 }\int \dd\mu(z) 
\frac{\hat{q}z^2+\Delta_a \prod\limits_{\ell=1}^{L}\left(\kappa_1^{(\ell)}\right)^2 \Delta_\ell\left(\hat{V}^2z^3+-2\lambda\hat{V}z^2-2\hat{V}^2z^3\right)}{(\lambda I_d+\hat{V}z)^2}
 \notag\\
&=\frac{1}{\kappa_1^2 }\int \dd\mu(z) 
\frac{\frac{\hat{V}^2}{\alpha}\epsilon_gz^2+\Delta_a \prod\limits_{\ell=1}^{L}\left(\kappa_1^{(\ell)}\right)^2 \Delta_\ell\hat{V}^2z^3}{(\lambda I_d+\hat{V}z)^2}
\notag\\
&=\Delta_a \prod\limits_{\ell=1}^{L} \Delta_\ell\int \dd\mu(z) 
\frac{\hat{V}\lambda+\hat{V}^2z^3}{(\lambda I_d+\hat{V}z)^2}
\notag\\
&=\Delta_a \prod\limits_{\ell=1}^{L} \Delta_\ell\int \dd\mu(z) 
\frac{\hat{V}\lambda z^2+\hat{V}^2z^3}{(\lambda I_d+\hat{V}z)^2}
\notag\\
&=\int \dd\mu(z)\frac{\hat{q}z^2}{I_d+\Delta_a \prod\limits_{\ell=1}^{L} \Delta_\ell \hat{q}z}\notag\\
&=\int \dd\mu(z)\frac{\alpha\prod\limits_{\ell=1}^{L}\left(\kappa_1^{(\ell)}\right)^2z^2}{\epsilon_g I_d+\Delta_a \prod\limits_{\ell=1}^{L} \Delta_\ell \alpha\prod\limits_{\ell=1}^{L}\left(\kappa_1^{(\ell)}\right)^2z}\notag\\
\end{align}
which is exactly the Bayes Optimal equation \eqref{eq:SP_Bayes}.

\subsection{Random nets / features}
The linearization \eqref{eq:Omega_Psi_multilayer}, further detailed in Appendix \ref{App:GET}, translate in the framework of \cite{Loureiro2021CapturingTL} into
\begin{align}
\begin{cases}
     &\rho=\scriptstyle \prod\limits_{\ell=1}^{L^\star}\left(\kappa_1^{\star(\ell)}\right)^2\rho^{L^\star}_1+\sum\limits_{\ell_0=1}^{L-1}\kappa_*^{\star(\ell_0)}\prod\limits_{\ell=\ell_0+1}^{L^\star}\left(\kappa_1^{(\ell)}\right)^2\rho^{L^\star}_{\ell_0+1}+\left(\kappa_*^{L^\star}\right)^2\rho_a+\Delta,\\
    &\Omega=\kappa_1^2 \frac{F\Sigma F^\top}{d} +\kappa_*^2I_k\notag\\    &\Phi=\sqrt{\Delta_a\prod\limits_{\ell=1}^L\Delta_\ell}\kappa_1\prod\limits_{\ell=1}^L\kappa_1^{\star (\ell)} \Sigma \frac{F^\top}{\sqrt{d}} 
\end{cases}   
\end{align}
yielding the saddle point equations
\begin{align}
\label{eq:App:Multi_ERMSP_RF_teacherML_spec}
&\begin{cases}
\hat{V}=\frac{\frac{\alpha}{\gamma}}{1+V}\\
\hat{q}=\frac{\alpha}{\gamma}\frac{\prod\limits_{\ell=1}^{L^\star}\left(\kappa_1^{\star(\ell)}\right)^2\rho^{L^\star}_1+\sum\limits_{\ell_0=1}^{L-1}\kappa_*^{\star(\ell_0)}\prod\limits_{\ell=\ell_0+1}^{L^\star}\left(\kappa_1^{(\ell)}\right)^2\rho^{L^\star}_{\ell_0+1}+\left(\kappa_*^{L^\star}\right)^2\rho_a+q-2\prod\limits_{\ell=1}^{L^\star}\kappa_1^{\star(\ell)} m+\Delta}{(1+V)^2}\\
\hat{m}=\sqrt{\gamma}\frac{\prod\limits_{\ell=1}^{L^\star}\kappa_1^{\star(\ell)}\frac{\alpha}{\gamma}}{1+V}
\end{cases}
\\
&\begin{cases}
V=\int d\rho(s)
\frac{\kappa_1^2 s +\kappa_*^2}{\lambda +\hat{
V}(\kappa_1^2 s +\kappa_*^2)}\\
q=\int d\rho(s)
\frac{\Delta_a \prod\limits_{\ell=1}^{L^\star}\Delta_\ell \hat{m}^2\kappa_1^2 s(\kappa_1^2 s +\kappa_*^2)+\hat{q}(\kappa_1^2 s +\kappa_*^2)^2}{(\lambda +\hat{
V}(\kappa_1^2 s +\kappa_*^2))^2}
\\
m=\sqrt{\gamma}\Delta_a \prod\limits_{\ell=1}^{L^\star}\Delta_\ell \hat{m}\int d\rho(s)
\frac{\kappa_1^2 s }{\lambda +\hat{
V}(\kappa_1^2 s +\kappa_*^2)}
\end{cases}
\end{align}

\subsection{GP kernels}
By generalizing the 2-layer equations to the multi-layer case,
\begin{align}
\label{eq:App:Multi_ERMSP_K_teacherML_spec_final}
\begin{cases}
\hat{V}=\frac{\alpha}{1+V}\\
\hat{q}=\alpha\frac{\prod\limits_{\ell=1}^{L^\star}\left(\kappa_1^{\star(\ell)}\right)^2\rho^{L^\star}_1+\sum\limits_{\ell_0=1}^{L-1}\kappa_*^{\star(\ell_0)}\prod\limits_{\ell=\ell_0+1}^{L^\star}\left(\kappa_1^{(\ell)}\right)^2\rho^{L^\star}_{\ell_0+1}+\left(\kappa_*^{L^\star}\right)^2\rho_a+q-2\prod\limits_{\ell=1}^{L^\star}\kappa_1^{\star(\ell)} m+\Delta}{(1+V)^2}\\
\hat{m}=\alpha\frac{\prod\limits_{\ell=1}^{L^\star}\kappa_1^{\star(\ell)}}{1+V}
\end{cases}
&&\begin{cases}
V&=\frac{\kappa_*^2}{\lambda}+\frac{\kappa_1^2\Delta_f }{\lambda+\hat{V}\kappa_1^2\Delta_f }
\\
q&=\frac{\Delta_a \prod\limits_{\ell=1}^{L^\star}\Delta_\ell \hat{m}^2\kappa_1^4\Delta_f^2+\hat{q}\kappa_1^4\Delta_f^2}{(\lambda+\hat{V}\kappa_1^2\Delta_f )^2}
\\
m&=\Delta_a \prod\limits_{\ell=1}^{L^\star}\Delta_\ell \hat{m}
\frac{\kappa_1^2\Delta_f }{\lambda+\hat{V}\kappa_1^2\Delta_f }
\end{cases}
\end{align}

\paragraph{Optimal regularization for RF kernel ERM}
Combining \eqref{eq:App:ERM:eg_ridge_teacher2l} and \eqref{eq:App:ERM:SP_K_teacher2l_spec_final}, one can as in the ridge case derive a self-consistent equation for the generalization error:
\begin{align}
    \epsilon_g-\Delta&=\epsilon_r+\frac{\scriptscriptstyle\Delta_a \prod\limits_{\ell=1}^{L}\left(\kappa_1^{(\ell)}\right)^2 \Delta_\ell\lambda^2+\Delta_a \prod\limits_{\ell=1}^{L}\left(\kappa_1^{(\ell)}\right)^2 \Delta_\ell\kappa_1^4\Delta_f^2\hat{V}+2\Delta_a \prod\limits_{\ell=1}^{L}\left(\kappa_1^{(\ell)}\right)^2 \Delta_\ell\lambda\hat{V}\kappa_1\Delta_f -\Delta_a \prod\limits_{\ell=1}^{L}\left(\kappa_1^{(\ell)}\right)^2 \Delta_\ell\kappa_1^4\Delta_f^2\hat{V}+\frac{\hat{V}^2}{\alpha}\kappa_1^4\Delta_f^2\epsilon_g-2\Delta_a \prod\limits_{\ell=1}^{L}\left(\kappa_1^{(\ell)}\right)^2 \Delta_\ell\lambda\hat{V}\kappa_1\Delta_f }{(\lambda+\hat{V}\kappa_1^2\Delta_f )^2}\notag\\
    &\overset{\Tilde{\lambda}\equiv \frac{\lambda}{\hat{V}}}{=}\epsilon_r+\frac{\Delta_a \prod\limits_{\ell=1}^{L}\left(\kappa_1^{(\ell)}\right)^2 \Delta_\ell\Tilde{\lambda}^2+\frac{\kappa_1^4\Delta_f^2\epsilon_g}{\alpha}}{(\tilde{\lambda}+\kappa_1^2\Delta_f )^2}
\end{align}
Requiring $\frac{\partial \epsilon_g}{\partial \Tilde{\lambda}}=0$ leads to
\begin{align}
    0=\frac{2\Delta_a \prod\limits_{\ell=1}^{L}\left(\kappa_1^{(\ell)}\right)^2 \Delta_\ell\left(\Tilde{\lambda}^2+2\kappa_1^2\Delta_f \Tilde{\lambda}-2\Tilde{\lambda}^2 \right)-2\frac{\kappa_1^4\Delta_f^2\epsilon_g}{\alpha}}{(\Tilde{\lambda}+\kappa_1^2\Delta_f )^3}
    =2\kappa_1^2\Delta_f \frac{\Delta_a \prod\limits_{\ell=1}^{L}\left(\kappa_1^{(\ell)}\right)^2 \Delta_\ell\Tilde{\lambda}-\frac{\epsilon_g\kappa_1^2\Delta_f }{\alpha}}{(\Tilde{\lambda}+\kappa_1^2\Delta_f )^3}.
\end{align}
The optimal regularization thus reads
\begin{equation}
    \lambda=\frac{\hat{V}\epsilon_g\kappa_1^2\Delta_f }{ \alpha\Delta_a \prod\limits_{\ell=1}^{L}\left(\kappa_1^{(\ell)}\right)^2 \Delta_\ell}=\frac{\epsilon_g\kappa_1^2\Delta_f }{\Delta_a \prod\limits_{\ell=1}^{L}\left(\kappa_1^{(\ell)}\right)^2 \Delta_\ell(1+V)}.
\end{equation}

\noindent 
The optimal $\lambda$ is therefore the solution on the $\lambda$ variable of the system of equations
\begin{align}
    \begin{cases}
     \lambda=\frac{\hat{V}\epsilon_g\kappa_1^2\Delta_f }{ \alpha\Delta_a \prod\limits_{\ell=1}^{L}\left(\kappa_1^{(\ell)}\right)^2 \Delta_\ell}\\
   \epsilon_g=\epsilon_r+\frac{\Delta_a \prod\limits_{\ell=1}^{L}\left(\kappa_1^{(\ell)}\right)^2 \Delta_\ell\Tilde{\lambda}^2+\frac{\kappa_1^4\Delta_f^2\epsilon_g}{\alpha}}{(\tilde{\lambda}+\kappa_1^2\Delta_f )^2}\\
  \frac{1}{\hat{V}}=\frac{1}{\alpha}\left(
  1+\frac{\kappa_*^2}{\lambda}+\frac{\kappa_1^2\Delta_f}{\lambda+\hat{V}\kappa_1^2\Delta_f}
  \right)
    \end{cases}.
\end{align}
Introducing the variable $\tilde{\lambda}\equiv\lambda/\hat{V}$, and plugging the first line into the second, this simplifies to
\begin{align}
    \begin{cases}
        \epsilon_g=\alpha\frac{\Delta_a \prod\limits_{\ell=1}^{L} \Delta_\ell}{\Delta_f} \frac{\prod\limits_{\ell=1}^{L}\left(\kappa_1^{(\ell)}\right)^2}{\kappa_1^2}\tilde{\lambda}\\ \tilde{\lambda}=\frac{\Delta_f\kappa_1^2}{\Delta_a \prod\limits_{\ell=1}^{L}\left(\kappa_1^{(\ell)}\right)^2 \Delta_\ell}\frac{1}{\alpha}\left(\epsilon_r+\frac{\Delta_a \prod\limits_{\ell=1}^{L}\left(\kappa_1^{(\ell)}\right)^2 \Delta_\ell\tilde{\lambda}}{\tilde{\lambda}+\kappa_1^2\Delta_f}
        \right)\\
    \tilde{\lambda}=\frac{1}{\alpha}\left(
    \lambda+\kappa_*^2+\frac{\kappa_1^2\Delta_f\tilde{\lambda}}{\tilde{\lambda}+\kappa_1^2\Delta_f}
    \right)
    \end{cases}.
\end{align}
Substracting the third line from the second yields an expression for $\lambda$:
\begin{equation}
    \frac{\Delta_f\kappa_1^2}{\Delta_a \prod\limits_{\ell=1}^{L}\left(\kappa_1^{(\ell)}\right)^2 \Delta_\ell}\epsilon_r=\lambda+\kappa_*^2.
\end{equation}
In other words the optimal regularization is explicitly given by
\begin{equation}
    \lambda=\kappa_1^2\Delta_f\left(
    \frac{\epsilon_r}{\Delta_a \prod\limits_{\ell=1}^{L}\left(\kappa_1^{(\ell)}\right)^2 \Delta_\ell}-\frac{\kappa_*^2}{\kappa_1^2\Delta_f}
    \right)
\end{equation}
corresponding to \eqref{eq:Kernel_opt}.\\

\paragraph{Equivalence GP kernel/BO}
Under optimal regularization \eqref{eq:Kernel_opt} the saddle-point equations for the Bayes-optimal MSE can be recovered:
\begin{align}
    -\frac{q-2\prod\limits_{\ell=1}^{L}\kappa_1^{(\ell)} m}{\prod\limits_{\ell=1}^{L}\left(\kappa_1^{(\ell)}\right)^2}&=-\frac{1}{\prod\limits_{\ell=1}^{L}\left(\kappa_1^{(\ell)}\right)^2}\frac{\scriptstyle \Delta_a \prod\limits_{\ell=1}^{L}\left(\kappa_1^{(\ell)}\right)^2 \Delta_\ell\kappa_1^4\Delta_f^2\hat{V}^2+\kappa_1^3\Delta_f^2\frac{\hat{V}^2}{\alpha}\epsilon_g-2\Delta_a \prod\limits_{\ell=1}^{L}\left(\kappa_1^{(\ell)}\right)^2 \Delta_\ell\kappa_1^2\Delta_f \hat{V}\lambda-2\Delta_a \prod\limits_{\ell=1}^{L}\left(\kappa_1^{(\ell)}\right)^2 \Delta_\ell\kappa_1^4\Delta_f^2\hat{V}^2}{(\lambda+\hat{V}\kappa_1\Delta_f )^2}\notag\\
    &=\frac{\Delta_a \prod\limits_{\ell=1}^{L}\Delta_\ell (\Delta_f^2\kappa_1^4\hat{V}^2+2\kappa_1^2\Delta_f \lambda\hat{V})-\frac{\kappa_1^4\Delta_f^2}{\prod\limits_{\ell=1}^{L}\left(\kappa_1^{(\ell)}\right)^2}\frac{\hat{V}}{\alpha}\epsilon_g}{(\lambda+\hat{V}\kappa_1\Delta_f )^2}\notag\\
    &=\frac{\Delta_a \prod\limits_{\ell=1}^{L}\Delta_\ell  \frac{\alpha\prod\limits_{\ell=1}^{L}\left(\kappa_1^{(\ell)}\right)^2}{\epsilon_g}}{1+\Delta_a \prod\limits_{\ell=1}^{L}\Delta_\ell \frac{\alpha\prod\limits_{\ell=1}^{L}\left(\kappa_1^{(\ell)}\right)^2}{\epsilon_g}}=\frac{\Delta_a \prod\limits_{\ell=1}^{L}\Delta_\ell  \alpha\prod\limits_{\ell=1}^{L}\left(\kappa_1^{(\ell)}\right)^2}{\epsilon_g+\Delta_a \prod\limits_{\ell=1}^{L}\Delta_\ell \alpha\prod\limits_{\ell=1}^{L}\left(\kappa_1^{(\ell)}\right)^2}
\end{align}
which recovers \eqref{eq:BO_SP}.

%% file: Appendix/Classification.tex
\subsection{Bayes-optimal error for classification}
In this section, we treat the case of classification, corresponding to a sign readout $f_\star(x)=\mathrm{sign}(x)$ in \eqref{eq:teacher}. Since the finite temperature free energy of Appendix \ref{App:multilayer} is derived under no assumption over the output channel, it is amenable to being readily specialized to the present setting. The output partition function reads
\begin{align}
    \mathcal{Z}(y,\omega,V)=\frac{1}{2}\left[
    1-\mathrm{erf}\left(\frac{\omega}{\sqrt{2(V+\Delta)}}\right)
    \right].
\end{align}
The saddle point equations can be massaged into the compact form
\begin{align}
\label{eq:classification_BO_SP}
    \begin{cases}
    q=\int\frac{\hat{q}\Delta_a^2\prod\limits_{\ell=1}^L \Delta_\ell^2z^2}{\hat{q}z\Delta_a\prod\limits_{\ell=1}^L \Delta_\ell+1}\dd\mu(z)\\
    \hat{q}=\frac{2\alpha\prod\limits_{\ell=1}^L\left(\kappa_1^{(\ell)}\right)^2}{\Delta_a\int 
 z\dd \mu(z)\prod\limits_{\ell=1}^L \left(\kappa_1^{(\ell)}\right)^2\Delta_\ell+\epsilon_r-\prod\limits_{\ell=1}^L\left(\kappa_1^{(\ell)}\right)^2 q}\notag\\
    \qquad \int \frac{d\xi}{(2\pi)^{\frac{3}{2}}}\frac{2e^{-\frac{1}{2}\frac{\Delta_a\int 
 z\dd \mu(z)\prod\limits_{\ell=1}^L \left(\kappa_1^{(\ell)}\right)^2\Delta_\ell+\epsilon_r+\prod\limits_{\ell=1}^L\left(\kappa_1^{(\ell)}\right)^2 q}{\Delta_a\int 
 z\dd \mu(z)\prod\limits_{\ell=1}^L \left(\kappa_1^{(\ell)}\right)^2\Delta_\ell+\epsilon_r-\prod\limits_{\ell=1}^L\left(\kappa_1^{(\ell)}\right)^2 q}\xi^2}}{1-\mathrm{erf}\left(\frac{\prod\limits_{\ell=1}^L\kappa_1^{(\ell)}\sqrt{q}\xi}{\sqrt{2\left(\Delta_a\int 
 z\dd \mu(z)\prod\limits_{\ell=1}^L \left(\kappa_1^{(\ell)}\right)^2\Delta_\ell+\epsilon_r-\prod\limits_{\ell=1}^L\left(\kappa_1^{(\ell)}\right)^2 q\right)}}\right)}
    \end{cases}
\end{align}
The interested reader is referred to \cite{Cui2019LargeDF} for a full derivation in a closely related case.

\subsection{Bayes optimal classification error}
We detail here the derivation of the Bayes misclassification error \eqref{eq:error_classif}. The Bayes optimal estimator for a test input $x$ is \cite{Opper1991GeneralizationPO}
\begin{equation}
    \mathrm{sign}\left[\left\langle  
    \mathrm{sign}\left(
    a^\top h_L(x)
    \right)
    \right\rangle_{a,\{W_\ell\}_\ell\sim\mathbb{P}}\right]
\end{equation}
where $h_L(x)$ is the post-activation of the student network 
 and $\mathbb{P}$ is the Bayes posterior
 $$
 \mathbb{P}(a,\{W_\ell\})=\frac{1}{Z}e^{-\frac{||a||^2}{2\Delta_a}-\sum\limits_{\ell}\frac{||W_\ell||_F^2}{2\Delta_\ell}
}\prod\limits_{\mu=1}^n\mathbb{E}_{\xi\sim\mathcal{N}(0,\Delta)}\left[
\delta\left(
y^\mu-\mathrm{sign}(\hat{y}(x^\mu)+\xi)
\right)
\right].
 $$
 Then
\begin{align}
    1-\epsilon_g&=\mathbb{E}_{\mathcal{D},a_\star,\{W^\star_\ell\}_\ell}\mathbb{E}_{x} \Theta\left[
   \frac{1}{\sqrt{k_L}}a_\star^\top h_L^\star(x)\times  \left\langle\mathrm{sign}\left(\frac{1}{\sqrt{k_L}}a^\top h_L(x)\right)\right\rangle
    \right]\notag\\
    &=\mathbb{E}_{\mathcal{D},a_\star,\{W^\star_\ell\}_\ell}\mathbb{E}_{x} \int \dd v \dd \hat{v}e^{-iv\hat{v}}\Theta\left[
   \frac{1}{\sqrt{k_L}}a_\star^\top h_L^\star(x)\times v
    \right]
    e^{i\hat{v}\left\langle\mathrm{sign}\left(\frac{1}{\sqrt{k_L}}a^\top h_L(x)\right)\right\rangle}\notag\\
    &=\mathbb{E}_{\mathcal{D},a_\star,\{W^\star_\ell\}_\ell}\mathbb{E}_{x} \int \dd v \dd \hat{v}e^{-iv\hat{v}}\Theta\left[
   \frac{1}{\sqrt{k_L}}a_\star^\top h_L^\star(x)\times v
    \right]\sum\limits_{k=0}^\infty 
    \frac{(i\hat{v})^k}{k!}\left\langle\mathrm{sign}\left(\frac{1}{\sqrt{k_L}}a^\top h_L(x)\right)\right\rangle^k\notag\\
    &= \int \dd v \dd \hat{v}e^{-iv\hat{v}}\sum\limits_{k=0}^\infty 
    \frac{(i\hat{v})^k}{k!}\mathbb{E}_{x}\mathbb{E}_{
    \mathcal{D},a_\star,\{W^\star_\ell\}_\ell}
    \Theta\left[
   \frac{1}{\sqrt{k_L}}a_\star^\top h_L^\star(x)\times v
    \right]\left\langle\mathrm{sign}\left(\frac{1}{\sqrt{k_L}}a^\top h_L(x)\right)\right\rangle^k.
\end{align}
The computation of the bracketed average can be carried out using the replica trick, observing
\begin{align}
    \left\langle\mathrm{sign}\left(\frac{1}{\sqrt{k_L}}a^\top h_L(x)\right)\right\rangle&=\mathbb{E}_{\mathcal{D},a_\star,\{W^\star_\ell\}_\ell}\underset{s\rightarrow 0}{\lim} Z^{s-1}\int da P_a(a)\prod\limits_{\ell}^LdW_\ell P_w(W_\ell) \prod\limits_{\mu=1}^nP_{\mathrm{out}}(y^\mu|\hat{y}(x^\mu))\mathrm{sign}\left(\hat{y}(x)\right)\notag\\
    &=\mathbb{E}_{\mathcal{D},a_\star,\{W^\star_\ell\}_\ell}\int \prod\limits_{a=1}^sda^a P_a(a^a)\prod\limits_{a=1}^s\prod\limits_{\ell}^LdW_\ell^a P_w(W_\ell^a) \prod\limits_{\mu=1}^n\prod\limits_{a=1}^sP_{\mathrm{out}}(y^\mu|\hat{y}^a(x^\mu)) \mathrm{sign}\left(\hat{y}^1(x)\right)
\end{align}
where $\hat{y}^1$ is the first replica of the student \eqref{eq:student}, with weights $a^a, \{W^a_\ell\}$. It follows that
\begin{align}
    \left\langle\mathrm{sign}\left(\frac{1}{\sqrt{k_L}}a^\top h_L(x)\right)\right\rangle ^k 
    &=\mathbb{E}_{\mathcal{D},a_\star,\{W^\star_\ell\}_\ell}\int \prod\limits_{a=1}^s\prod\limits_{\alpha=1}^k da^{a\alpha}P_a(a^{a\alpha})\prod\limits_{a=1}^s 
\prod\limits_{\alpha=1}^k\prod\limits_{\ell}^LdW_\ell^{a\alpha} P_w(W_\ell^{a\alpha})\notag\\ &~~~~\prod\limits_{\mu=1}^n\prod\limits_{\alpha=1}^k \mathrm{sign}\left(\hat{y}^{1\alpha}(x)\right)\prod\limits_{a=1}^sP_{\mathrm{out}}(y^\mu|\hat{y}^{a\alpha}(x^\mu)) ,
\end{align}
with the power $k$ effectively introducing a second level of replication. 
\begin{align}
    \mathbb{E}_{x}\mathbb{E}_{
    a_\star,\{W^\star_\ell\}_\ell}
    &\Theta\left[
   \frac{1}{\sqrt{k_L}}a_\star^\top h_L^\star(x)\times v
    \right]\left\langle\mathrm{sign}\left(\frac{1}{\sqrt{k_L}}a^\top h_L(x)\right)\right\rangle^k\notag\\ &
    =\mathbb{E}_{\mathcal{D},a_\star,\{W^\star_\ell\}_\ell}\int \prod\limits_{a=1}^s\prod\limits_{\alpha=1}^k da^{a\alpha}P_a(a^{a\alpha})\prod\limits_{a=1}^s 
\prod\limits_{\alpha=1}^k\prod\limits_{\ell}^LdW_\ell^{a\alpha} P_w(W_\ell^{a\alpha})\notag\\ &~~~~~~\mathbb{E}_{\{x^\mu\}_{\mu=1}^n}\prod\limits_{\mu=1}^n  \prod\limits_{a=1}^s\prod\limits_{\alpha=1}^kP_{\mathrm{out}}(y^\mu|\hat{y}^{a\alpha}(x^\mu)) \underbrace{\mathbb{E}_x \prod\limits_{\alpha=1}^k \mathrm{sign}\left(\hat{y}^{1\alpha}(x)\right)\Theta(v\times y(x) )}_{(a)}.\notag\\
\end{align}
Following the discussion in Appendix \ref{App:GET} (see also Section \ref{sec:setting} in the main text), and under the assumption that $\hat{y}^{1\alpha}(x), y(x)$ may be as a first approximation treated as Gaussian variables, it follows from the linearization \eqref{eq:Omega_Psi_multilayer} that $\hat{y}^{1\alpha}(x), y(x)$ can be replaced inside the expectation $\mathbb{E}_x$ in $(a)$ by their equivalent linear models,
\begin{align}  &\hat{y}^{1\alpha}_{\mathrm{lin}}(x):=\frac{1}{\sqrt{k_L}}a^\top_{1\alpha}\left[ 
    \kappa_*^{(L)}\xi^{\alpha}_{L}
    +\sum\limits_{\ell=2}^{L-1}\kappa_*^{(\ell-1)}\left(\prod\limits_{\ell^\prime=L}^{\ell}\kappa_1^{(\ell^\prime)}\frac{W_{\ell^\prime}^{1\alpha}}{\sqrt{k_{\ell^\prime-1}}}\right)
    \xi^\alpha_{\ell-1}
+\left(\prod\limits_{\ell=L}^{1}\kappa_1^{(\ell)}\frac{W_{\ell}^{1\alpha}}{\sqrt{k_{\ell-1}}}\right)x
    \right]\notag\\
& y_{\mathrm{lin}}(x):=\frac{1}{\sqrt{k_L}}a_\star^\top\left[ 
    \kappa_*^{(L)}\xi^{\star}_{L}
    +\sum\limits_{\ell=2}^{L-1}\kappa_*^{(\ell-1)}\left(\prod\limits_{\ell^\prime=L}^{\ell}\kappa_1^{(\ell^\prime)}\frac{W_{\ell^\prime}^\star}{\sqrt{k_{\ell^\prime-1}}}\right)
    \xi^\star_{\ell-1}
+\left(\prod\limits_{\ell=L}^{1}\kappa_1^{(\ell)}\frac{W_{\ell}^\star}{\sqrt{k_{\ell-1}}}\right)x
    \right]+\sqrt{\Delta}\xi
\end{align}
where $\xi^{\star/\alpha}_\ell\sim \mathcal{N}(0,I_{k_{\ell}})$ are independently sampled Gaussian noise vectors. Note that the replication procedure leading to the introduction  of the second replication level, indexed by $\alpha$, means that the $\xi^\alpha$ are also mutually independent. Finally, one can write 
\begin{align}
    (a)=\mathbb{E}_{x,\{\xi^\alpha_\ell\}_{\ell,\alpha},\{\xi^\star_\ell\}_\ell} \prod\limits_{\alpha=1}^k \mathrm{sign}\left(\hat{y}_{\mathrm{lin}}^{1\alpha}(x)\right)\Theta(v\times y_{\mathrm{lin}}(x) ).
\end{align}
The full equation for the test error is then
\begin{align}
    1-\epsilon_g&= \mathbb{E}_{x,\{\xi^\alpha_\ell\}_{\ell,\alpha},\{\xi^\star_\ell\}_\ell,\xi}\int \dd v \dd \hat{v}e^{-iv\hat{v}}\sum\limits_{k=0}^\infty 
    \frac{(i\hat{v})^k}{k!}
    \Theta\left[
   y_{\mathrm{lin}}(x)\times v
    \right]
    \underbrace{
    \mathbb{E}_{
    \mathcal{D},a_\star,\{W^\star_\ell\}_\ell}\left\langle\mathrm{sign}\left(\hat{y}_{\mathrm{lin}}(x)\right)\right\rangle^k}_{(b)}.
\end{align}
We henceforth focus on computing $(b)$. Resuming the replica computation,
\begin{align}
(b)&=\int \prod\limits_{a=1}^s\prod\limits_{\alpha=1}^k da^{a\alpha}P_a(a^{a\alpha})\prod\limits_{a=1}^s 
\prod\limits_{\alpha=1}^k\prod\limits_{\ell}^LdW_\ell^{a\alpha} P_w(W_\ell^{a\alpha})\notag\\ &~~~~\prod\limits_{\mu=1}^n\prod\limits_{\alpha=1}^k \mathrm{sign}\left(\hat{y}_{\mathrm{lin}}^{1\alpha}(x)\right)\mathbb{E}_{\mathcal{D},a_\star,\{W^\star_\ell\}_\ell}\prod\limits_{a=1}^s\prod\limits_{\alpha=1}^kP_{\mathrm{out}}(y^\mu|\hat{y}_{\mathrm{lin}}^{a\alpha}(x^\mu)).
\end{align}
Define the local fields
\begin{align}
    & z_a^\alpha:=\frac{1}{\sqrt{k_L}}a_{1\alpha}^\top\xi^\alpha_L,\notag\\
&z_\ell^\alpha:=\frac{1}{\sqrt{k_L}}a_{1\alpha}^\top\left(\prod\limits_{\ell^\prime=L}^{\ell}\frac{W_{\ell^\prime}^{1\alpha}}{\sqrt{k_{\ell^\prime-1}}}\right)
    \xi^\alpha_{\ell-1},\qquad \forall  2\le \ell\le L\notag\\
&z_1^\alpha:=\frac{1}{\sqrt{k_L}}a_{1\alpha}^\top\left(\prod\limits_{\ell=L}^{1}\frac{W_{\ell}^{1\alpha}}{\sqrt{k_{\ell-1}}}\right)x,
\end{align}
so that
\begin{align}
\hat{y}_{\mathrm{lin}}^{1\alpha}=\kappa_*^{(L)}z^\alpha_a +\sum\limits_{\ell=2}^{L-1}\kappa_*^{(\ell-1)} 
 \prod\limits_{\ell^\prime=\ell}^{L}\kappa_{1}^{(\ell)} z^\alpha_\ell +\prod\limits_{\ell=1}^{L}\kappa_{1}^{(\ell)} z^\alpha_1.
\end{align}
Enforcing the definitions of the local fields using the Fourier representation of Dirac peaks, $(b)$ becomes
\begin{align}
(b)&= \int \prod\limits_{\alpha=1}^k dz^\alpha_1 d\hat{z}_1^\alpha e^{ -z^\alpha_1 
 \hat{z}_1^\alpha} \left(\prod\limits_{\ell=2}^L   dz^\alpha_\ell d\hat{z}_\ell^\alpha  e^{ -z^\alpha_\ell\hat{z}_\ell^\alpha}\right)dz_a^\alpha d\hat{z}_a^\alpha e^{ -z^\alpha_a
 \hat{z}_a^\alpha} 
\int \prod\limits_{a=1}^s\prod\limits_{\alpha=1}^k da^{a\alpha}P_a(a^{a\alpha})\prod\limits_{a=1}^s 
\prod\limits_{\alpha=1}^k\prod\limits_{\ell}^LdW_\ell^{a\alpha} P_w(W_\ell^{a\alpha})
 \notag\\
 &~~~~\prod\limits_{\alpha=1}^k e^{\hat{z}_a^\alpha \frac{1}{\sqrt{k_L}}a_{1\alpha}^\top\xi^\alpha_L+\sum\limits_{\ell=2}^L \hat{z}^\alpha_\ell \frac{1}{\sqrt{k_L}}a_{1\alpha}^\top\left(\prod\limits_{\ell^\prime=L}^{\ell}\frac{W_{\ell^\prime}^{1\alpha}}{\sqrt{k_{\ell^\prime-1}}}\right)
    \xi^\alpha_{\ell-1}+\hat{z}_1^\alpha \frac{1}{\sqrt{k_L}}a_{1\alpha}^\top\left(\prod\limits_{\ell=L}^{1}\frac{W_{\ell}^{1\alpha}}{\sqrt{k_{\ell-1}}}\right)x
 }\notag\\ &~~~~\prod\limits_{\mu=1}^n\prod\limits_{\alpha=1}^k \mathrm{sign}\left(\hat{y}_{\mathrm{lin}}^{1\alpha}(x)\right)\mathbb{E}_{\mathcal{D},a_\star,\{W^\star_\ell\}_\ell}\prod\limits_{a=1}^s\prod\limits_{\alpha=1}^kP_{\mathrm{out}}(y^\mu|\hat{y}_{\mathrm{lin}}^{a\alpha}(x^\mu)).
\end{align}
The computation for the trace and energy potentials $\Psi_{t,y}$ follows the same steps as detailed in Appendix \ref{App:multilayer}, leading to the same contributions to the multi-layer free energy. On the other hand, the entropy potential $\Psi_w$ is changed, as it now includes the second line. As in Appendix \ref{App:multilayer}, we decompose $\Psi_w$ in input, hidden, and readout layer contributions and detail the derivation for each.\\

\paragraph{Input layer}. We detail the computation of the entropy contribution at the input layer $\Tilde{\Psi}_1$, where the tilde distinguishes the $\hat{z}$ dependent potential from the one computed in Appendix \ref{App:multilayer}. Note crucially that in particular $\Tilde{\Psi}_1$ has a part which is not proportional to the first replica index $s$, and therefore contributes non-vanishingly in the $s\rightarrow 0$ limit.
\begin{align}
e^{\Tilde{\Psi}_1}=&\prod\limits_{j=1}^d
    \int \prod\limits_{a=1}^s\prod\limits_{\alpha=1}^kdw^1_{a\alpha}P_w(w^1_{a\alpha})e^{\sum\limits_{a\alpha\le b\beta}\sigma_j\left(\hat{q}^L_1\right)_{a\alpha,b\beta}\frac{\left(w_{a\alpha}^{1\top}\prod\limits_{\ell=2}^LW_{a\alpha}^\ell a_{a\alpha}\right)\times\left(w_{b\beta}^{1\top}\prod\limits_{\ell=2}^LW_{b\beta}^\ell a_{b\beta}\right) }{\prod\limits_{\ell=1}^L k_\ell}}\notag\\
    &~~~~\times e^{+\sum\limits_{a\alpha} \sigma_j\hat{m}\frac{\left(w_{a\alpha}^{1\top}\prod\limits_{\ell=2}^LW_{a\alpha}^\ell a_{a\alpha}\right)\times\left((w^1_\star)_j^\top\prod\limits_{\ell=2}^{L^\star}W_\star^\ell a_\star\right)}{\prod\limits_{\ell=1}^L\sqrt{k_\ell}\prod\limits_{\ell=1}^{L^\star}\sqrt{k^\star_\ell}}
    +\sum\limits_{\alpha} \hat{z}_1^\alpha \frac{1}{\sqrt{k_L}}a_{1\alpha}^\top\left(\prod\limits_{\ell=L}^{2}\frac{W_{\ell}^{1\alpha}}{\sqrt{k_{\ell-1}}}\right)w^{1}_{1\alpha}\frac{1}{\sqrt{d}}x_j
    }.
\end{align}
Again, we assume replica symmetry with respect to the double replica, which together with the Nishimori conditions imply
\begin{equation}
\left(\hat{q}^L_1\right)_{a\alpha,b\beta}=\delta_{ab}\delta_{\alpha\beta}\hat{q},
\end{equation}
together with $\hat{m}=\hat{q}$. As in Appendix \ref{App:multilayer}, introduce the local fields
\begin{equation}
    \eta_{a\alpha}\equiv \frac{w_{a\alpha}^{1\top}\prod\limits_{\ell=2}^LW_{a\alpha}^\ell a_{a\alpha}}{\prod\limits_{\ell=1}^L\sqrt{k_\ell}},
\end{equation}
with statistics
\begin{align}
    \langle \eta_{a\alpha} \eta_{b\beta}\rangle
    =\delta_{ab}\delta_{\alpha\beta}\Delta_1\rho^L_2.
\end{align}
Therefore
\begin{align}
e^{\Tilde{\Psi}_1}&=\prod\limits_{j=1}^d
    \int \prod\limits_{a\alpha}\frac{d\eta_{a\alpha}}{\sqrt{2\pi \Delta_1\rho_2^L}}e^{-\frac{1}{2\Delta_1\rho_2^L}\eta_{a\alpha}^2}e^{\sum\limits_{a\alpha\le b\beta}\sigma_j\left(\hat{q}^L_1\right)_{a\alpha,b\beta}\eta_{a\alpha}\eta_{b\beta}+\sum\limits_{a\alpha} \sigma_j\hat{m}\eta_{\alpha}\eta^*_j
    +\sum\limits_{\alpha} \hat{z}_1^\alpha \eta_{1\alpha}\frac{1}{\sqrt{d}}x_j
    }\notag\\
    &=e^{sd\Psi_1}\times e^{\frac{1}{2}\sum\limits_{j=1}^d \frac{x_j^2}{d}\frac{\Delta_1 \rho_2^L+\hat{q}\sigma_j}{((\Delta_1 \rho_2^L)^{-1}+\sigma_j\hat{q})^2}\sum\limits_{\alpha=1}^k(\hat{z}^\alpha_1)^2
    +\frac{1}{2}\sum\limits_{j=1}^d \frac{x_j^2}{d}
    \frac{\hat{q}\sigma_j}{((\Delta_1 \rho_2^L)^{-1}+\sigma_j\hat{q})^2}\sum\limits_{\alpha,\beta}\hat{z}^\alpha_1\hat{z}^\beta_1
    +\sum\limits_{j=1}^d\sigma_j\hat{q}\eta^*_j\frac{x_j}{\sqrt{d}}\frac{1}{\Delta_1 \rho_2^L)^{-1}+\sigma_j\hat{q}}\sum\limits_{\alpha=1}^k\hat{z}^\alpha_1
    }\notag\\
    &=\int \frac{d\eta}{\sqrt{2\pi}}e^{-\frac{\eta^2}{2}}\prod\limits_{\alpha=1}^k\left[
e^{\frac{1}{2}\sum\limits_{j=1}^d \frac{x_j^2}{d}\frac{\Delta_1 \rho_2^L+\hat{q}\sigma_j}{((\Delta_1 \rho_2^L)^{-1}+\sigma_j\hat{q})^2}(\hat{z}^\alpha_1)^2
    +\sqrt{\sum\limits_{j=1}^d \frac{x_j^2}{d}
    \frac{\hat{q}\sigma_j}{((\Delta_1 \rho_2^L)^{-1}+\sigma_j\hat{q})^2}}\eta\hat{z}^\alpha_1
    +\sum\limits_{j=1}^d\sigma_j\hat{q}\eta^*_j\frac{x_j}{\sqrt{d}}\frac{1}{\Delta_1 \rho_2^L)^{-1}+\sigma_j\hat{q}}\hat{z}^\alpha_1
    }
    \right],
\end{align}
where we introduced a Hubbard-Stratonovitch field $\eta$ in the last line and took the $s\rightarrow 0 $ limit.\\

\paragraph{Hidden layers} In this paragraph, we turn to the middle layers $\ell\ge 2$.
\begin{align}
    e^{\tilde{\Psi}_{\ell_0}}&=\prod\limits_{j=1}^d\left[
    \int \prod\limits_{a\alpha }dw_{a\alpha} e^{-\frac{\Delta_{\ell_0}^{-1}}{2}\sum\limits_{a\alpha}w_{a\alpha}^\top w_{a\alpha} -\frac{1}{2}\hat{r}^L_{\ell_0}\sum\limits_{a\alpha}\left(
    \frac{w_{a\alpha}^\top\prod\limits_{\ell=\ell_0+1}^LW_{a\alpha}^\ell a_{a\alpha}}{\prod\limits_{\ell=\ell_0}^L\sqrt{k_\ell}}
    \right)^2  +\sum\limits_{a\alpha}\hat{z}^\alpha_{\ell_0}
    \frac{w_{1\alpha}^\top\prod\limits_{\ell=\ell_0+1}^LW_{1\alpha}^\ell a_{1\alpha}}{\prod\limits_{\ell=\ell_0}^L\sqrt{k_\ell}}
    \frac{(\xi^\alpha_{\ell_0-1})_j}{\sqrt{k_{\ell_0-1}}}}
    \right].
\end{align}
Introducing the local field
\begin{equation}
    \eta_{a\alpha}:=\frac{w_{a\alpha}^\top\prod\limits_{\ell=\ell_0+1}^LW_{a\alpha}^\ell a_{a\alpha}}{\prod\limits_{\ell=\ell_0}^L\sqrt{k_\ell}},
\end{equation}
with the statistics
\begin{align}
    \langle \eta_{a\alpha} \eta_{b\beta}\rangle
    =\delta_{ab}\delta_{\alpha\beta}\Delta_{\ell_0}\rho^L_{\ell_0+1},
\end{align}
this becomes
\begin{align}
    e^{\tilde{\Psi}_{\ell_0}}&=\prod\limits_{j=1}^d\left[
    \int \prod\limits_{a\alpha } \frac{d\eta_{a\alpha}}{\sqrt{2\pi \Delta_{\ell_0}\rho_{\ell_0+1}^L}}e^{--\frac{1}{2}\hat{r}^L_{\ell_0}\sum\limits_{a\alpha}\eta_{a\alpha}^2+\sum\limits_{a\alpha}\hat{z}^\alpha_{\ell_0}
    \eta_{a\alpha}
    \frac{(\xi^\alpha_{\ell_0-1})_j}{\sqrt{k_{\ell_0-1}}}}
    \right]\notag\\
    &=e^{sd\Psi_{\ell_0}}\times e^{\frac{1}{2}\sum\limits_{j=1}^d\Delta_{\ell_0}\rho_{\ell_0+1}^L\sum\limits_{\alpha=1}^k(\hat{z}^\alpha_{\ell_0})^2\frac{(\xi^\alpha_{\ell_0-1})_j^2}{k_{\ell_0-1}}}.
\end{align}

\paragraph{Readout layer} By the same token, the new readout layer entropy potential reads
\begin{align}
    e^{\Tilde{\Psi}_a}=e^{\frac{1}{2}\sum\limits_{j=1}^d\Delta_{L}\sum\limits_{\alpha=1}^k(\hat{z}^\alpha_{a})^2\frac{(\xi^\alpha_{L})_j^2}{k_{L}}}
\end{align}

\paragraph{End of the computation} Finally, putting all the pieces together, the computation of $(b)$ follows as
\begin{align}
    (b)=&\int \frac{d\eta}{\sqrt{2\pi}}e^{-\frac{1}{2}\eta^2}\Bigg[
\int \prod\limits_{\ell=1}^L d\hat{z}_\ell dz_\ell d\hat{z}_a dz_a e^{-\sum\limits_{\ell=1}^L z_\ell\hat{z}_\ell-\hat{z}_az_a+\frac{1}{2}\Delta_a \frac{\sum\limits_{j=1}^d(\xi_L)_j^2}{k_L}\hat{z}_a^2 +\frac{1}{2}\sum\limits_{\ell=2}^L\Delta_\ell\rho_{\ell+1}^L \frac{\sum\limits_{j=1}^d(\xi_\ell)_j^2}{k_{\ell-1}}\hat{z}_\ell^2  +\frac{1}{2}\Tilde{\Delta}_1(\hat{z}_1)^2+\mu \hat{z}_1  }\notag\\
& \qquad\qquad \qquad ~~~~
 \times \mathrm{sign}\left(\kappa_*^{(L)}z_a +\sum\limits_{\ell=2}^{L-1}\kappa_*^{(\ell-1)} 
 \prod\limits_{\ell^\prime=\ell}^{L}\kappa_{1}^{(\ell)} z_\ell +\prod\limits_{\ell=1}^{L}\kappa_{1}^{(\ell)} z_1\right)
    \Bigg]^k.
\end{align}
We introduced the shorthands
\begin{equation}
\begin{cases}
    \mu=\sqrt{\sum\limits_{j=1}^d \frac{x_j^2}{d}
    \frac{\hat{q}\sigma_j}{((\Delta_1 \rho_2^L)^{-1}+\sigma_j\hat{q})^2}}\eta
    +\sum\limits_{j=1}^d\sigma_j\hat{q}\eta^*_j\frac{x_j}{\sqrt{d}}\frac{1}{\Delta_1 \rho_2^L)^{-1}+\sigma_j\hat{q}}\\
    \Tilde{\Delta}_1=\sum\limits_{j=1}^d \frac{x_j^2}{d}\frac{\Delta_1 \rho_2^L+\hat{q}\sigma_j}{((\Delta_1 \rho_2^L)^{-1}+\sigma_j\hat{q})^2}
\end{cases}
\end{equation}
Note that the quantities $\sfrac{\sum\limits_{j=1}^d(\xi_L)_j^2}{k_L}$ and $\sfrac{\sum\limits_{j=1}^d(\xi_\ell)_j^2}{k_{\ell-1}}$ concentrate around their unit mean. Carrying out the $\hat{z}$ integrals:
\begin{align}
    (b)&=\int \frac{d\eta}{\sqrt{2\pi}}e^{-\frac{1}{2}\eta^2}\Bigg[
\int \frac{dz_a\prod\limits_{\ell}dz_\ell}{\sqrt{(2\pi)^{L+1}\Delta_a\Tilde{\Delta}_1\prod\limits_{\ell=2}^L\Delta_\ell\rho_{\ell+1}^L}} e^{-\frac{1}{2\Delta_a}z_a^2-\frac{1}{2}\sum\limits_{\ell=2}^L\frac{z_\ell^2}{\Delta_\ell \rho_{\ell+1}^L}-\frac{1}{2\Tilde{\Delta}_1}(z_1-\mu)^2}\notag\\
&\qquad\qquad\qquad\qquad\times\mathrm{sign}\left( \kappa_*^{(L)}z_a +\sum\limits_{\ell=2}^{L-1}\kappa_*^{(\ell-1)} 
 \prod\limits_{\ell^\prime=\ell}^{L}\kappa_{1}^{(\ell)} z_\ell +\prod\limits_{\ell=1}^{L}\kappa_{1}^{(\ell)} z_1\right)
    \Bigg]^k\notag \\
&=\int \frac{d\eta}{\sqrt{2\pi}}e^{-\frac{1}{2}\eta^2}\Bigg[
\int \frac{dz}{\sqrt{\scriptstyle 2\pi \Delta_z}} e^{-\frac{1}{2\Delta_z}\left(z-\mu\prod\limits_{\ell=1}^{L}\kappa_{1}^{(\ell)}\right)^2}\times\mathrm{sign}\left( z \right)
    \Bigg]^k\notag\\
    &=\int \frac{d\eta}{\sqrt{2\pi}}e^{-\frac{1}{2}\eta^2} \mathrm{erf}\left(
    \frac{\mu\prod\limits_{\ell=1}^{L}\kappa_{1}^{(\ell)}}{\sqrt{2 \Delta_z}}
    \right)^k
\end{align}

We introduced 
\begin{align}
    \Delta_z=\left(\kappa_*^{(L)}\right)^2\Delta_a +\sum\limits_{\ell=2}^{L-1}\left(\kappa_*^{(\ell-1)} 
 \prod\limits_{\ell^\prime=\ell}^{L}\kappa_{1}^{(\ell)}\right)^2 \Delta_\ell\rho_{\ell+1}^L+\prod\limits_{\ell=1}^{L}\left(\kappa_{1}^{(\ell)} \right)^2 \Tilde{\Delta}_1.
\end{align}
This concludes the computation of $(b)$. Returning to the computation of the classification error:
\begin{align}
    1-\epsilon_g&=\mathbb{E}_{x,\{\xi^\alpha_\ell\}_{\ell,\alpha},\{\xi^\star_\ell\}_\ell,\xi} \int \frac{d\eta}{\sqrt{2\pi}}e^{-\frac{1}{2}\eta^2} \int \dd v \dd \hat{v}e^{-iv\hat{v}}\sum\limits_{k=0}^\infty 
    \frac{(i\hat{v})^k}{k!}
    \Theta\left[
   y_{\mathrm{lin}}(x)\times v
    \right]
    \mathrm{erf}\left(
    \frac{\mu\prod\limits_{\ell=1}^{L}\kappa_{1}^{(\ell)}}{\sqrt{2 \Delta_z}}
    \right)^k\notag\\
    &= \mathbb{E}_{x,\{\xi^\alpha_\ell\}_{\ell,\alpha},\{\xi^\star_\ell\}_\ell,\xi} \int \frac{d\eta}{\sqrt{2\pi}}e^{-\frac{1}{2}\eta^2} \Theta\left[ 
    y_{\mathrm{lin}}(x)\times \mathrm{erf}\left(
    \frac{\mu\prod\limits_{\ell=1}^{L}\kappa_{1}^{(\ell)}}{\sqrt{2 \Delta_z}}
    \right)
    \right]\notag\\
    &= \mathbb{E}_{x,\{\xi^\alpha_\ell\}_{\ell,\alpha},\{\xi^\star_\ell\}_\ell,\xi} \int \frac{d\eta}{\sqrt{2\pi}}e^{-\frac{1}{2}\eta^2} \Theta\left[ 
    y_{\mathrm{lin}}(x)\times \mu
    \right]\notag\\
    &=1-\frac{1}{\pi}\arccos \left(
    \frac{\mathbb{E}_{x,\{\xi^\alpha_\ell\}_{\ell,\alpha},\{\xi^\star_\ell\}_\ell,\xi,\eta}[y_{\mathrm{lin}}(x)\times \mu]}{\sqrt{\mathbb{E}_{x,\{\xi^\alpha_\ell\}_{\ell,\alpha},\{\xi^\star_\ell\}_\ell,\xi,\eta}[y_{\mathrm{lin}}(x)^2]}\times \sqrt{\mathbb{E}_{x,\{\xi^\alpha_\ell\}_{\ell,\alpha},\{\xi^\star_\ell\}_\ell,\xi,\eta}[\mu^2]}}
    \right).
\end{align}
The average squared label can be computed as
\begin{align}
    \mathbb{E}_{x,\{\xi^\alpha_\ell\}_{\ell,\alpha},\{\xi^\star_\ell\}_\ell,\xi,\eta}[y_{\mathrm{lin}}(x)^2]&=\frac{1}{k_L}\left(\kappa_*^{(L)}\right)^2 a_\star^\top a_\star + \sum\limits_{\ell=2}^{L-1}\left(\kappa_*^{(\ell-1)}\right)^2\frac{1}{k_L}a_\star^\top\left(\prod\limits_{\ell^\prime=L}^{\ell}\kappa_1^{(\ell^\prime)}\frac{W_{\ell^\prime}^\star}{\sqrt{k_{\ell^\prime-1}}}\right)\left(\prod\limits_{\ell^\prime=L}^{\ell}\kappa_1^{(\ell^\prime)}\frac{W_{\ell^\prime}^\star}{\sqrt{k_{\ell^\prime-1}}}\right)^\top a_\star\notag\\
&\qquad+\frac{1}{k_L}a_\star^\top \left(\prod\limits_{\ell=L}^{1}\kappa_1^{(\ell)}\frac{W_{\ell}^\star}{\sqrt{k_{\ell-1}}}\right)
\Sigma
\left(\prod\limits_{\ell=L}^{1}\kappa_1^{(\ell)}\frac{W_{\ell}^\star}{\sqrt{k_{\ell-1}}}\right)^\top a_\star+\Delta\notag\\
&=\left(\kappa_*^{(L)}\right)^2 \Delta_a + \sum\limits_{\ell=2}^{L-1}\left(\kappa_*^{(\ell-1)}\right)^2\Delta_a\left(\prod\limits_{\ell^\prime=L}^{\ell}\left(\kappa_1^{(\ell^\prime)}\right)^2\Delta_{\ell^\prime}\right)+\Delta_a\left(\prod\limits_{\ell=1}^{L}\left(\kappa_1^{(\ell)}\right)^2\Delta_{\ell}\right)\frac{1}{d}\Tr\Sigma+\Delta\notag\\
&=\left(\kappa_*^{(L)}\right)^2 \Delta_a + \sum\limits_{\ell=2}^{L-1}\left(\kappa_*^{(\ell-1)}\right)^2\Delta_a\left(\prod\limits_{\ell^\prime=L}^{\ell}\left(\kappa_1^{(\ell^\prime)}\right)^2\Delta_{\ell^\prime}\right)+\Delta_a\left(\prod\limits_{\ell=1}^{L}\left(\kappa_1^{(\ell)}\right)^2\Delta_{\ell}\right)\int d\mu(z) z+\Delta,
\end{align}
where we used the fact that the scalar products in the first line all concentrate asymptotically, leveraging the assumption on $\Sigma$. The teacher/student correlation term can be computed as
\begin{align}
    \mathbb{E}_{x,\{\xi^\alpha_\ell\}_{\ell,\alpha},\{\xi^\star_\ell\}_\ell,\xi,\eta}[y_{\mathrm{lin}}(x)\times \mu]&=\prod\limits_{\ell=1}^L \kappa_1^{(\ell)} 
 \eta^{*\top }\frac{\hat{q}\Sigma}{(\Delta_1\rho_2^L)^{-1}+\hat{q}\Sigma}\Sigma\eta^*\notag\\
    &=\prod\limits_{\ell=1}^L \kappa_1^{(\ell)}\Tr[\frac{\hat{q}\Sigma^2\Delta_a^2\prod\limits_{\ell=1}^L\Delta_\ell^2}{1+\hat{q}\Sigma \Delta_a\prod\limits_{\ell=1}^L\Delta_\ell}]\notag\\
    &=\prod\limits_{\ell=1}^L \kappa_1^{(\ell)} \int d \mu(z)\frac{\hat{q}z^2\Delta_a^2\prod\limits_{\ell=1}^L\Delta_\ell^2}{1+\hat{q}z \Delta_a\prod\limits_{\ell=1}^L\Delta_\ell}\notag\\
    &=\prod\limits_{\ell=1}^L \kappa_1^{(\ell)} \times q,
\end{align}
where we used the definition of $\eta^*$ and the self-averaging of the scalar product in going from the first to the second line. Finally, the averaged squared student output is 
\begin{align}
    \mathbb{E}_{x,\{\xi^\alpha_\ell\}_{\ell,\alpha},\{\xi^\star_\ell\}_\ell,\xi,\eta}[\mu^2]&=\hat{q}\Tr[\frac{\Sigma^2}{\left((\Delta_1\rho_2^L)^{-1}+\hat{q}\Sigma\right)^2}]+\hat{q}^2\Tr[\frac{\Sigma^3 \Delta_a\prod\limits_{\ell=1}^L\Delta_\ell}{\left((\Delta_1\rho_2^L)^{-1}+\hat{q}\Sigma\right)^2}]\notag\\
    &=\Tr[\frac{\hat{q}\Sigma^2 \Delta_a^2\prod\limits_{\ell=1}^L\Delta_\ell^2}{1+\hat{q}\Delta_a\prod\limits_{\ell=1}^L\Delta_\ell\Sigma}]=\int d \mu(z)\frac{\hat{q}z^2\Delta_a^2\prod\limits_{\ell=1}^L\Delta_\ell^2}{1+\hat{q}z \Delta_a\prod\limits_{\ell=1}^L\Delta_\ell}=q.
\end{align}
Putting these contributions together, one finally reaches
\begin{align}
    \epsilon_g=\frac{1}{\pi}\arccos\left(
    \frac{\prod\limits_{\ell=1}^L \kappa_1^{(\ell)} \sqrt{q}}{\sqrt{\left(\kappa_*^{(L)}\right)^2 \Delta_a + \sum\limits_{\ell=2}^{L-1}\left(\kappa_*^{(\ell-1)}\right)^2\Delta_a\left(\prod\limits_{\ell^\prime=L}^{\ell}\left(\kappa_1^{(\ell^\prime)}\right)^2\Delta_{\ell^\prime}\right)+\Delta_a\left(\prod\limits_{\ell=1}^{L}\left(\kappa_1^{(\ell)}\right)^2\Delta_{\ell}\right)\int d\mu(z) z+\Delta}}
    \right)
\end{align}
which is \eqref{eq:error_classif}.

\subsection{ERM}
In full similarity to the regression case discussed in Appendix \ref{App:ERM}, sharp asymptotics of the classification error are given by the concomitant work of \cite{DRF2023}. The same characterizations can be obtained by using Theorem $1$ of\cite{Loureiro2021CapturingTL} on the equivalent shallow network \eqref{eq:linear_model}, using in their notations
\begin{align}
\begin{cases}
     &\rho=\scriptstyle \prod\limits_{\ell=1}^{L}\left(\kappa_1^{(\ell)}\right)^2\rho^{L}_1+\sum\limits_{\ell_0=1}^{L-1}\kappa_*^{(\ell_0)}\prod\limits_{\ell=\ell_0+1}^{L}\left(\kappa_1^{(\ell)}\right)^2\rho^{L}_{\ell_0+1}+\left(\kappa_*^{L}\right)^2\rho_a,\\
    &\Omega=\Sigma \notag\\    &\Phi=\Delta_a^{\frac{1}{2}}\prod\limits_{\ell=1}^{L}\Delta_\ell^{\frac{1}{2}}\kappa_1^{(\ell)} \Sigma
\end{cases} .
\end{align}
We recall that we consider, for classification, the noiseless case $\Delta=0$. The test error is given by
\begin{align}
    \epsilon_g=\frac{1}{\pi}\arccos\left(
\frac{m\prod\limits_{\ell=1}^{L}\kappa_1^{(\ell)}}{\sqrt{\left(\scriptstyle \prod\limits_{\ell=1}^{L}\left(\kappa_1^{(\ell)}\right)^2\rho^{L}_1+\sum\limits_{\ell_0=1}^{L-1}\kappa_*^{(\ell_0)}\prod\limits_{\ell=\ell_0+1}^{L}\left(\kappa_1^{(\ell)}\right)^2\rho^{L}_{\ell_0+1}+\left(\kappa_*^{L}\right)^2\rho_a\right) q}}
    \right),
\end{align}
with $\rho$ and $q$ being given by a risk-dependent system of equations. We examine in succession ridge classification and logistic regression.\\

\paragraph{Ridge classification} The saddle point equations which need to be solved in order to access the asymptotic limit of the test error for ridge classification then read
\begin{align}
\begin{cases}
\hat{V}=\frac{\alpha}{1+V}\\
\hat{q}=\alpha\frac{1+q-2\prod\limits_{\ell=1}^{L}\kappa_1^{(\ell)} \sqrt{\frac{2}{\pi\left({\scriptstyle \prod\limits_{\ell=1}^{L}\left(\kappa_1^{(\ell)}\right)^2\rho^{L}_1+\sum\limits_{\ell_0=1}^{L-1}\kappa_*^{(\ell_0)}\prod\limits_{\ell=\ell_0+1}^{L}\left(\kappa_1^{(\ell)}\right)^2\rho^{L}_{\ell_0+1}+\left(\kappa_*^{L}\right)^2\rho_a}\right)}}m+\Delta}{(1+V)^2}\\
\hat{m}=\sqrt{\frac{2}{\pi\left({\scriptstyle \prod\limits_{\ell=1}^{L}\left(\kappa_1^{(\ell)}\right)^2\rho^{L}_1+\sum\limits_{\ell_0=1}^{L-1}\kappa_*^{(\ell_0)}\prod\limits_{\ell=\ell_0+1}^{L}\left(\kappa_1^{(\ell)}\right)^2\rho^{L}_{\ell_0+1}+\left(\kappa_*^{L}\right)^2\rho_a}\right)}}\frac{\prod\limits_{\ell=1}^{L}\kappa_1^{(\ell)}\alpha}{1+V}
\end{cases}
&&\begin{cases}
V=\int \dd\mu(z)
\frac{z}{\lambda I_d+\hat{
V}z}
\\
q=\int \dd\mu(z)
\frac{\Delta_a \prod\limits_{\ell=1}^{L}\Delta_\ell \hat{m}^2z^3+\hat{q}z^2}{(\lambda I_d+\hat{
V}z)^2}
\\
m=\Delta_a \prod\limits_{\ell=1}^{L}\Delta_\ell \hat{m}\int \dd\mu(z)
\frac{z^2}{\lambda I_d+\hat{
V}z}
.
\end{cases}
\end{align}

\paragraph{Logistic regression} By the same token, introducing following \cite{Loureiro2021CapturingTL} the auxiliary functions
$$ Z(y,\omega,V):=\frac{1}{2}\left(
    1+\mathrm{erf}\left(\frac{y\omega}{\sqrt{2V}}\right)
    \right)
$$
and $f(y,\omega,V)$ defined as the solution of
$$
f(y,\omega,V)=\frac{y}{1+e^{y(Vf(y,\omega,V)+\omega)}}.
$$
Using the abuse of notations
\begin{align}
    &Z:=Z\left(y,\frac{m\prod\limits_{\ell=1}^{L}\kappa_1^{(\ell)}}{\sqrt{q}}\xi, {\scriptstyle \prod\limits_{\ell=1}^{L}\left(\kappa_1^{(\ell)}\right)^2\rho^{L}_1+\sum\limits_{\ell_0=1}^{L-1}\kappa_*^{(\ell_0)}\prod\limits_{\ell=\ell_0+1}^{L}\left(\kappa_1^{(\ell)}\right)^2\rho^{L}_{\ell_0+1}+\left(\kappa_*^{L}\right)^2\rho_a-\frac{m^2\prod\limits_{\ell=1}^{L}\left(\kappa_1^{(\ell)}\right)^2}{q}}\right)\\
    &f:=f(y,\sqrt{q}\xi,V)
\end{align}
the corresponding saddle point equations read
\begin{align}
\begin{cases}
\hat{V}=-\alpha \int \frac{d\xi e^{-\frac{\xi^2}{2}}}{\sqrt{2\pi}}\sum\limits_{y=\pm 1} Z \partial_\omega f\\
\hat{q}=\alpha \int \frac{d\xi e^{-\frac{\xi^2}{2}}}{\sqrt{2\pi}} \sum\limits_{y=\pm 1} Z  f^2 \\
\hat{m}=\alpha \int \frac{d\xi e^{-\frac{\xi^2}{2}}}{\sqrt{2\pi}}\sum\limits_{y=\pm 
           1}\partial_\omega Z  f
\end{cases}
&&\begin{cases}
V=\int \dd\mu(z)
\frac{z}{\lambda I_d+\hat{
V}z}
\\
q=\int \dd\mu(z)
\frac{\Delta_a \prod\limits_{\ell=1}^{L}\Delta_\ell \hat{m}^2z^3+\hat{q}z^2}{(\lambda I_d+\hat{
V}z)^2}
\\
m=\Delta_a \prod\limits_{\ell=1}^{L}\Delta_\ell \hat{m}\int \dd\mu(z)
\frac{z^2}{\lambda I_d+\hat{
V}z}
.
\end{cases}
\end{align}